\documentclass{article}

\usepackage{amsmath,amsfonts,bm}

\def\eqref#1{equation~\ref{#1}}

\def\1{\bm{1}}

\DeclareMathAlphabet{\mathsfit}{\encodingdefault}{\sfdefault}{m}{sl}
\SetMathAlphabet{\mathsfit}{bold}{\encodingdefault}{\sfdefault}{bx}{n}

\usepackage[dvipsnames, table]{xcolor}
\usepackage{hyperref}
\hypersetup{
    colorlinks=true, citecolor=MidnightBlue, linkcolor=OrangeRed, urlcolor=OrangeRed }
\usepackage{iclr2027_conference,times}
\setcitestyle{citesep={,},yysep={,}}
\iclrfinalcopy 
\usepackage[utf8]{inputenc}
\usepackage[T1]{fontenc}
\usepackage{url}
\usepackage{booktabs}
\usepackage{graphicx}
\usepackage{amsmath}
\usepackage{amssymb}
\usepackage{multirow}
\usepackage{array}
\usepackage{colortbl}

\usepackage[capitalise,nameinlink]{cleveref}
\crefname{figure}{Fig.}{Figs.} \Crefname{figure}{Figure}{Figures} \crefname{table}{Tab.}{Tabs.} \Crefname{table}{Table}{Tables}

\newcommand{\thickhline}{\noalign{\hrule height 0.9pt}}
\providecommand{\noident}{\noindent}
\newcommand{\g}[1]{\textcolor{gray}{#1}}
\newcommand{\wicon}[2]{\cellcolor{#1}\rule{0pt}{1.6em}\raisebox{-0.05em}{\makebox[1.2em][c]{\includegraphics[height=1.1em,width=1.1em,keepaspectratio]{figures/wbench_icon/#2}}}}
\definecolor{harnessfill}{RGB}{232,240,252}
\newcommand{\ocell}{\cellcolor{harnessfill}}

\usepackage{makecell}
\usepackage{subcaption}
\usepackage{float}
\usepackage{wrapfig}
\newfloat{algorithm}{t}{loa}\floatname{algorithm}{Algorithm}
\usepackage{tikz}

\usetikzlibrary{positioning,fit,arrows.meta,calc,backgrounds,decorations.pathmorphing,patterns}
\usepackage{pgfplots}
\pgfplotsset{compat=1.17}

\definecolor{harnessblue}{RGB}{219,231,250}
\definecolor{harnessbluedark}{RGB}{38,84,160}
\definecolor{stripbg}{RGB}{240,246,241}
\definecolor{stripframe}{RGB}{160,190,170}

\makeatletter
\newcommand{\figcaption}{\def\@captype{figure}\caption}
\newcommand{\tabcaption}{\def\@captype{table}\caption}
\makeatother
\definecolor{spatialorange}{RGB}{252,229,205}
\definecolor{spatialorangedark}{RGB}{196,106,25}
\definecolor{contgreen}{RGB}{214,238,222}
\definecolor{contgreendark}{RGB}{34,120,66}
\definecolor{modelgrey}{RGB}{236,236,238}
\definecolor{modelgreydark}{RGB}{90,90,96}
\definecolor{failred}{RGB}{233,88,88}
\definecolor{inputlav}{RGB}{238,232,248}
\definecolor{inputlavdark}{RGB}{108,84,170}

\newcommand{\sysname}{\textsc{Stateful Harness}}
\newcommand{\benchname}{WorldState-Bench}
\newcommand{\vsm}{VisualScript}

\title{State World Models: Execute the Rules, Generate the Pixels}

\definecolor{reslink}{HTML}{1B2A8C}
\newcommand{\projecturl}{https://ivg-yanranzhang.github.io/StateWorldModel/}
\newcommand{\githuburl}{https://github.com/Zhangyr2022/StateWorldModel}
\newcommand{\hfurl}{https://huggingface.co/datasets/Louis0411/StateWorldModel}
\newcommand{\msurl}{https://modelscope.cn/datasets/Louis0411/StateWorldModel/}
\newcommand{\globeicon}{\tikz[baseline=-0.62ex, line width=0.55pt, x=0.47em, y=0.47em, black]{%
  \draw (0,0) circle (1); \draw (0,0) ellipse (0.42 and 1);
  \draw (-1,0) -- (1,0); \draw (-0.86,0.5) -- (0.86,0.5); \draw (-0.86,-0.5) -- (0.86,-0.5);}}
\newcommand{\resicon}[1]{\raisebox{-0.2em}{\includegraphics[height=1.05em]{figures/icons/#1}}}
\newcommand{\reslink}[3]{\href{#1}{#2\hspace{0.35em}\textcolor{reslink}{#3}}}
\newcommand{\ressep}{\hspace{1.1em}{\color{black!50}\textbar}\hspace{1.1em}}
\newcommand{\resourcelinks}{\normalfont
  \reslink{\projecturl}{\globeicon}{Project Page}\ressep
  \reslink{\githuburl}{\resicon{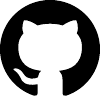}}{GitHub}\ressep
  \reslink{\hfurl}{\resicon{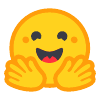}}{Hugging Face}\ressep
  \reslink{\msurl}{\resicon{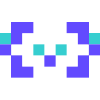}}{ModelScope}}

\author{\textbf{Chan Yan Ki}$^{1,*}$ ~~~
\textbf{Yanran Zhang}$^{1,*}$ ~~~
\textbf{Wenzhao Zheng}$^{1,\dagger}$ ~~~
\textbf{Xiaofeng Wang}$^{1}$ ~~~\\
\textbf{Zheng Zhu}$^{2}$ ~~~
\textbf{Lei Chen}$^{1}$ ~~~
\textbf{Jie Zhou}$^{1}$ ~~~
\textbf{Jiwen Lu}$^{1}$\\
$^1$ Department of Automation, Tsinghua University, China ~~~
$^2$ GigaAI\\[0.5em]
\resourcelinks
}

\begin{document}

\maketitle
\begingroup
\renewcommand{\thefootnote}{}
\footnotetext{$^*$ Equal contribution. 
\ \ 
$^\dagger$ Project leader.
}
\endgroup
\vspace{-9mm}
\begin{figure}[H]
\centering
\includegraphics[width=\textwidth]{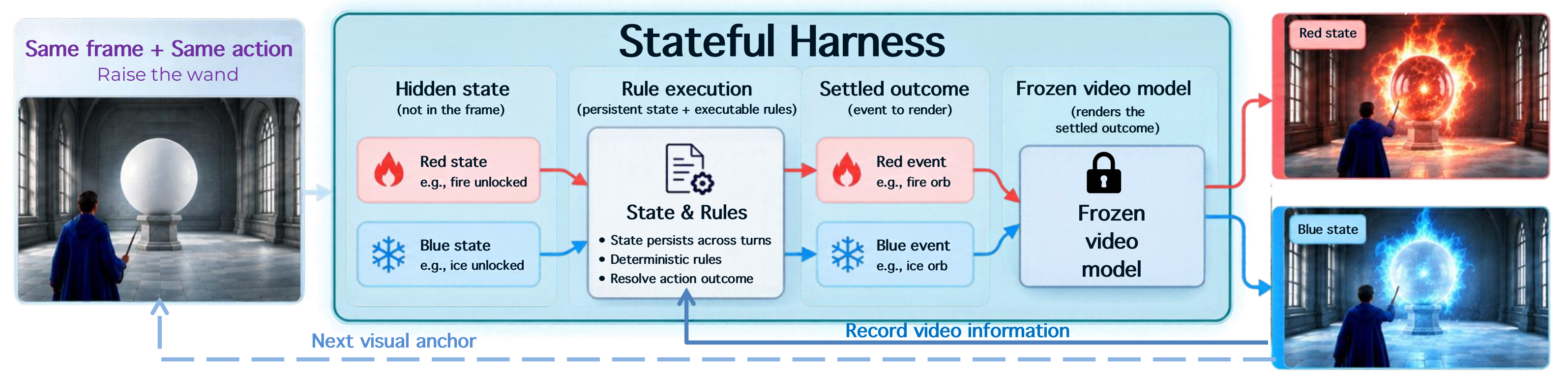}
     \vspace{-6mm}
\caption{\textbf{State World Models with \sysname{}.}
The same initial frame and action can lead to different outcomes under different world states.
The harness executes predefined rules to determine the event before a frozen video model renders it, illustrated by the red and blue elemental states.
The generated clip provides the visual anchor for the next turn.}
\label{fig:teaser}
\end{figure}
\vspace{-2mm}

\begin{abstract}
Existing world models can generate action-conditioned video from a single image.
Many focus on camera movement, allowing users to explore the generated scene.
However, navigation alone does not ensure that an action follows persistent rules when its outcome depends on state that is invisible in the current frame.
We present \textbf{\sysname{}}, a framework that separates world evolution from video rendering to support stateful and rule-governed interaction.
It maintains an evolving world state under specified rules, renders each resulting event with a frozen video model, and uses the generated clip to update the world state for the next turn.
At each turn, the harness resolves the action selected by the user and compiles the resulting event into an instruction, while a preview clip from an editable 3D scene provides spatial and camera guidance.
To evaluate rule compliance, we build \textbf{\benchname{}}, a benchmark of 50 counterfactual twin pairs that share an initial frame and action but differ in hidden state and required outcome.
Under rule-only prompting, the best of the 11 evaluated video backends produces the required endings for both cases in 32\% of pairs.
For MiniMax-H3, specifying the required endings raises this rate from 32\% to 64\%, while the full harness reaches 92\% with the same frozen model.
These results support separating rule execution from video rendering for world interactions with hidden state.
\end{abstract}
\section{Introduction}
\label{sec:intro}

Video generation has advanced rapidly, with recent models producing visually realistic videos with coherent motion \citep{wan2_2,hunyuanvideo15}.
Building on this progress, interactive video world models condition generation on navigation signals, allowing users to move through and explore generated environments \citep{parkerholder2025genie3,wang2026matrixgame3}.

Navigation control makes a generated scene explorable, but does not determine the consequences of interacting with it \citep{parkerholder2025genie3,wang2026matrixgame3}.
Some video world models support richer interactions by conditioning generation on player or agent actions \citep{valevski2024gamengen,zhu2026incantationnaturallanguageaction}.
However, a responsive video does not by itself ensure that an action has the outcome required by the world state and rules, or that its consequences persist across later interactions \citep{ding2026playworld,xu2026worldroambenchopenworldbenchmarklonghorizon}.
For example, the same action should open a locked chest only when the player holds a key, even if the initial views are identical (Figure~\ref{fig:counterfactual}).
Such distinctions require \emph{semantic state} to record variables such as inventory and progress, together with an addressable \emph{spatial state} to preserve scene layout and camera paths across turns.
While prior systems incorporate explicit state through programmable engines, learned game variables, or reconstructed NPC state \citep{huang2026programmable,lin2026stateplaystateawaregameworld,deng2026worldminddecoupledgameworld}, their evaluations focus on selected mechanics and do not establish whether a frozen video model can render outcomes prescribed by newly specified rules involving hidden state.
Explicit state alone does not ensure that the consequence required by its rules appears in the generated video.
The central challenge is therefore to make state transitions authoritative and render their consequences faithfully.

We present \sysname{}, a framework that separates world evolution from video rendering to support stateful interaction under specified rules (Figure~\ref{fig:teaser}).
It maintains an evolving world state under specified rules, renders each resulting event with a frozen video model, and uses the generated clip to update the world state for the next turn.
Given a world description and an initial image, a coding agent defines the world state, available actions, and executable rules, while an editable 3D scene represents the spatial layout.
The rules remain fixed as the world state evolves through user actions.
At each turn, the user observes the current scene and selects an action, which the runtime resolves against the current state before rendering.
The resolved event is compiled into a \vsm{} instruction, and a preview clip from the updated 3D scene provides spatial and camera guidance.
A frozen video model renders the event from these inputs and a visual anchor drawn from the initial image or preceding clip.
A vision language model then interprets the generated clip and proposes updates to the semantic and spatial state based on the observed events.
The runtime incorporates updates compatible with the fixed rules, and the resulting state determines the consequences of subsequent actions.
Together, rule execution, video generation, and visual observation form a closed loop that yields \emph{state world models}.

To evaluate whether an action yields a visible event consistent with hidden state, we construct \benchname{} with a Change-to-Comply protocol.
Change assesses whether the action produces a discernible event, and Comply assesses whether that event satisfies an executable rule.
Each of the 50 cases comprises a counterfactual pair with the same initial frame, action, and rule program but different hidden values that require distinct outcomes.
We introduce \textsc{PairFollow} and \textsc{PairStrict} as two metrics for joint evaluation of each counterfactual pair.
\textsc{PairFollow} requires both clips to show their required endings, whereas \textsc{PairStrict} additionally requires the forbidden ending to be absent from both clips.
Among eleven unwrapped video backends, the highest \textsc{PairFollow} score under the Rule-only condition is 32\%.
For MiniMax-H3, explicitly specifying the required ending increases \textsc{PairFollow} from 32\% to 64\%, although \textsc{PairStrict} reaches only 4\%.
With MiniMax-H3, the complete \sysname{} framework achieves 92\% \textsc{PairFollow} and 90\% \textsc{PairStrict} on the same cases.
Beyond the single-turn probe, the same closed loop carries state across longer episodes, as in the multi-turn duel of Figure~\ref{fig:demo}, and extends training-free to worlds whose custom rules and state are written as a new world definition.
\section{Related Work}
\label{sec:related}

\textbf{World Models.}
World models study action-conditioned predictions~\citep{micheli2023iris,hafner2025dreamer4,ha2018worldmodels}.
The Genie series introduces latent actions and promptable environments \citep{bruce2024genie,parkerholder2024genie2,parkerholder2025genie3}, while Oasis demonstrates interaction in an open world \citep{decart2024oasis}.
Other systems expand scene range, streaming, and instruction following \citep{he2025matrixgame2,wang2026matrixgame3,gao2026lingbot,yin2026evoke,dreamxteam2026dreamxworld10generalpurposeinteractive,chen2026h3world}, separate appearance from actions \citep{yu2025gamefactory}, use game corpora \citep{che2024gamegenx,kanervisto2025wham,guo2025mineworld,yang2024playgen,tang2025hgc2}, or address memory and visual consistency \citep{xiao2025worldmem,yu2025context,li2025vmem,hu2026multiplayerwm,garcin2026persist}.
Building on action-conditioned rendering, \sysname{} uses executable rules to resolve each action against explicit semantic and spatial state before a frozen video model renders the resulting event.

\textbf{Geometric Control of Video Generation.}
Camera-conditioned systems inject trajectories through learned pose or motion representations \citep{wang2024motionctrl,yang2024directavideo,he2024cameractrl,xu2024camco,bahmani2024vd3d}.
Other methods reconstruct geometry for training-free control \citep{hou2024camtrol} or use boxes, source videos, and reference clips \citep{lin2024videodirectorgpt,bai2025recammaster,jiang2025vace}.
MoGe4D generates trajectories over recovered geometry \citep{moge4d}.
The \sysname{} uses a related interface to carry an editable scene across turns, with rule execution determining which spatial edits and semantic events are sent to the renderer.

\textbf{Explicit State and Rules.}
Rules can be searched as generative objects or compiled from language \citep{salen2003rules,cook2013mechanicminer,hsu2026ifcargo}.
PWM couples an engine and a 3D oriented-box representation with conditioning at the pixel level \citep{huang2026programmable}.
Code World Model compiles a code-maintained state into proxy video and fine-tunes its renderer \citep{chen2026codeworld}.
Other work generates rule code without a video renderer \citep{lehrach2025cwm}, recovers state from generated frames \citep{deng2026worldminddecoupledgameworld}, or conditions generation on selected state variables \citep{lin2026stateplaystateawaregameworld,meng2026marionettepredictingworldstates,zhu2026incantationnaturallanguageaction}.
Narrative and character state can also be generated together \citep{li2025unbounded,cheng2025animegamer}.
We combine a persistent editable scene, a semantic state ledger, and the standard reference inputs of a frozen renderer.
This combination is minimal: new rule-governed worlds reduce to a small typed program, no engine or fine-tuning.
Table~\ref{tab:pwm} compares these interfaces and evaluation protocols.

\textbf{World Modeling Evaluation.}
Most existing benchmarks measure visual quality, prompt adherence, navigation, subject actions, event editing, and perspective changes \citep{wbench2026,xu2026worldmark,xu2026worldroambenchopenworldbenchmarklonghorizon,zhao2026worldolympiad,ding2026playworld,fang2026iworldbench,yang2026worldexam}.
Camera and geometry states can be tracked or remembered \citep{nam2026worldcam,yu2025context,li2025vmem}.
These measurements are valuable for navigability and continuity, but they do not directly test whether identical images and actions produce different rule-mandated outcomes when only hidden state changes.
Concurrent work tests whether a shell-game arrangement is retained in the model's internal state across a swap chain \citep{shin2026shellgame}. \benchname{} instead keeps the hidden state in an explicit rule program and tests whether the mandated outcome is executed in rendered video.
Benchmarks inside executable games can compare behavior with engine state \citep{hafner2022crafter,bard2020hanabi,light2023avalonbench,ouyang2026gameworld,la2026gameenginebench,luo2026gamecraft}, but require an existing game or simulator.
\benchname{} supplies an explicit rule program and tests its consequences in generated video.
\section{Method: The \sysname{}}
\label{sec:method}
\label{sec:formal}
\label{sec:dualchannel}

\begin{figure}[t]
\centering
\includegraphics[width=\textwidth]{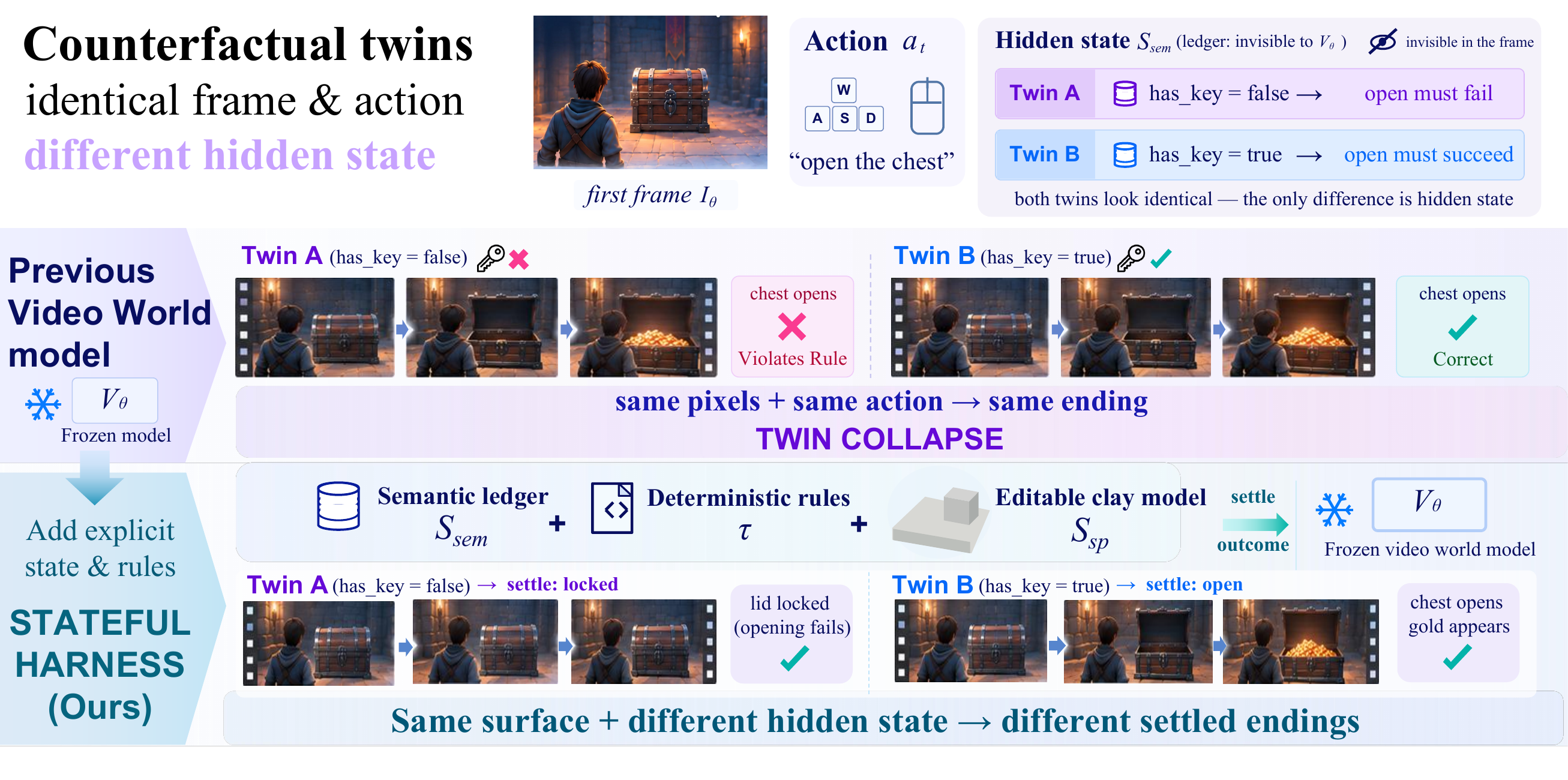}
    \vspace{-7mm}
\caption{\textbf{Counterfactual Pairs Reveal the Role of Hidden State.}
The two cases share the first frame $I_0$ and action $a_t$ but differ in the hidden flag \texttt{has\_key}.
In the illustrated failure, an unwrapped video model opens the chest in both cases, violating the rule when no key is available.
The harness maintains semantic and spatial state, executes the shared rule to determine whether the chest remains locked or opens, and renders each outcome with the same frozen video model.}
\label{fig:counterfactual}
    \vspace{-2mm}
\end{figure}

\noident\textbf{Framework Overview.}
A video model can depict an interaction, but a generated clip alone does not specify which changes should persist or whether those changes satisfy the rules of the world.
Figure~\ref{fig:counterfactual} shows how identical frames and actions require different outcomes when the hidden state differs.
Inspired by the role of LLM harnesses, we place a stateful runtime around a frozen video model to separate rule execution, visual realization, and observation of the generated result.

\begin{figure}[t]
\centering
\includegraphics[width=\textwidth]{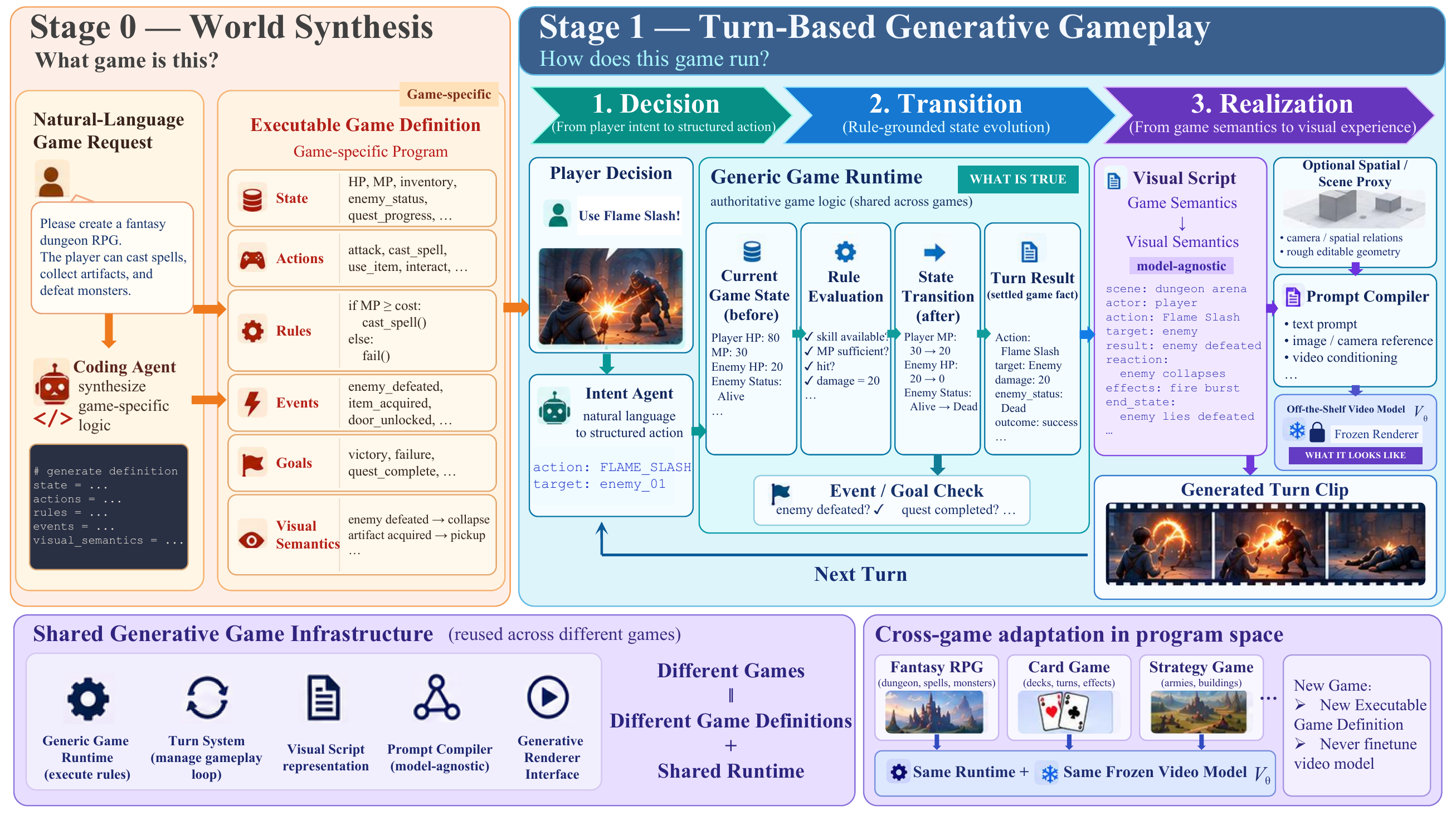}
    \vspace{-6mm}
\caption{\textbf{The \sysname{}.}
In \textbf{Stage 0}, a coding agent constructs an executable game definition specifying the state, actions, rules, events, goals, and visual semantics.
In \textbf{Stage 1}, the runtime evaluates the rules, updates the state, and determines the event and goal status for each action.
The resolved event is compiled into an instruction for the frozen video model $V_\theta$.
Visual observations of the generated clip inform the state used in the next turn.
Adaptation to a new game changes the game definition while retaining the shared runtime and video model weights.}
\label{fig:framework}
    \vspace{-1mm}
\end{figure}

Given a world description and an initial image, a coding agent constructs an executable game definition containing the state schema, available actions, rules, events, goals, and visual semantics.
The harness initializes a semantic ledger and an editable 3D scene, while the game definition remains fixed throughout the episode.
At each turn, the user observes the current scene and selects an action, which the runtime resolves against the current state and rules to determine the event and provisional state.
The resolved event is compiled into a \vsm{} instruction, and a preview clip from the updated 3D scene provides spatial and camera guidance.
A frozen video model generates the next clip from these inputs and a visual anchor drawn from the initial image or preceding clip.
A vision language model then interprets the generated events and proposes updates to the semantic and spatial state.
The runtime checks these proposals against the fixed rules and incorporates compatible updates into the world state used to resolve the next action.
The final frame also becomes the visual anchor for the next generation step.
Together, rule execution, video generation, and visual observation form a closed loop that couples explicit world state with a frozen video model, yielding a \emph{state world model}.
Figure~\ref{fig:framework} illustrates world construction and the rule to render path.

We represent the evolving world state as $s_t=(z_t,u_t)\in\mathcal{S}_{\mathrm{sem}}\times\mathcal{S}_{\mathrm{sp}}$.
The semantic component $z_t$ records resources, entity attributes, progress, attitudes, and terminal status that may be invisible in the current frame.
The spatial component $u_t$ records an editable entity layout, scale, and camera path.
A fixed rule program $R$ determines which events an action permits and which state changes those events require.
The harness updates the world state through a closed loop,
\begin{equation}
s_{t+1} = \mathcal{H}_R(s_t,\,a_t,\,o_{t+1}).
\label{eq:state}
\end{equation}
Here $o_{t+1}$ denotes observations extracted by a vision language model from the generated clip, and $\mathcal{H}_R$ combines rule execution with compatible updates derived from these observations.
The next state therefore incorporates both the consequences of the action and feedback from the generated video under the fixed rules $R$.
The visual anchor $f_t$ is initialized from the supplied image and subsequently updated from the final frame of each clip.

\subsection{Stage 0: World Construction}
\label{sec:setup}

World construction produces an executable game definition, an initial frame, a semantic ledger, and an editable spatial scene.
The game definition fixes the rules for an episode, whereas the ledger and scene change as interactions unfold.

\textbf{Executable Rules.}
From the world description, a coding agent writes the game definition in a small typed language.
Each rule has a guard that reads the current state and action, and an effect that specifies ledger updates, spatial edits, and an event.
Goal conditions inspect the ledger to determine whether an episode has ended.
The agent validates and freezes the definition before interaction begins.
In the counterfactual benchmark, both members of a pair share this definition.

\textbf{Initial Frame and Spatial State.}
The benchmark supplies an audited first image shared by each counterfactual pair.
For a new world, setup accepts a supplied image.
A feed forward geometry model lifts the image into instance boxes and initializes $u_0$ as an untextured, editable 3D scene with a ground plane and labeled cuboids.
We call this scene the \emph{clay model}.
Its explicit geometry supports object edits and procedural camera previews across turns.

\textbf{Semantic State.}
The ledger is a serializable record of typed fields for resources, entity attributes, progress counters, attitudes, and terminal status.
A twin specification declares the theme, twin identifier, and seeds.
The initializer assigns different hidden values $z_A \neq z_B$ while preserving the same first frame and spatial layout.
The rules therefore require different outcomes for the same visible action, even though rendering remains stochastic.

\subsection{Stage 1: State Evolution and Video Generation}
\label{sec:ir}

A turn begins when the user selects an action $a_t$ after observing the current scene.
In open interaction an intent agent can map a user instruction to a structured action.
The runtime first settles the rule governed outcome, then compiles and renders it, and finally observes the generated clip for changes relevant to the next turn.
This order gives fixed rules authority over action consequences while allowing generated events to inform the evolving world state.

\textbf{Rule Settlement.}
The rule program evaluates its guards against $s_t$ and $a_t$ before video generation.
It produces a provisional semantic state $\bar z_{t+1}$, a rule event $e_t$, and a spatial edit script $\epsilon_t$.
The edit script may add, move, or remove a box or change the camera path.
The provisional spatial state is obtained by applying these edits to the clay model.
\begin{equation}
(\bar z_{t+1},\,\epsilon_t,\,e_t) \;=\; \tau_R(s_t,\,a_t),
\qquad
\bar u_{t+1} \;=\; \mathrm{apply}(u_t,\,\epsilon_t),
\qquad
\bar s_{t+1} \;=\; (\bar z_{t+1},\,\bar u_{t+1}).
\label{eq:settle}
\end{equation}
An optional neural kernel $\rho$ may propose dialogue or freeform narration for the settled event.
Schema validation, value clamping, bounded retry, and fallback to the rule defined default constrain its proposals to the allowed set $\mathcal{A}_{\tau}(s,a)$.
Thus the gated kernel $\mathcal{G}\circ\rho$ satisfies
\begin{equation}
\mathrm{Supp}\big((\mathcal{G}\circ\rho)(\cdot \mid s,a)\big)
\;\subseteq\; \mathcal{A}_{\tau}(s,a).
\label{eq:bounded}
\end{equation}
The proposal may vary how an allowed event is expressed but cannot change the rule mandated transition in Eq.~\ref{eq:settle}.
\label{sec:ledger}

\textbf{Spatial Guidance.}
The runtime applies $\epsilon_t$ to the clay model before rendering so that the preview represents the provisional spatial state.
A procedural camera program renders this into a preview clip $r_t$, which is an editable representation of layout and cameras rather than the final appearance.
\label{sec:spatial}

\textbf{Instruction Compilation.}
The \vsm{} compiler translates the settled state and event into a typed visual script with slots for actor, action, target, reaction, dialogue, effects, end state, and shots.
A fixed compiler maps this record to the text instruction $p_t$, so identical inputs produce identical instructions.
The text carries semantic facts that an oriented box stream cannot express, while the preview supplies layout and camera guidance.
The instruction identifies the preview as a spatial reference to discourage the backend from reproducing its clay appearance.

\textbf{Frozen Video Rendering.}
The video model receives the instruction as text, the preview clip as video reference, and the visual anchor $f_t$ as image reference.
It generates a clip $v_{t+1}$ depicting the settled event, and the last frame becomes the anchor for the next turn.
\begin{equation}
(p_t,\,r_t) \;=\; \mathcal{C}(\bar z_{t+1},\,\bar u_{t+1},\,e_t),
\qquad
v_{t+1} \;=\; V_\theta(p_t,\,r_t,\,f_t),
\qquad
f_{t+1} \;=\; \mathrm{last}(v_{t+1}).
\label{eq:realize}
\end{equation}
The model weights remain frozen, and the runtime uses the stock text, image, and video reference interfaces of a compatible backend.
Normal operation makes one generation request per turn, with bounded retries for transient errors.
On a scene change, the anchor chain can be reset and appearance information is carried in the instruction instead.

\textbf{Visual Observation and State Update.}
A vision language model examines the generated clip together with the settled event and provisional state.
It identifies visible events and proposes changes to semantic or spatial fields that the video makes observable.
The runtime admits a proposed change only when it is well formed and compatible with the fixed rule program.
A visual detail may therefore become part of the next state, while an apparent event that contradicts the settled outcome is recorded as a rendering inconsistency and cannot overturn that outcome.
We express this step as
\begin{equation}
o_{t+1} \;=\; O_\psi(v_{t+1},\,\bar s_{t+1},\,e_t),
\qquad
s_{t+1} \;=\; U_R(\bar s_{t+1},\,o_{t+1}),
\label{eq:observe}
\end{equation}
where $O_\psi$ proposes observations and $U_R$ validates and incorporates permitted updates.
The next action is evaluated against $s_{t+1}$, closing the interaction loop without changing the rules $R$.
The counterfactual evaluation in this paper tests rendered compliance.

\section{Benchmark: \benchname{}}
\label{sec:bench}

\noident\textbf{Overview.}
Existing benchmarks evaluate visual quality, prompt adherence, navigation, subject actions, event editing, and camera or geometric consistency \citep{wbench2026,xu2026worldmark,xu2026worldroambenchopenworldbenchmarklonghorizon,zhao2026worldolympiad,ding2026playworld,fang2026iworldbench,yang2026worldexam,nam2026worldcam}.
These evaluations do not directly isolate whether a change in hidden state produces the distinct visible consequence prescribed by an executable rule.
Benchmarks based on executable games assess behavior conditioned on state provided by existing engines \citep{hafner2022crafter,bard2020hanabi,light2023avalonbench,ouyang2026gameworld,la2026gameenginebench,luo2026gamecraft}.
Worlds initialized from a description and an image lack this predefined engine state.
To evaluate such worlds, \benchname{} uses counterfactual pairs with identical visible inputs, different hidden values, and expected outcomes determined by an executable rule before generation.
Joint evaluation of each pair tests whether both required outcomes are rendered correctly under the corresponding hidden states.

We formalize this with the Change-to-Comply protocol.
\emph{Change} assesses whether an action produces a discernible event beyond camera motion.
\emph{Comply} assesses whether the event satisfies the rule for the specified hidden state.
This separates the absence of an event from the generation of an event that contradicts the rule.
The primary metrics quantify Comply, while clips with no discernible event or unjudgeable content are recorded as \textsc{Unclear} and receive no credit.

\phantomsection
\noindent\textbf{Counterfactual Hidden-State Pairs.}
\label{sec:twolayer}
Each case contains two counterfactual instances, A and B, that share the same first frame and action script.
They differ in one hidden value $z_A \neq z_B$ that is not observable in the shared frame.
The same rule program requires different visible endings for the two instances.
A correct ending in an isolated clip does not establish sensitivity to hidden state, whereas identical endings across a pair necessarily violate the required outcome for at least one instance.
We therefore count a pair as compliant only when both clips show their required endings.
A stricter criterion additionally requires the forbidden ending to be absent throughout both clips.

\phantomsection
\noindent\textbf{Case Construction.}
\label{sec:data}
The benchmark comprises 50 cases drawn from 49 hidden variables and 21 scenario themes.
Hidden values select which object, character, or elemental form carries the effect.
Each hidden variable must be unobservable in the first frame and must select rule branches with visibly different outcomes.
FLUX.2 \citep{flux2} generates the shared first frame at $1280{\times}720$, and each frame is audited for visual cues that reveal the hidden value.
This audit forms part of benchmark construction.
Authors specify each case program in a typed language that determines the initial states, required endings, and fixed choice judging questions.
The harness executes the same rules used to derive the expected outcomes, so the evaluation measures whether the generated videos depict these outcomes.

We compile two instruction conditions from each case specification.
\emph{Rule-only} provides the original case instruction, hidden value, selected rule branch, and action.
\emph{Rule+end} additionally specifies the required terminal outcome in an explicit final state sentence.
Table~\ref{tab:main} reports results under these conditions, while the harness uses the \vsm{} instruction described in \S\ref{sec:ir}.

\phantomsection
\noindent\textbf{Judging.}
\label{sec:judging}
We use the multimodal Qwen3.8-27B-VL \citep{qwen3} to answer fixed choice questions from sampled video frames.
The questions identify the ending shown and determine whether the forbidden outcome appears at any point in the clip.
At the clip level, \textsc{EndFollow} requires the prescribed ending, \textsc{CntAbsent} requires the absence of the forbidden outcome, and \textsc{Strict} requires both conditions.
At the pair level, \textsc{PairFollow} and \textsc{PairStrict} require both clips to satisfy \textsc{EndFollow} and \textsc{Strict}, respectively.
\textsc{TwinContrast} records whether the judged endings differ between the two clips.
The judge receives the expected outcomes for the relevant rule branch, but neither the hidden variable nor the rule text.

\section{Experiments}
\label{sec:experiments}

\subsection{Experiment Setup}
\label{sec:expsetup}

The rule compliance experiments use the 50 counterfactual pairs in \benchname{}, with 100 clips for individual evaluation and 50 pairs for joint evaluation (\S\ref{sec:data}).
Each backend generates one clip of a few seconds per action at its native resolution and frame rate.
Clips contain 77 to 128 frames at 15 to 24\,fps, with resolutions from $592{\times}336$ to $1280{\times}720$.
The harness generates clips at $864{\times}480$ and 24\,fps.
We evaluate eleven existing backends without additional training or fine-tuning (Table~\ref{tab:main}).
Four image-to-video generators receive instruction text and the initial frame, namely Wan2.2 \citep{wan2_2}, LongCat-Video \citep{longcatvideo}, HunyuanVideo 1.5 \citep{hunyuanvideo15}, and MiniMax-H3 \citep{chen2026h3world}.
Seven interactive world models receive inputs through their standard action interfaces, namely LingBot-World v2 \citep{gao2026lingbot}, Echo-WM Flash \citep{echowm}, FantasyWorld \citep{fantasyworld}, AlayaWorld \citep{alayaworld}, InSpatio-World \citep{inspatio}, Astra \citep{astra}, and HY-WorldPlay \citep{worldplay2025}.
The Qwen3.8-27B-VL judge evaluates frames sampled at 3\,fps together with the final frame.
Generation and judging use eight A800 80\,GB GPUs.

\begin{table}[t]
\centering
\caption{Main results on \benchname{} under Rule-only instructions (Track A) and Rule+end instructions (Track B).
Bold indicates the best result and underline the best among the unwrapped baselines.
The highlighted row reports \sysname{} with text and first-frame reference inputs.
Rows within a track share the probe, first frame, and judge. The matched comparison for the highlighted row is the same backbone under the Rule+end probe (Appendix Table~\ref{tab:bothprobes}).
}
\label{tab:main}
\scriptsize
\setlength{\tabcolsep}{2pt}
\resizebox{\textwidth}{!}{%
\begin{tabular}{@{}l ccc ccc@{}}
\toprule
 & \multicolumn{3}{c}{\textbf{A. Rule-only probe}} & \multicolumn{3}{c}{\textbf{B. Rule+end probe}} \\
\cmidrule(lr){2-4}\cmidrule(lr){5-7}
Backend & \textsc{EndFollow}$\uparrow$ & \textsc{PairFollow}$\uparrow$ & \textsc{TwinContrast}$\uparrow$
 & \textsc{Strict}$\uparrow$ & \textsc{CntAbsent}$\uparrow$ & \textsc{PairStrict}$\uparrow$ \\
\midrule
\multicolumn{7}{@{}l}{\emph{Image-to-video generators}}\\
\hspace{0.9em}Wan2.2 A14B \citep{wan2_2}       & 65\,\% & \underline{32\,\%} & 38\,\% & \underline{54\,\%} & 63\,\% & \underline{26\,\%} \\
\hspace{0.9em}LongCat-Video 13.6B \citep{longcatvideo} & 47\,\% & 8\,\% & 24\,\% & 46\,\% & 62\,\% & 8\,\% \\
\hspace{0.9em}HunyuanVideo 1.5 \citep{hunyuanvideo15}  & 57\,\% & 20\,\% & 26\,\% & 29\,\% & 30\,\% & 4\,\% \\
\hspace{0.9em}MiniMax-H3 \citep{chen2026h3world}        & \underline{66\,\%} & \underline{32\,\%} & 32\,\% & 29\,\% & 30\,\% & 4\,\% \\
\midrule
\multicolumn{7}{@{}l}{\emph{Interactive world models}}\\
\hspace{0.9em}LingBot-World v2 \citep{gao2026lingbot}  & 48\,\% & 16\,\% & 38\,\% & 44\,\% & 76\,\% & 14\,\% \\
\hspace{0.9em}Echo-WM Flash \citep{echowm}     & 57\,\% & 26\,\% & \underline{42\,\%} & 43\,\% & 60\,\% & 8\,\% \\
\hspace{0.9em}FantasyWorld \citep{fantasyworld}      & 34\,\% & 8\,\% & 30\,\% & 35\,\% & 81\,\% & 8\,\% \\
\hspace{0.9em}AlayaWorld \citep{alayaworld}        & 37\,\% & 12\,\% & 28\,\% & 33\,\% & 85\,\% & 6\,\% \\
\hspace{0.9em}InSpatio-World \citep{inspatio}    & 19\,\% & 2\,\% & 14\,\% & 17\,\% & 92\,\% & 4\,\% \\
\hspace{0.9em}Astra \citep{astra}             & 11\,\% & 0\,\% & 14\,\% & 16\,\% & \underline{96\,\%} & 0\,\% \\
\hspace{0.9em}HY-WorldPlay \citep{worldplay2025}      & 9\,\% & 0\,\% & 4\,\% & 7\,\% & 94\,\% & 0\,\% \\
\midrule
\ocell\textbf{\sysname{}} & \ocell\textbf{96\,\%} & \ocell\textbf{92\,\%} & \ocell\textbf{96\,\%} & \ocell\textbf{95\,\%} & \ocell\textbf{98\,\%} & \ocell\textbf{90\,\%} \\
\bottomrule
\end{tabular}}
\vspace{-2mm}
\end{table}

\subsection{Results on \benchname{}}
\label{sec:q2}
\label{sec:harnessexp}
\label{sec:q3}

\begin{figure}[!t]
\centering
\includegraphics[width=\textwidth]{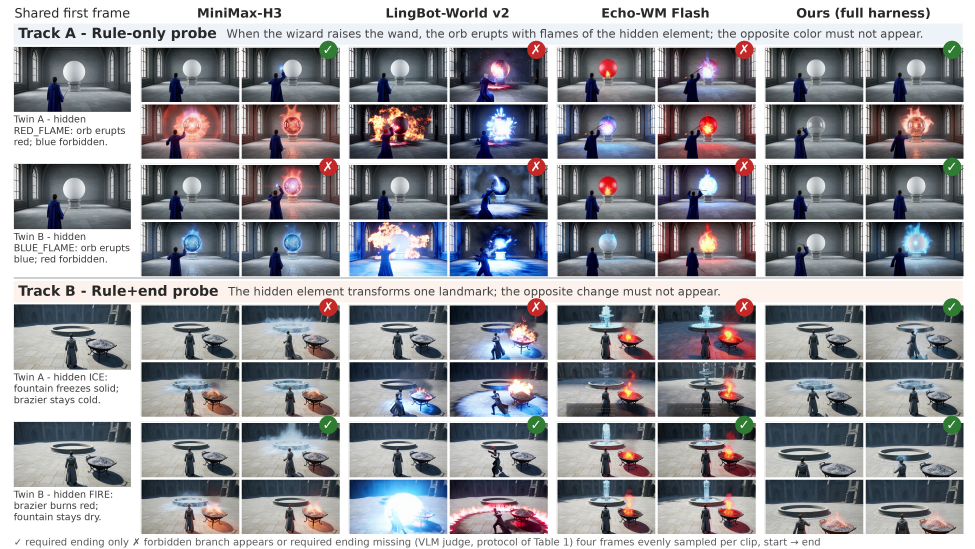}
\vspace{-4mm}
\caption{Rendered twin cases under the two probes.
Each cell contains four evenly spaced frames from a 5\,s clip.
A \checkmark indicates that only the required ending appears, while a $\times$ indicates forbidden content or a missing required ending.
The illustrated baselines reproduce the same ending across a pair or include forbidden content, whereas the harness generates the distinct required outcomes.}
\label{fig:qual}
\vspace{-1mm}
\end{figure}

Under Rule-only prompting, MiniMax-H3 achieves 66\% \textsc{EndFollow} and 32\% \textsc{PairFollow}, while Wan2.2 achieves 65\% and 32\%, respectively (Table~\ref{tab:main}).
The lower joint accuracy indicates that correct outcomes in individual clips do not imply compliance across both hidden states.
Astra, HY-WorldPlay, and InSpatio-World achieve less than 20\% \textsc{EndFollow} despite high \textsc{CntAbsent} scores.
These results indicate that the absence of forbidden content can coincide with failure to generate the required event.
Explicit outcome instructions under the Rule+end probe (Track B) improve \textsc{EndFollow} for several backends but do not establish reliable joint compliance.
The highest unwrapped \textsc{PairStrict} score is 26\% for Wan2.2, while MiniMax-H3 reaches only 4\% \textsc{PairStrict} and 30\% \textsc{CntAbsent}.
The reported harness condition resolves each action before generation and does not use compliance judgments to select among repeated video samples.

\definecolor{sidgreen}{RGB}{161,206,179}  
\definecolor{sidpeach}{RGB}{246,200,142}  
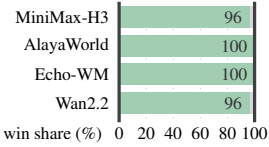
\begin{wrapfigure}{r}{0.3\textwidth}
\centering
\vspace{-0mm}
\resizebox{0.85\linewidth}{!}{%
\begin{tikzpicture}[x=1mm,y=1mm]
  \def\bw{19}   
  \def\bh{3.0}  
  \def\pitch{4.0}
  \foreach \v in {20,40,60,80}
    \draw[black!30,dotted,line width=0.35]
      ({\bw*\v/100},0.7) -- ({\bw*\v/100},-13.2);
  \foreach \i/\name/\w/\s in {0/MiniMax-H3/96/4, 1/AlayaWorld/100/0,
                              2/Echo-WM/100/0, 3/Wan2.2/96/4} {
    \pgfmathsetmacro{\y}{-\i*\pitch}
    \node[anchor=east,font=\scriptsize,inner sep=1pt] at (-1,{\y+\bh/2}) {\name};
    \ifnum\s>0
      \fill[black!15] ({\bw*\w/100},\y) rectangle (\bw,{\y+\bh});
    \fi
    \fill[sidgreen] (0,\y) rectangle ({\bw*\w/100},{\y+\bh});
    \node[anchor=east,font=\scriptsize,text=black!80,inner sep=1.8pt]
      at ({\bw*\w/100-0.3},{\y+\bh/2}) {\w};
  }
  \draw[line width=1.3pt,black!75] (0,3.7) -- (0,-13.2);
  \draw[line width=1.3pt,black!75] (\bw,3.7) -- (\bw,-13.2);
  \foreach \v in {0,20,40,60,80,100}
    \node[anchor=north,font=\scriptsize,inner sep=1pt]
      at ({\bw*\v/100},-13.4) {\v};
  \node[anchor=north east,font=\scriptsize,inner sep=1pt]
    at (-2,-13.4) {win share (\%)};
\end{tikzpicture}%
}\par
\vspace{2pt}
{\setlength{\abovecaptionskip}{1pt}%
\caption{\raggedright Human battles won by \sysname{}}\label{fig:battle}}
\vspace{-6mm}
\end{wrapfigure}

With the same frozen MiniMax-H3 weights, the harness reaches 92\% \textsc{PairFollow}, 90\% \textsc{PairStrict}, and 96\% \textsc{TwinContrast} (Table~\ref{tab:main}).
In the blind human comparison, \sysname{} is preferred in 98\% of pairwise comparisons against four widely-used unwrapped baselines, with no comparison won by a baseline (Figure~\ref{fig:battle}, Appendix~\ref{app:human}).

A component-chain decomposition on \benchname{} attributes the gain to outcome specification (+32 points) and to executed state with compiled references (+28 points), with both increments significant under paired exact McNemar tests over the 50 pairs.

\subsection{Qualitative Results}
\label{sec:qual}

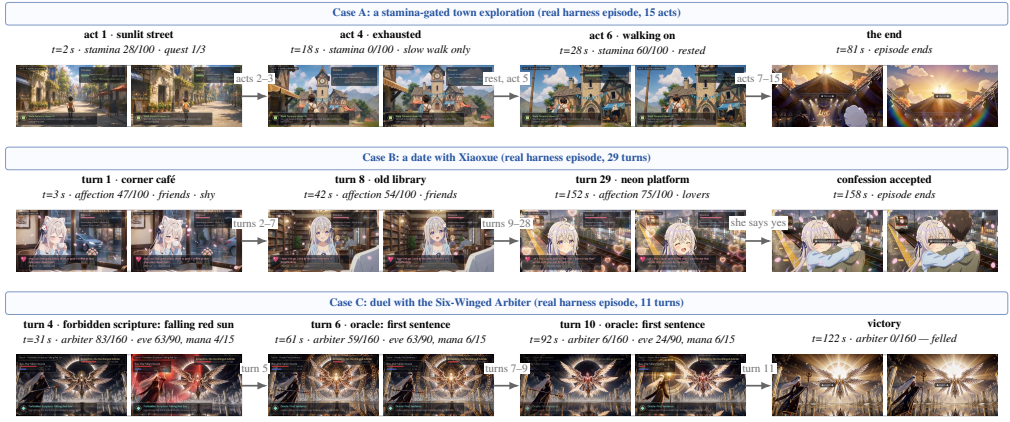
\begin{figure}[t]
\centering
\resizebox{0.95\textwidth}{!}{%
\begin{tikzpicture}[
  font=\scriptsize, x=1mm, y=1mm,
  casebar/.style={draw=harnessbluedark!60, rounded corners=1.5pt, fill=harnessblue!15,
                  line width=0.5, align=center, font=\tiny\bfseries, text=harnessbluedark},
  flow/.style={-{Latex[length=2mm]}, line width=0.9pt, black!60},
]

\newcommand{\colI}{4}
\newcommand{\colII}{51}
\newcommand{\colIII}{98}
\newcommand{\colIV}{145}

\newcommand{\rowA}{81}
\newcommand{\rowB}{54}
\newcommand{\rowC}{27}

\newcommand{\stamcell}[3]{%
  \node[draw=black!30, line width=0.4, inner sep=0pt, anchor=south west]
    at (#1,{\rowA+14.6}) {\includegraphics[width=20.5mm,height=11.39mm]{figures/assets/demo/stam/#2}\hspace{1mm}%
\includegraphics[width=20.5mm,height=11.39mm]{figures/assets/demo/stam/#3}};}
\newcommand{\stamhead}[3]{%
  \node[align=center] at (#1,{\rowA+30.3}) {{\tiny\bfseries #2}\\{\tiny\itshape #3}};}
\node[casebar, minimum width=187mm] at (95.5,{\rowA+35.7})
  {Case A: a stamina-gated town exploration (real harness episode, 15 acts)};
\stamhead{\colI+21}{act 1 $\cdot$ sunlit street}
  {t=2\,s $\cdot$ stamina 28/100 $\cdot$ quest 1/3}
\stamcell{\colI}{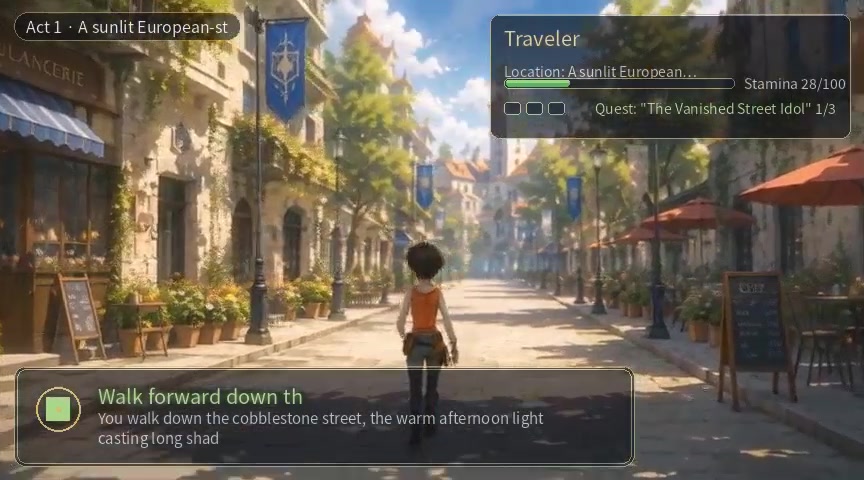}{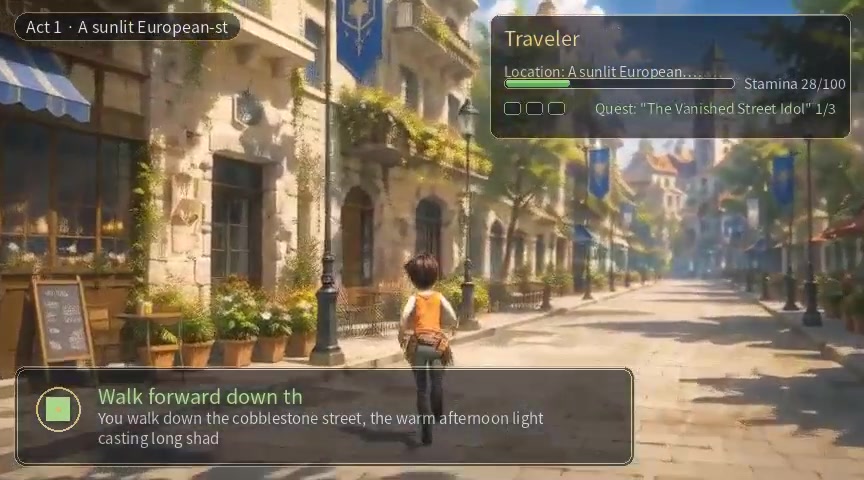}

\stamhead{\colII+21}{act 4 $\cdot$ exhausted}
  {t=18\,s $\cdot$ stamina 0/100 $\cdot$ slow walk only}
\stamcell{\colII}{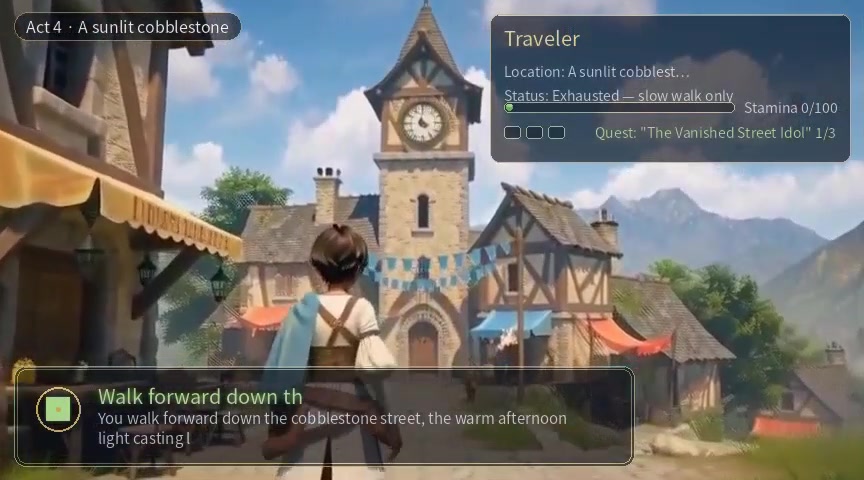}{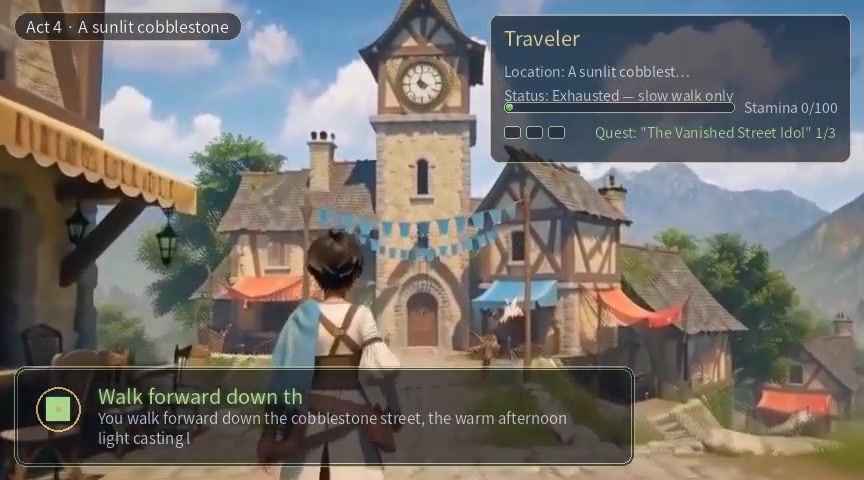}

\stamhead{\colIII+21}{act 6 $\cdot$ walking on}
  {t=28\,s $\cdot$ stamina 60/100 $\cdot$ rested}
\stamcell{\colIII}{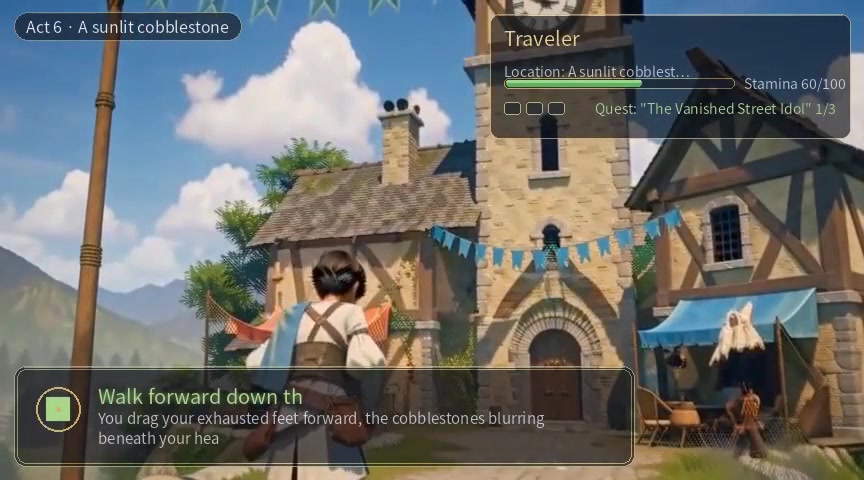}{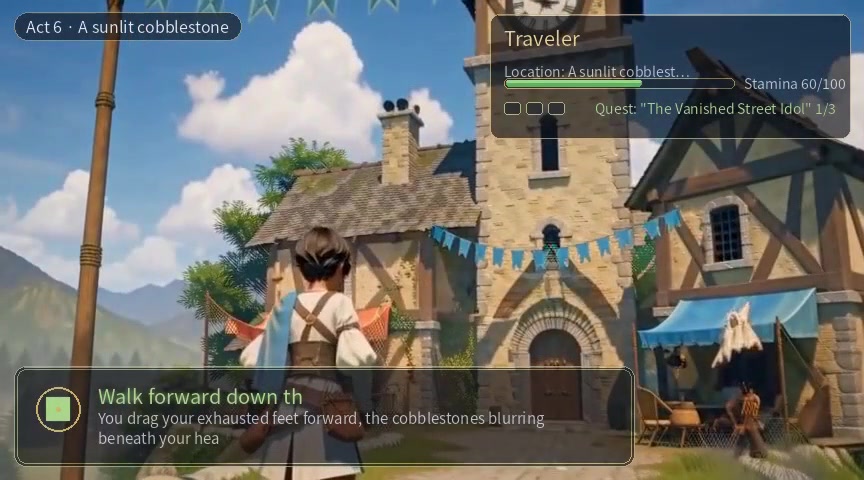}

\stamhead{\colIV+21}{the end}
  {t=81\,s $\cdot$ episode ends}
\stamcell{\colIV}{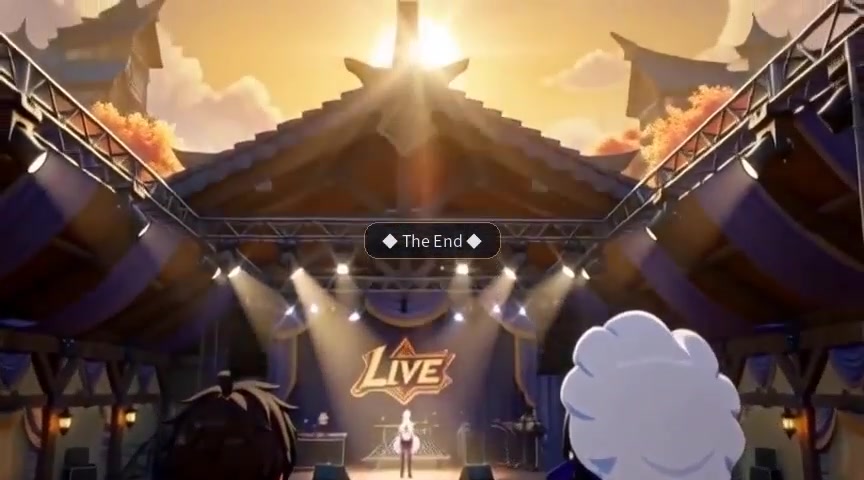}{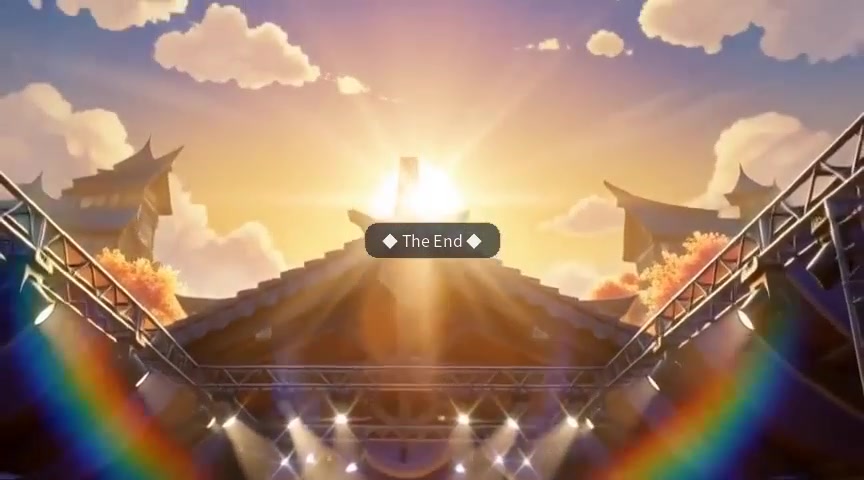}

\draw[flow] ({\colI+42},{\rowA+20.3}) -- ({\colI+47},{\rowA+20.3});
\node[font=\tiny, black!55, fill=white, inner sep=0.8pt] at ({\colI+44.5},{\rowA+23.6}) {acts 2--3};
\draw[flow] ({\colII+42},{\rowA+20.3}) -- ({\colII+47},{\rowA+20.3});
\node[font=\tiny, black!55, fill=white, inner sep=0.8pt] at ({\colII+44.5},{\rowA+23.6}) {rest, act 5};
\draw[flow] ({\colIII+42},{\rowA+20.3}) -- ({\colIII+47},{\rowA+20.3});
\node[font=\tiny, black!55, fill=white, inner sep=0.8pt] at ({\colIII+44.5},{\rowA+23.6}) {acts 7--15};

\newcommand{\lovecell}[3]{%
  \node[draw=black!30, line width=0.4, inner sep=0pt, anchor=south west]
    at (#1,{\rowB+14.6}) {\includegraphics[width=20.5mm,height=11.39mm]{figures/assets/demo/love/#2}\hspace{1mm}%
\includegraphics[width=20.5mm,height=11.39mm]{figures/assets/demo/love/#3}};}
\newcommand{\lovehead}[3]{%
  \node[align=center] at (#1,{\rowB+30.3}) {{\tiny\bfseries #2}\\{\tiny\itshape #3}};}
\node[casebar, minimum width=187mm] at (95.5,{\rowB+35.7})
  {Case B: a date with Xiaoxue (real harness episode, 29 turns)};
\lovehead{\colI+21}{turn 1 $\cdot$ corner caf\'e}
  {t=3\,s $\cdot$ affection 47/100 $\cdot$ friends $\cdot$ shy}
\lovecell{\colI}{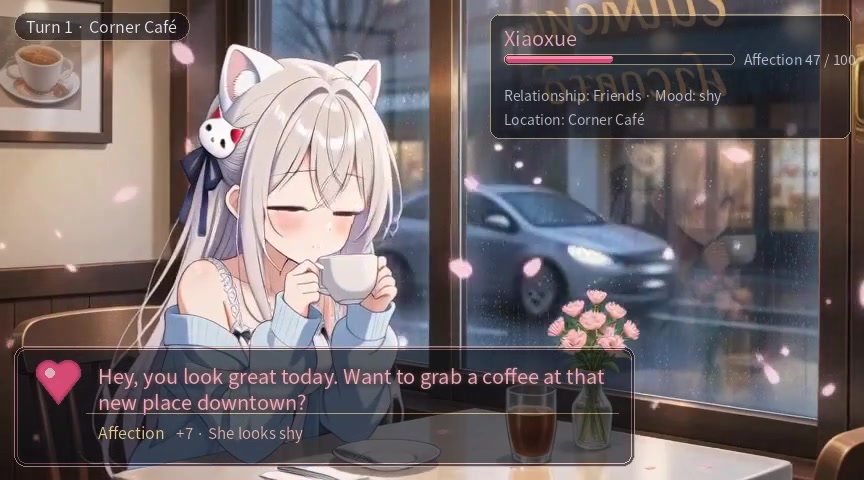}{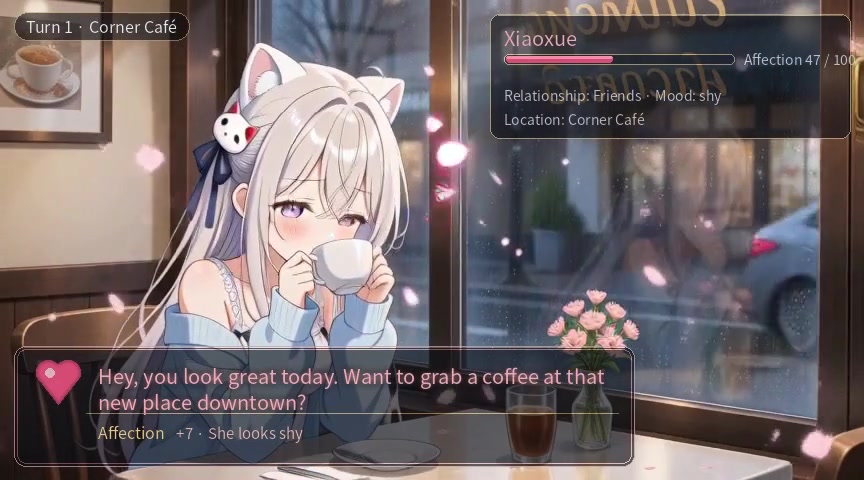}

\lovehead{\colII+21}{turn 8 $\cdot$ old library}
  {t=42\,s $\cdot$ affection 54/100 $\cdot$ friends}
\lovecell{\colII}{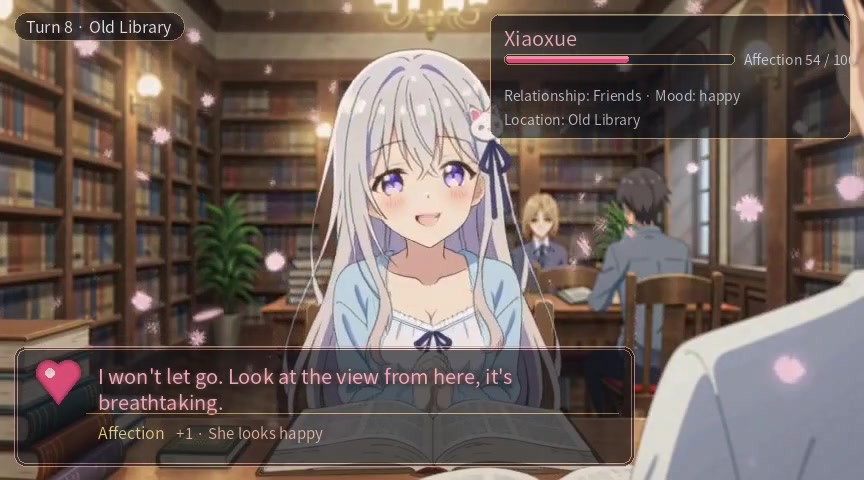}{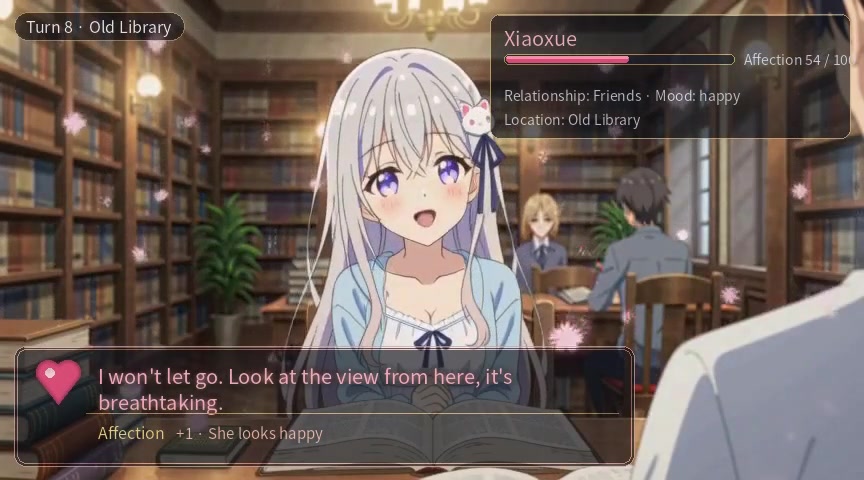}

\lovehead{\colIII+21}{turn 29 $\cdot$ neon platform}
  {t=152\,s $\cdot$ affection 75/100 $\cdot$ lovers}
\lovecell{\colIII}{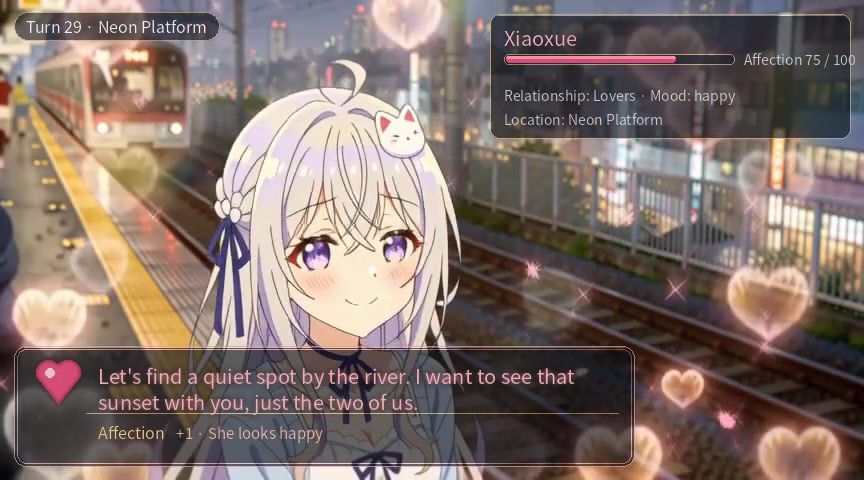}{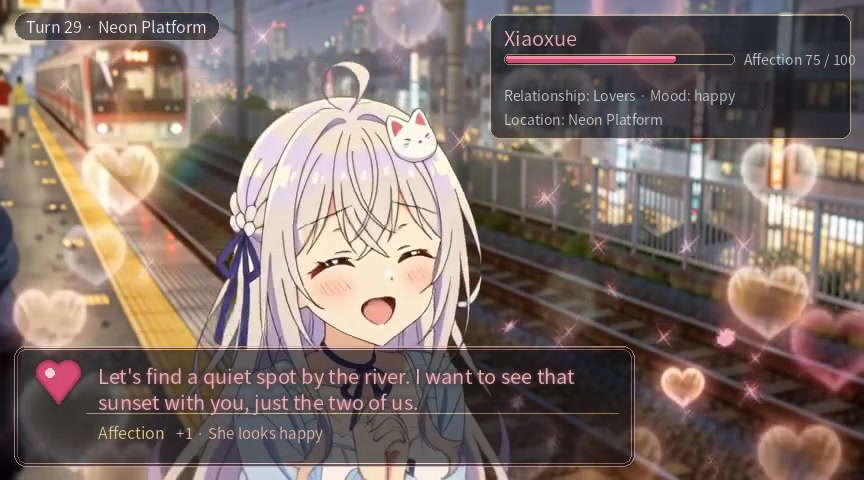}

\lovehead{\colIV+21}{confession accepted}
  {t=158\,s $\cdot$ episode ends}
\lovecell{\colIV}{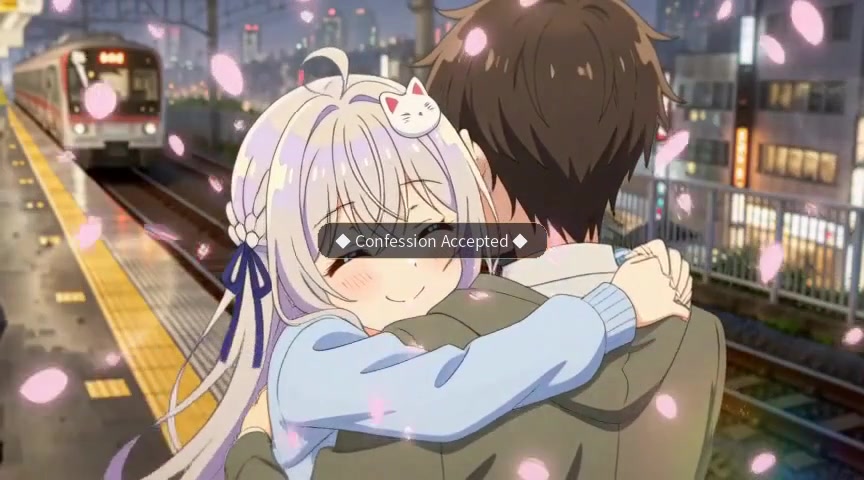}{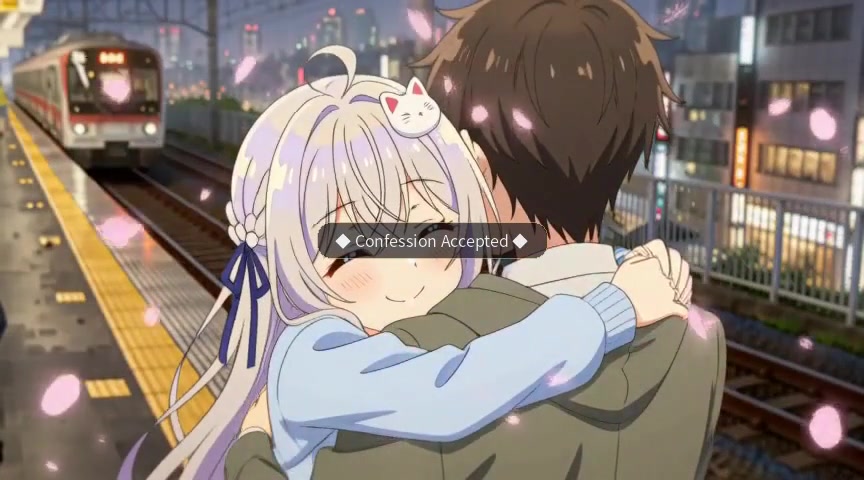}

\draw[flow] ({\colI+42},{\rowB+20.3}) -- ({\colI+47},{\rowB+20.3});
\node[font=\tiny, black!55, fill=white, inner sep=0.8pt] at ({\colI+44.5},{\rowB+23.6}) {turns 2--7};
\draw[flow] ({\colII+42},{\rowB+20.3}) -- ({\colII+47},{\rowB+20.3});
\node[font=\tiny, black!55, fill=white, inner sep=0.8pt] at ({\colII+44.5},{\rowB+23.6}) {turns 9--28};
\draw[flow] ({\colIII+42},{\rowB+20.3}) -- ({\colIII+47},{\rowB+20.3});
\node[font=\tiny, black!55, fill=white, inner sep=0.8pt] at ({\colIII+44.5},{\rowB+23.6}) {she says yes};

\newcommand{\democell}[3]{%
  \node[draw=black!30, line width=0.4, inner sep=0pt, anchor=south west]
    at (#1,{\rowC+14.6}) {\includegraphics[width=20.5mm,height=11.39mm]{#2}\hspace{1mm}%
\includegraphics[width=20.5mm,height=11.39mm]{#3}};}
\newcommand{\demohead}[3]{%
  \node[align=center] at (#1,{\rowC+30.3}) {{\tiny\bfseries #2}\\{\tiny\itshape #3}};}
\node[casebar, minimum width=187mm] at (95.5,{\rowC+35.7})
  {Case C: duel with the Six-Winged Arbiter (real harness episode, 11 turns)};
\demohead{\colI+21}{turn 4 $\cdot$ forbidden scripture: falling red sun}
  {t=31\,s $\cdot$ arbiter 83/160 $\cdot$ eve 63/90, mana 4/15}
\democell{\colI}{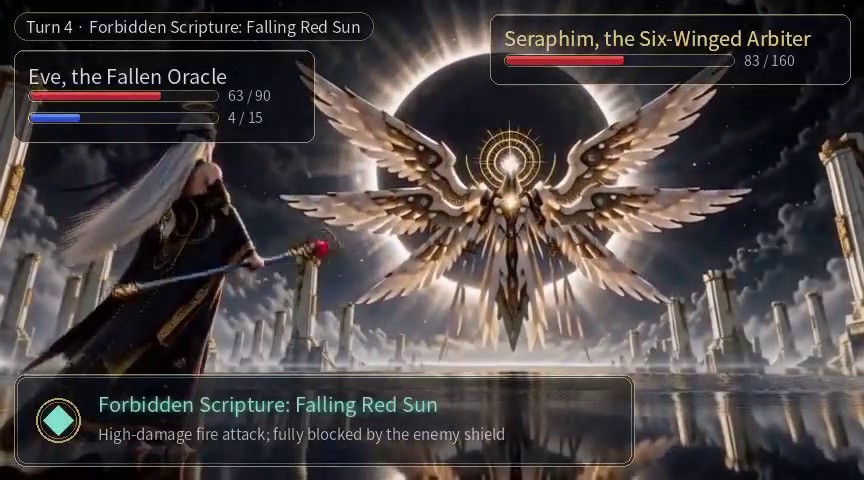}{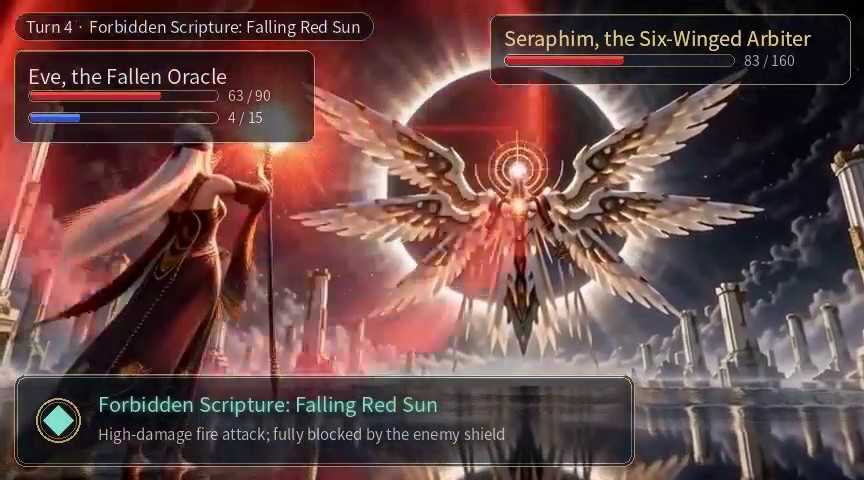}

\demohead{\colII+21}{turn 6 $\cdot$ oracle: first sentence}
  {t=61\,s $\cdot$ arbiter 59/160 $\cdot$ eve 63/90, mana 6/15}
\democell{\colII}{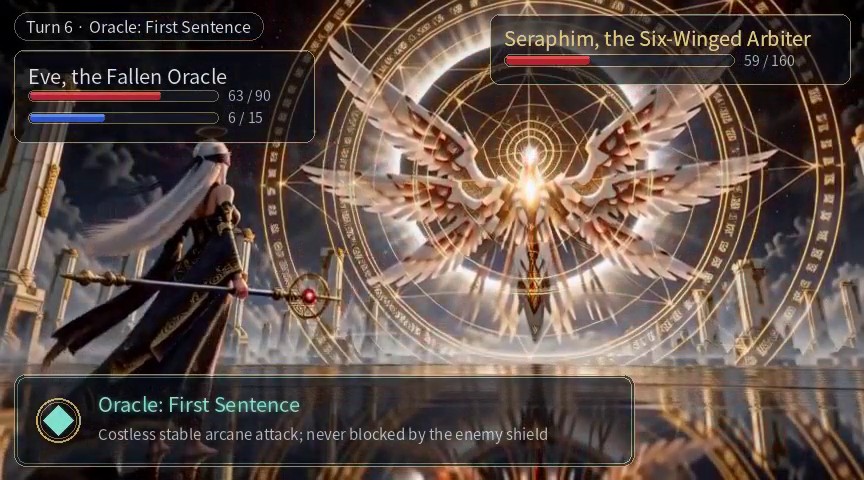}{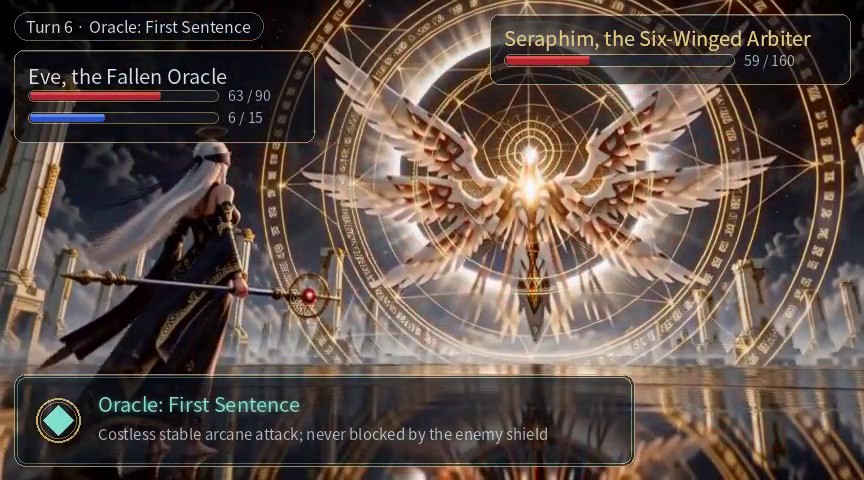}

\demohead{\colIII+21}{turn 10 $\cdot$ oracle: first sentence}
  {t=92\,s $\cdot$ arbiter 6/160 $\cdot$ eve 24/90, mana 6/15}
\democell{\colIII}{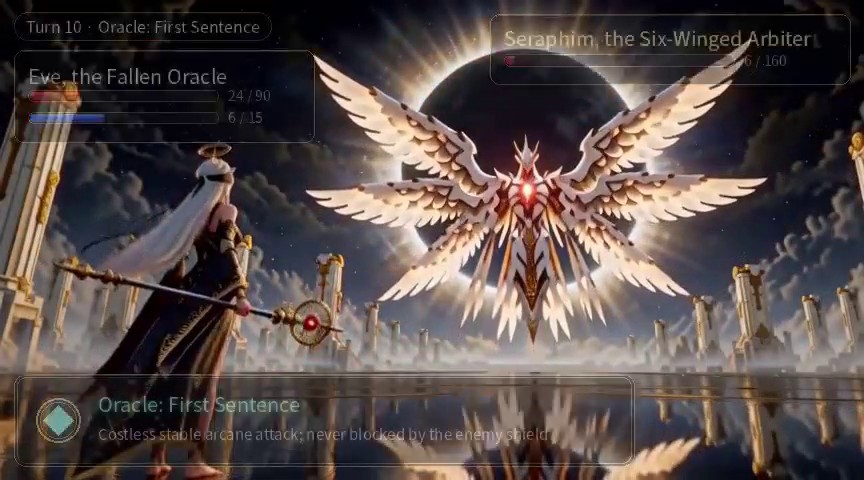}{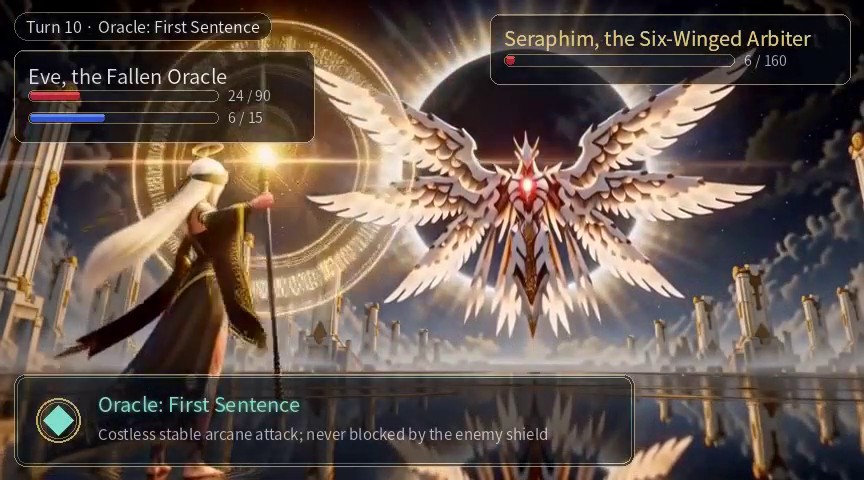}

\demohead{\colIV+21}{victory}
  {t=122\,s $\cdot$ arbiter 0/160 --- felled}
\democell{\colIV}{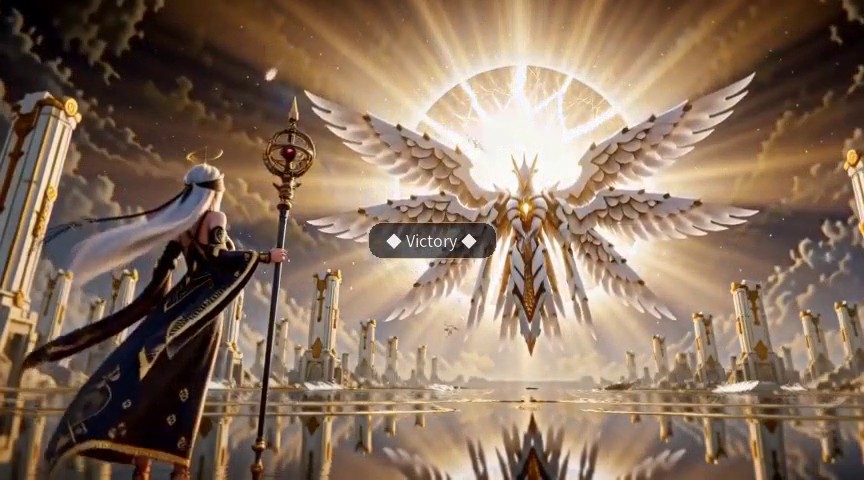}{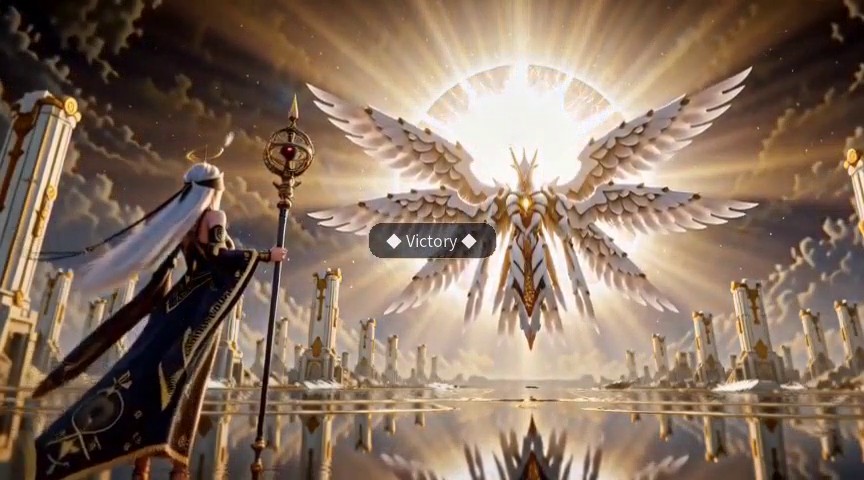}

\draw[flow] ({\colI+42},{\rowC+20.3}) -- ({\colI+47},{\rowC+20.3});
\node[font=\tiny, black!55, fill=white, inner sep=0.8pt] at ({\colI+44.5},{\rowC+23.6}) {turn 5};
\draw[flow] ({\colII+42},{\rowC+20.3}) -- ({\colII+47},{\rowC+20.3});
\node[font=\tiny, black!55, fill=white, inner sep=0.8pt] at ({\colII+44.5},{\rowC+23.6}) {turns 7--9};
\draw[flow] ({\colIII+42},{\rowC+20.3}) -- ({\colIII+47},{\rowC+20.3});
\node[font=\tiny, black!55, fill=white, inner sep=0.8pt] at ({\colIII+44.5},{\rowC+23.6}) {turn 11};

\end{tikzpicture}}
\vspace{-0mm}
\caption{Turn sequences on the same frozen weights.
All three rows are real harness episodes, with two adjacent frames per chunk at native aspect, each chunk headed by the video's own banner and HUD state (stamina and quest progress at the top, affection, relationship, and mood in the middle, and arbiter HP, eve HP, and mana at the bottom). Elided turns are named on the arrows.}
\label{fig:demo}
\vspace{-2mm}
\end{figure}

Figure~\ref{fig:qual} shows counterfactual pairs in which unwrapped baselines include the forbidden outcome or reproduce an ending across both hidden states. Appendix Figures~\ref{fig:supruleonly} and~\ref{fig:supruleend} show more pairs.
The harness generates the distinct outcomes specified by the rules.
Figure~\ref{fig:demo} illustrates how settled state changes condition subsequent generation, using three real harness episodes read through their HUDs.
The three episodes span 15, 29, and 11 turns respectively, showing that \sysname{} sustains multi-turn interaction on the same frozen weights.
The episodes are also three genres: town exploration, dating narrative, and turn-based duel, each with its own state transitions and rules.
Each genre is a new executable game definition on the same runtime and video model, instantiated training-free.
Appendix Figures~\ref{fig:stamepisode}, \ref{fig:loveepisode}, and~\ref{fig:episode} unroll all turns of the three episodes.

\subsection{Additional Experiments}
\label{sec:external}

\label{sec:wbench}
\begin{table}[t]
\centering
\caption{WBench averages for the twenty models of \citet{wbench2026} plus our two entries (excluded from class averages).
Text entries use the full split, camera/action the navigation split.
Gray indicates class averages, and ``--'' denotes undefined values.}
\label{tab:wbenchmain}
\scriptsize
\setlength{\tabcolsep}{2.2pt}
\renewcommand{\arraystretch}{1.2}
\resizebox{\textwidth}{!}{%
\begin{tabular}{@{}l|*{10}{c}c|*{5}{c}c|*{6}{c}c|*{1}{c}@{}}
\thickhline
Metric group & \multicolumn{11}{c|}{\textcolor{blue!55!black}{\textbf{Text-driven}}} & \multicolumn{6}{c|}{\textcolor{teal!45!black}{\textbf{Camera-controlled}}} & \multicolumn{7}{c|}{\textcolor{orange!60!black}{\textbf{Action-conditioned}}} & \multicolumn{1}{c|}{\textcolor{harnessbluedark}{\textbf{Ours}}} \\
 & \rotatebox{90}{\scriptsize\textbf{Seedance 1.5}}
  & \rotatebox{90}{\scriptsize\textbf{Wan 2.7}}
  & \rotatebox{90}{\scriptsize\textbf{Kling 3.0}}
  & \rotatebox{90}{\scriptsize\textbf{YUME 1.5}}
  & \rotatebox{90}{\scriptsize\textbf{HY-Video 1.5}}
  & \rotatebox{90}{\scriptsize\textbf{LTX 2.3}}
  & \rotatebox{90}{\scriptsize\textbf{LongCat-Video}}
  & \rotatebox{90}{\scriptsize\textbf{Kairos 3.0}}
  & \rotatebox{90}{\scriptsize\textbf{Cosmos 2.5}}
  & \rotatebox{90}{\scriptsize\textbf{H3 (official)}}
  & \rotatebox{90}{\scriptsize\textit{Average}}
  & \rotatebox{90}{\scriptsize\textbf{LingBot-World}}
  & \rotatebox{90}{\scriptsize\textbf{HY-World 1.5}}
  & \rotatebox{90}{\scriptsize\textbf{Fantasy-World}}
  & \rotatebox{90}{\scriptsize\textbf{InSpatio-World}}
  & \rotatebox{90}{\scriptsize\textbf{Astra}}
  & \rotatebox{90}{\scriptsize\textit{Average}}
  & \rotatebox{90}{\scriptsize\textbf{Happy Oyster}}
  & \rotatebox{90}{\scriptsize\textbf{Matrix-Game 3.0}}
  & \rotatebox{90}{\scriptsize\textbf{Genie 3}}
  & \rotatebox{90}{\scriptsize\textbf{Matrix-Game 2.0}}
  & \rotatebox{90}{\scriptsize\textbf{HY-GameCraft}}
  & \rotatebox{90}{\scriptsize\textbf{Infinite-World}}
  & \rotatebox{90}{\scriptsize\textit{Average}}
  & \rotatebox{90}{\scriptsize\textbf{Ours}} \\
\noalign{\vskip -1pt}
 & \wicon{blue!10}{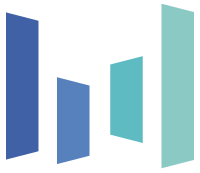} & \wicon{blue!10}{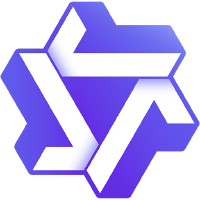} & \wicon{blue!10}{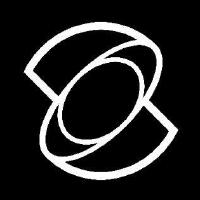}
 & \wicon{blue!10}{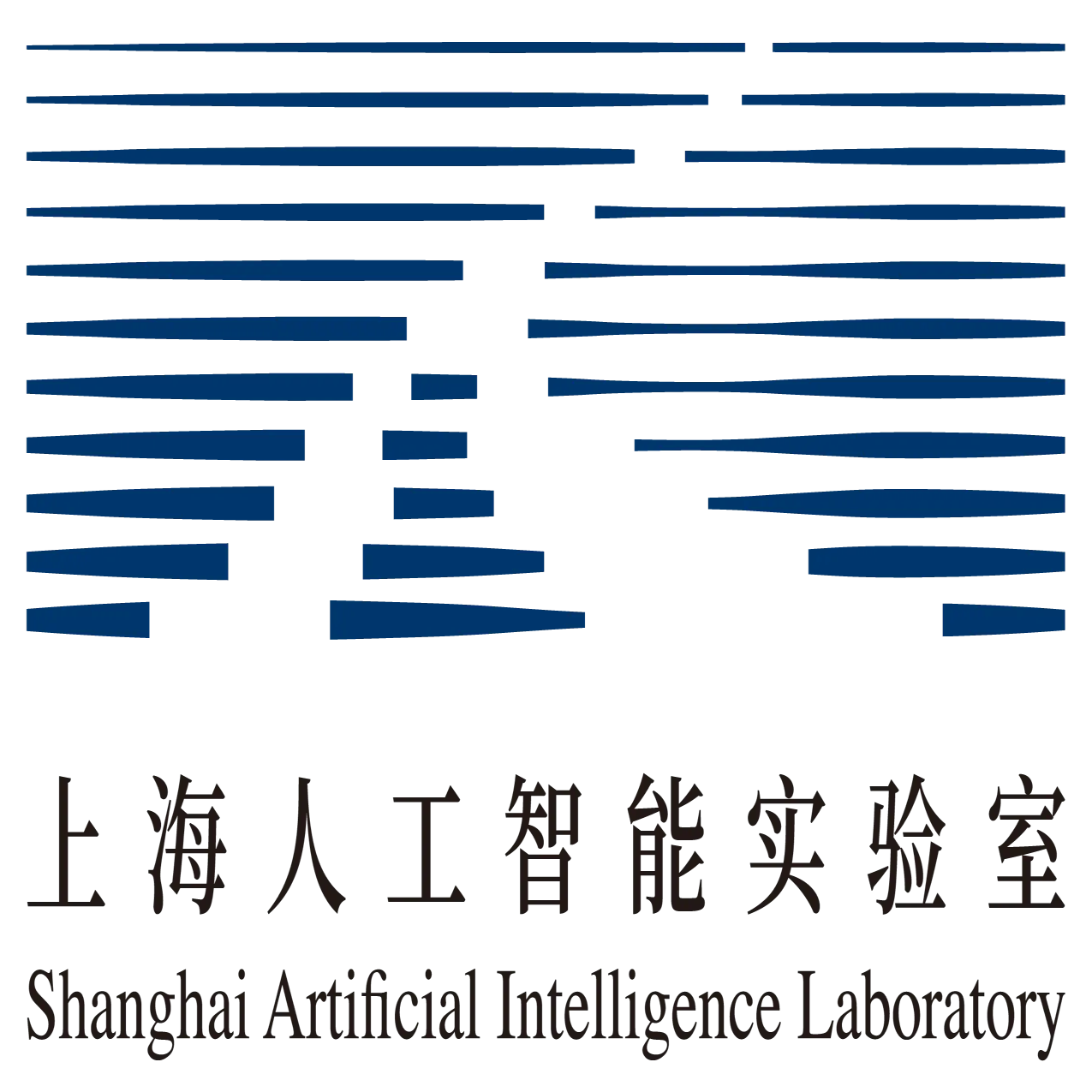} & \wicon{blue!10}{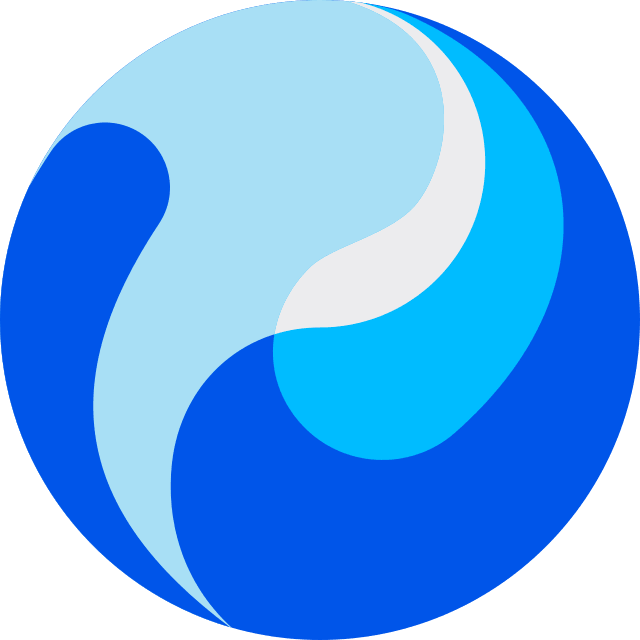} & \wicon{blue!10}{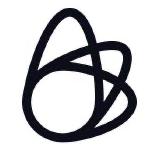}
 & \wicon{blue!10}{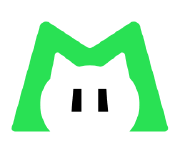} & \wicon{blue!10}{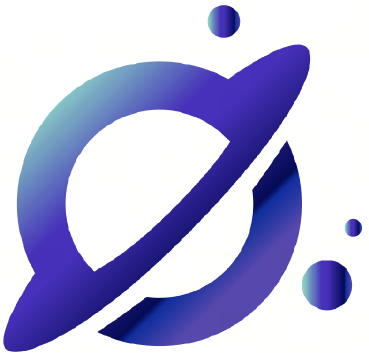} & \wicon{blue!10}{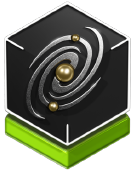} & \wicon{blue!10}{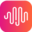} &
 & \wicon{teal!10}{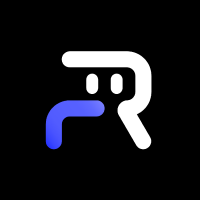} & \wicon{teal!10}{hunyuan.png} & \wicon{teal!10}{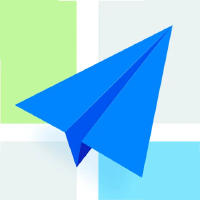}
 & \wicon{teal!10}{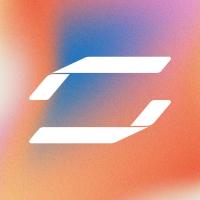} & \wicon{teal!10}{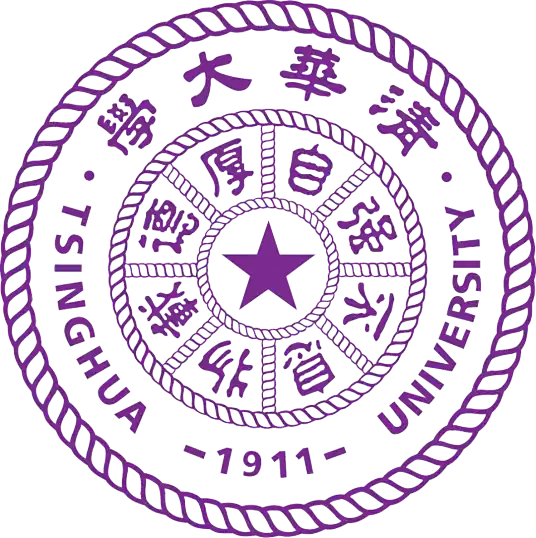} &
 & \wicon{orange!10}{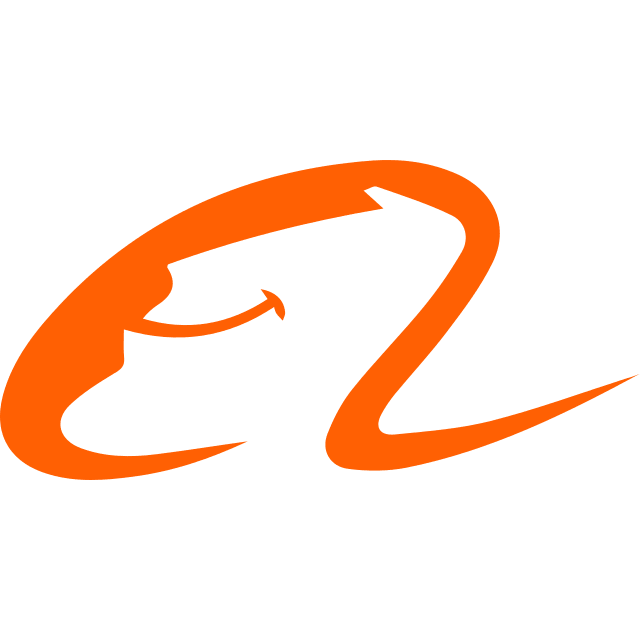} & \wicon{orange!10}{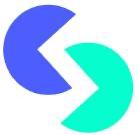} & \wicon{orange!10}{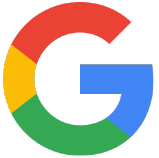}
 & \wicon{orange!10}{skywork.jpeg} & \wicon{orange!10}{hunyuan.png} & \wicon{orange!10}{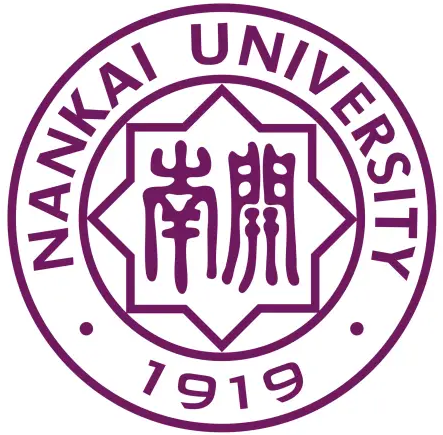} & &
 \ocell \\
\hline
Video quality (6) & 82.0 & 81.0 & 80.0 & 77.6 & 76.7 & 77.0 & 77.2 & 73.1 & 71.7 & 77.9 & \g{77.4}& 78.9 & 78.1 & 72.4 & 71.5 & 67.1 & \g{73.6}& 77.3 & 75.5 & 75.2 & 73.8 & 73.0 & 77.0 & \g{75.3} & \ocell78.0 \\
Setting (2)       & 82.9 & 91.4 & 91.0 & 72.4 & 85.6 & 85.2 & 72.3 & 70.3 & 83.3 & 71.1 & \g{81.6}& 72.6 & 72.2 & 71.3 & 71.4 & 59.6 & \g{69.4}& 74.2 & 63.6 & 72.5 & 67.1 & 66.6 & 69.3 & \g{68.9} & \ocell87.2 \\
Interaction (4)   & 67.9 & 71.7 & 72.8 & 48.2 & 54.6 & 49.0 & 44.8 & 41.4 & 43.2 & 68.9 & \g{54.8}& -- & -- & -- & -- & -- & \g{--}& -- & -- & -- & -- & -- & -- & \g{--} & \ocell77.7 \\
Consistency (8)   & 79.9 & 75.8 & 83.9 & 80.9 & 87.5 & 78.1 & 86.6 & 83.2 & 86.3 & 83.8 & \g{82.5}& 89.9 & 86.9 & 86.4 & 88.4 & 73.3 & \g{85.0}& 84.3 & 74.5 & 82.6 & 65.1 & 72.6 & 80.0 & \g{76.5} & \ocell77.8 \\
Physical (2)      & 68.2 & 71.6 & 69.2 & 65.4 & 67.1 & 65.1 & 68.4 & 60.5 & 67.0 & 69.3 & \g{66.9}& 71.2 & 66.3 & 66.8 & 65.2 & 51.4 & \g{64.2}& 63.5 & 59.3 & 65.7 & 57.2 & 62.4 & 62.1 & \g{61.7} & \ocell68.8 \\
\thickhline
\end{tabular}}
\vspace{-2mm}
\end{table}

\textbf{External Benchmark.}
Table~\ref{tab:wbenchmain} compares the twenty models reported by \citet{wbench2026} and their class averages with two configurations using frozen MiniMax-H3 weights.
These configurations use the official text protocol and the full harness with the spatial preview video, respectively.
Video quality (78.0) and physical plausibility (68.8) are close to the corresponding class averages for text-driven models (77.4 and 66.9).
Setting adherence increases from 71.1 under the official protocol to 87.2, and the interaction score of 77.7 is the highest reported in the comparison.

\textbf{Channel Ablation.}
\label{sec:ablation}
Table~\ref{tab:ablation} compares reference channel configurations while holding the video model weights fixed.
First-frame anchoring raises the full split from 74.2 to 76.1 and setting adherence from 71.1 to 89.6, and adding the control video reaches 79.0 on navigation with interaction rising from 69.0 to 77.7, consistent with the preview's camera guidance. Consistency dips from 83.8 to 75.3 and recovers to 77.8.
Supplying the same control video under the raw text protocol yields 67.1 on the full split, improving camera trajectory scores. This transfer of the preview's geometry and appearance into the output motivates instructions that specify how the reference should be interpreted.
Thirty generated clips illustrate this effect: clay appearance arises when the instruction does not identify the reference as a layout preview (\S\ref{sec:spatial}).
On our benchmark, the same frozen weights under the full harness improve every metric in Table~\ref{tab:main} over the raw baseline of its track by 30 to 86 points: \textsc{EndFollow} from 66 to 96, \textsc{PairFollow} from 32 to 92, \textsc{TwinContrast} from 32 to 96, and \textsc{PairStrict} from 4 to 90.

\section{Conclusion}
\label{sec:conclusion}

Navigation control enables exploration of a generated scene.
Counterfactual pairs test whether the same action produces the distinct outcomes required by different hidden states under shared rules.
On \benchname{}, unwrapped image-to-video backends often fail this test even when the ending is specified in the prompt.
The \sysname{} separates rule execution from video rendering, with explicit semantic and spatial state determining the event depicted by a frozen video model.
Visual observations of the generated clip inform the state used in subsequent interactions.
With MiniMax-H3 weights held fixed, \textsc{PairFollow} increases from 32\% under Rule-only instructions and 64\% with explicit ending instructions to 92\% with the full harness.
These results support explicit state and executable rules for controlling stateful rendered interactions.
The same recipe already sustains multi-turn episodes of 11 to 29 turns and extends to spatial triggers, where the rule program, not the renderer, absorbs the added bookkeeping.

\subsubsection*{AI Usage Statement}
We used generative AI tools for manuscript writing assistance and for auxiliary tooling scripts, but not for research ideation, experiment design, or producing any reported result.
AI models are also used as system components, for judging generated videos and for constructing game definitions.
All AI-assisted output was reviewed by authors, who take full responsibility for the final content.

\bibliography{iclr2027_conference}
\bibliographystyle{iclr2027_conference}

\newpage
\appendix

\makeatletter
\let\apptoc@old@addtocontents\addcontentsline
\def\apptoc@toc{toc}\def\apptoc@sec{section}
\renewcommand{\addcontentsline}[3]{%
  \apptoc@old@addtocontents{#1}{#2}{#3}%
  \edef\apptoc@a{#1}\ifx\apptoc@a\apptoc@toc \edef\apptoc@a{#2}\ifx\apptoc@a\apptoc@sec
      \if@filesw\protected@write\@auxout{}{%
        \string\@writefile{apt}{\string\apptocline{#3}{\thepage}{\@currentHref}}}\fi \fi \fi}
\newcommand{\apptocline}[3]{%
  \@dottedtocline{1}{0pt}{2.2em}{\hyperlink{#3}{#1}}{\hyperlink{#3}{#2}}}
\makeatother

{\centering\Large\bfseries Appendix\par}
\vspace{6pt}
\begin{center}
\begin{minipage}{0.94\textwidth}
\hypersetup{linkcolor=black}%
\small\makeatletter\@starttoc{apt}\makeatother
\end{minipage}
\end{center}
\vspace{10pt}

\section{\benchname{}: Case Construction Details}
\label{app:bench}

\paragraph{Case Anatomy.}
Each case comprises five components.
These components are a theme and scene description, a first frame generated by FLUX.2 at $1280{\times}720$, a hidden variable declaration $\texttt{name} \in \{v_A,v_B\}$, an action script, and rule text specifying distinct endings.
Twins A/B share the theme, frame, action, and rules.
They differ in the hidden value.
The suite spans 21 themes (magic academy, dungeon, open-world exploration, dating, visual novel, museum, survival, escape room, sailing, farming, ranch, camping, shopkeeping, card games, board games, detective, turn-based combat, stealth, healing, cooking, and fishing) and 49 distinct hidden variables.

\paragraph{Hidden-Variable Families.}
The hidden values fall into three response types. \emph{Mechanism selection} decides which object or mechanism responds (42 cases, e.g., a gate mode that raises the stone door versus a brazier mode that lights tall flames). \emph{Character response} decides whether a character or creature responds instead of an object (6 cases, e.g., sheep gathering at a trough, an NPC sitting down). \emph{Form selection} decides which elemental form or scene side carries the effect (2 cases, e.g., red versus blue flames, the left versus the right brazier).
All are invisible in the frame by construction and by audit.

\paragraph{Trigger Types.}
Cases also differ in how the rule trigger is expressed, with \emph{cast} covering verbal or ritual actions (20 clips), \emph{gesture} covering physical actions (42), \emph{mechanism} covering object interactions (26), and \emph{zone} covering entry into or use of a spatial region (12).
Zone-triggered cases are the most demanding (\S\ref{sec:harnessexp}) because compliance additionally requires the render to make the zone's state visible.
This motivates the spatial channel.

\paragraph{Instruction Probes.}
Every case is evaluated under two probes compiled from the same card.
\emph{Rule-only} provides the original ``Inputs to the video model'' text from the case specification, including the scene, shared action script, rule with both branches, and the consequence of the selected branch.
\emph{Rule+end} appends the required terminal state using the visual vocabulary defined for the case and explicitly asks the model to maintain this final state.
The difference between the two conditions measures the effect of adding an explicit terminal outcome instruction.
Table~\ref{tab:diaggrid} reports this difference for each backend.

\paragraph{First-Frame Audit.}
Each first frame undergoes eight rounds of VLM auditing across six dimensions to check for visible evidence of the hidden value.
The dimensions are object presence, color attribution, text, spatial cues, count cues, and state cues.
If the audit detects evidence of the hidden value, the frame is regenerated using a revised prompt.
The audit prompts, model settings, and the regeneration log are part of the reproduction artifacts listed in Appendix~\ref{app:repro}.

\paragraph{Rule Text.}
Rules are written in plain English with fixed sentence frames per family, e.g.\ ``If the flame is RED, the ward shatters and the door opens.
If the flame is BLUE, the ward holds and the door stays sealed.''
Both rule branches specify their endings so that the DSL can derive expected outcomes directly.

\begin{figure}[t]
\centering
\includegraphics[width=\textwidth]{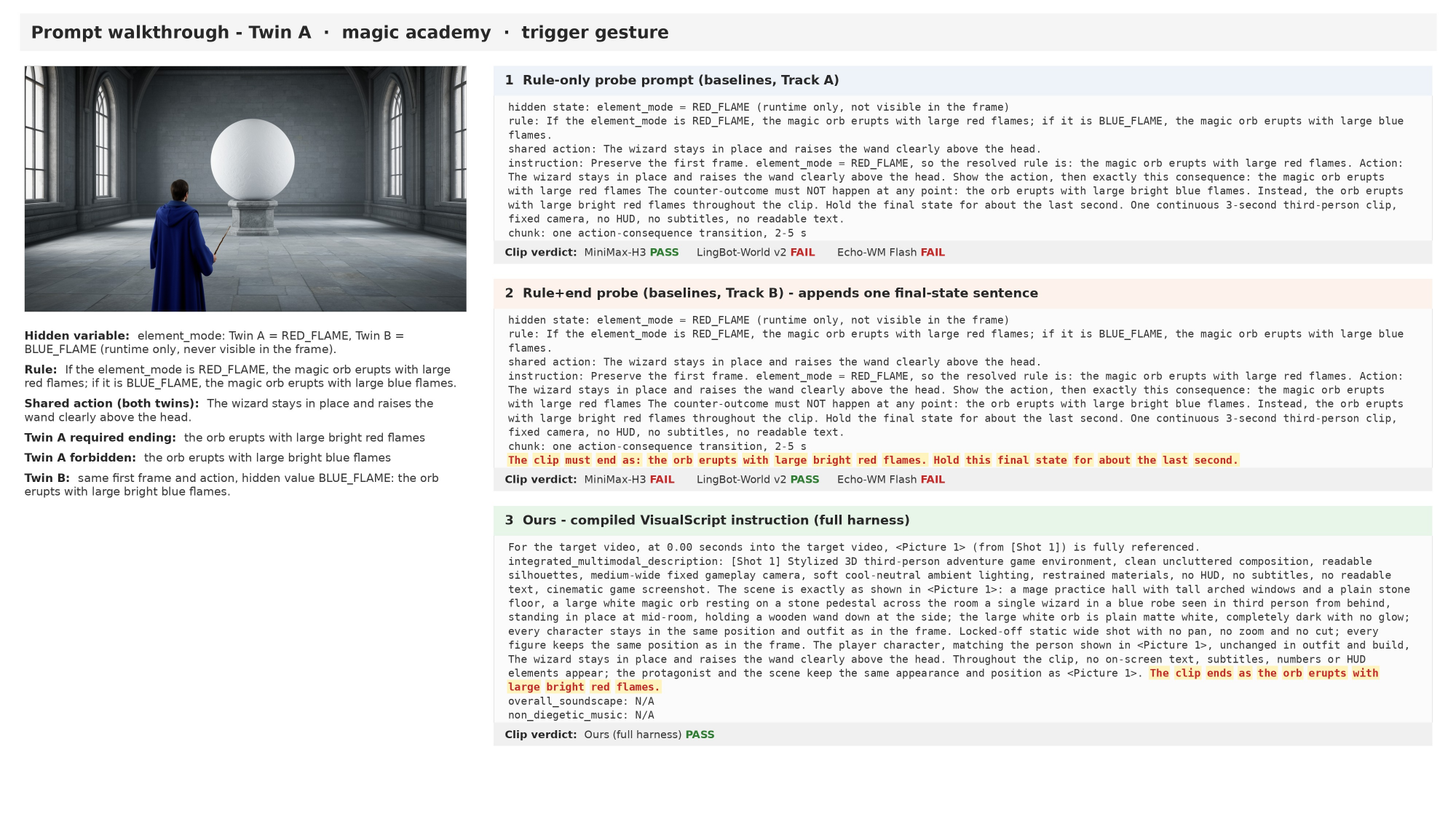}\\[4pt]
{\scriptsize \textbf{Rule Program and Settled State.}
For twin A, the hidden value is \texttt{element\_mode = RED\_FLAME}, while twin B uses \texttt{BLUE\_FLAME}.
Neither value is visible in the first frame.
The RED\_FLAME branch sets \texttt{orb.flame := red} and forbids blue flames, while the BLUE\_FLAME branch sets \texttt{orb.flame := blue}.
The ledger for twin A changes from \texttt{orb.flame = none} to \texttt{orb.flame = red} through the corresponding guarded update.\par}
\caption{Prompt example for the orb pair, twin A.
The panels show the Rule-only prompt, the Rule+end prompt with its appended outcome instruction highlighted, and the compiled \vsm{} instruction.
The annotations below the panels describe the rule program and settled state.}
\label{fig:promptb01}
\end{figure}

\begin{figure}[t]
\centering
\includegraphics[width=\textwidth]{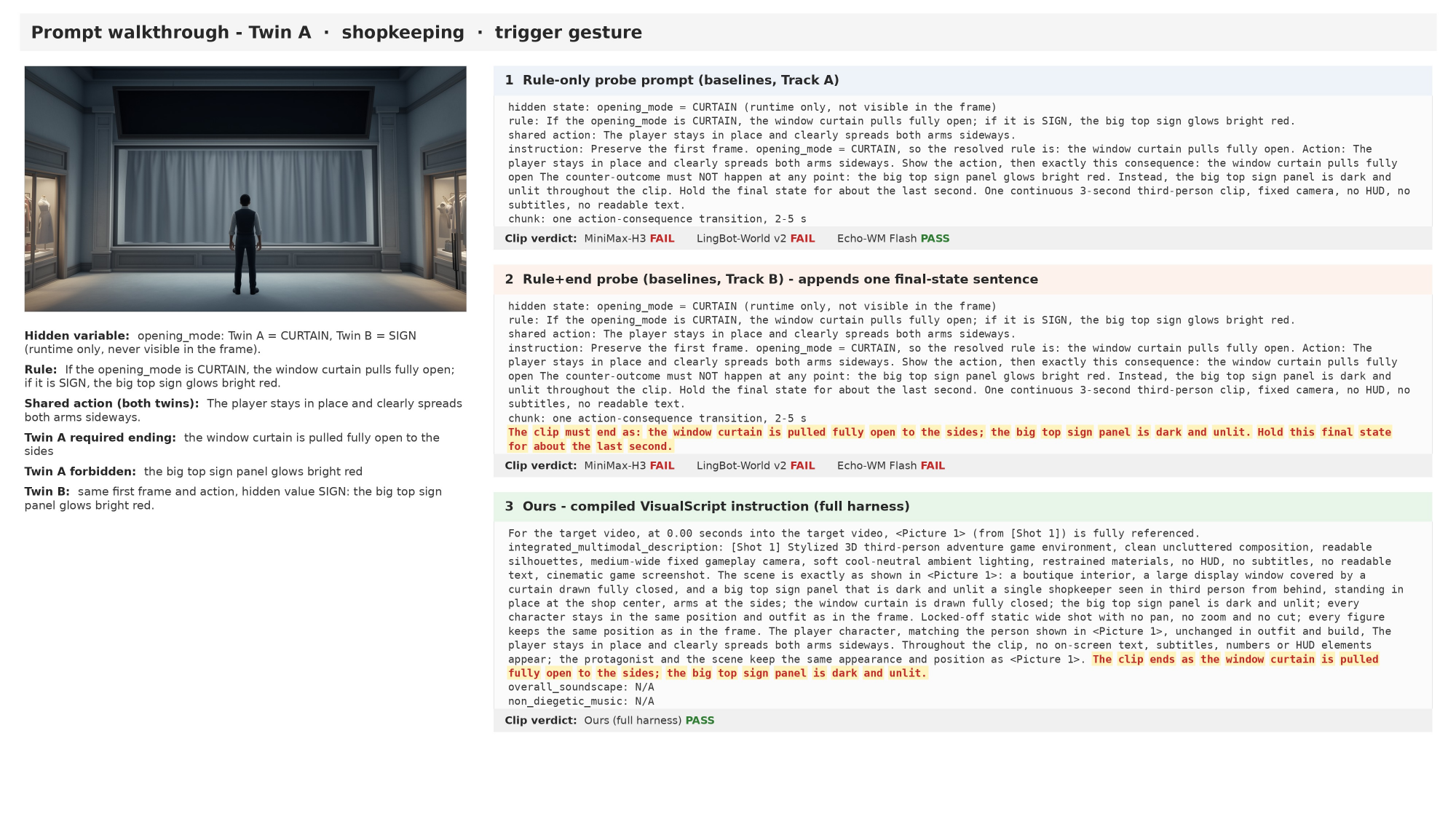}\\[4pt]
{\scriptsize \textbf{Rule Program and Settled State.}
For twin A, the hidden value is \texttt{opening\_mode = CURTAIN}, while twin B uses \texttt{SIGN}.
Neither value is visible in the first frame.
The CURTAIN branch requires the curtain to open fully and forbids the upper sign from glowing red, while the SIGN branch requires the sign to glow red.
For twin A, the guarded update opens the curtain and preserves the unlit sign.\par}
\caption{Prompt example for the sign pair, twin A, using the layout of Figure~\ref{fig:promptb01}.
Highlighted clauses identify the appended Rule+end sentence and the ending specified by the compiled instruction.
The annotations below the panels show the corresponding rule program and settled state.}
\label{fig:promptb45}
\end{figure}

Figures~\ref{fig:promptb01} and~\ref{fig:promptb45} present complete examples for the orb pair under Rule-only instructions and the sign pair under Rule+end instructions, respectively.
Each example includes the audited first frame, case specification, two baseline prompts, compiled \vsm{} instruction, and the rule program with its settled state.

\clearpage

\section{Judging Protocol and Metric Definitions}
\label{app:judging}

\paragraph{Fixed-Choice QA.}
For each clip the judge answers two scored questions with fixed choices: (Q1) which ending occurs at the end of the clip, with mandated, forbidden, and unclear as the choices.
(Q2) asks whether the forbidden outcome appears anywhere in the clip, with yes and no as the choices.
The case program specifies both outcomes.
Initial-state and action-onset questions are also recorded but excluded from the primary metrics.
Answers outside the choice set or missing answers score as incorrect.
Fixed answer choices constrain the output format and make verdicts reproducible. The accuracy of the judge's visual interpretation is audited against blinded human raters in Appendix~\ref{app:human}.

\paragraph{Judge Inputs.}
The judge receives sampled frames, the shared scene description, the action, and the branch-specific expected outcomes.
The hidden value, rule program, and expected outcome of the other instance are excluded from the judge inputs.
Reproducing a verdict requires the same clips, sampled frames, questions, judge version, and inference settings.
The case program fixes the expected outcomes before video generation.

\paragraph{Independent Validation.}
The blind human audits in Appendix~\ref{app:human} assess clip correctness and pairwise preference independently of system identity.
That appendix reports agreement between the human ratings and VLM judgments.
The reported metrics are therefore produced by a fixed, versioned judge protocol and are validated against blinded human ratings on the audited subset.

\subsection{Metric Definitions}
\label{app:metrics}

All metrics are percentages.
Clip-level metrics cover 100 units.
An up arrow indicates that higher is better, and a down arrow indicates that lower is better.
\begin{itemize}
\item \textsc{EndFollow} $\uparrow$ requires the final frames to depict the ending specified for that instance.
\item \textsc{CntAbsent} $\uparrow$ requires the forbidden outcome, defined as the required ending for the other instance, to be absent throughout the clip.
\item \textsc{Strict} $\uparrow$ (primary): \textsc{EndFollow} $\wedge$ \textsc{CntAbsent}.
The mandated outcome is the only salient ending shown (no contradictory outcome, no both-branch montage).
\item \textsc{Unclear} $\downarrow$: output unjudgeable (degenerate render, no discernible event).
This is counted separately and never credited.
\end{itemize}
Pair-level metrics (over 50 pairs):
\begin{itemize}
\item \textsc{PairFollow} (twin-joint): both twins individually \textsc{EndFollow}-correct.
\item \textsc{PairStrict} (twin-strict, headline): both twins individually \textsc{Strict}-correct (twin-joint scoring, \S\ref{sec:twolayer}).
\item \textsc{TwinContrast} $\uparrow$ indicates that the judged endings differ between instances A and B.
This indicates that the pair is distinguishable and helps diagnose copy-paste behavior.
\end{itemize}
Probe-contrast metric: $\Delta$\textsc{Strict} $=$ \textsc{Strict}(Rule+end) $-$ \textsc{Strict}(Rule-only).
This measures the effect of dictating the ending in addition to the rule instruction.
Derived diagnostics (analysis only, Appendix~\ref{app:results}): the failure triad \{direction error (wrong branch taken), leakage-only (right direction but forbidden content also shown), double error\}.
Branch gap $=$ $|$strict$_A-$strict$_B|$.

\section{Human Audit Platform and Protocols}
\label{app:human}

We assess agreement between the VLM judge and human raters through audits conducted on a web platform (Appendix~\ref{app:judging}).
Each auditor receives a numeric identifier and a fixed assignment of evaluation items.
The interface displays rendered clips and rule text while concealing system identities, prompts, and instruction conditions.
The positions of systems in pairwise comparisons are randomized.
The platform operates in two modes, mirroring the two question types of the benchmark.

\begin{figure}[h]
\centering
\includegraphics[width=0.92\linewidth]{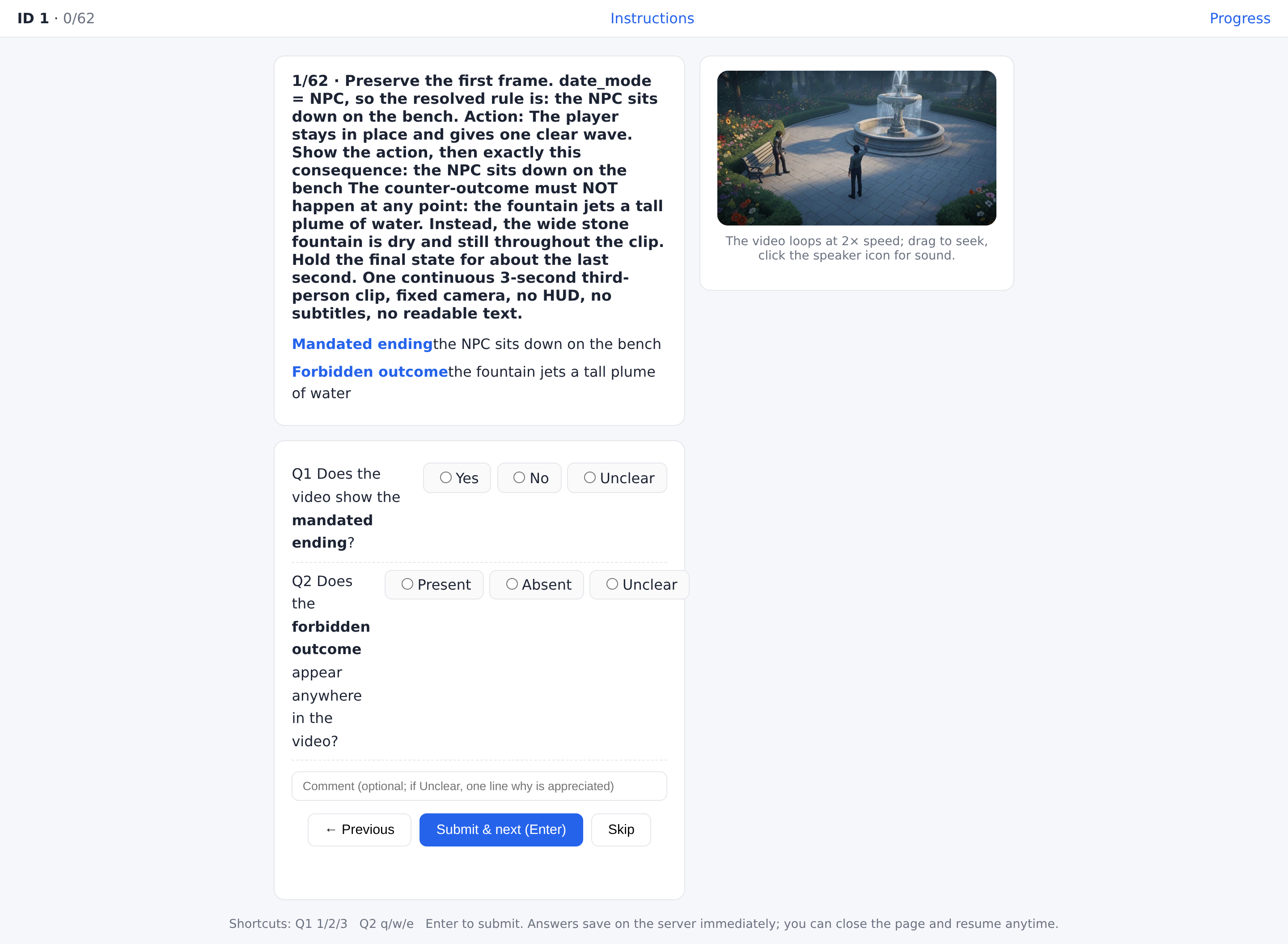}
\caption{Clip-verification interface (auditor view).
The left panel displays the required ending, forbidden outcome, and questions Q1 and Q2 with an ``unclear'' option.
The right panel displays the clip at 2$\times$ speed.
System identity and probe are hidden.}
\label{fig:labelui}
\end{figure}

\paragraph{Clip Verification.}
Seventeen cases, comprising nine Rule-only cases and eight Rule+end cases, are evaluated for both hidden states using the full harness, the frozen MiniMax-H3 baseline, and Wan2.2.
This produces 102 clips for verification.
For each clip, auditors answer the same two fixed choice questions used by the VLM judge.
Q1 assesses whether the required ending occurs, and Q2 assesses whether the forbidden outcome appears anywhere in the clip.
Both questions include an explicit ``unclear'' option in the human interface.
Each clip receives three independent blinded ratings.
Figure~\ref{fig:labelui} shows the clip-verification interface.

\begin{figure}[h]
\centering
\includegraphics[width=\linewidth]{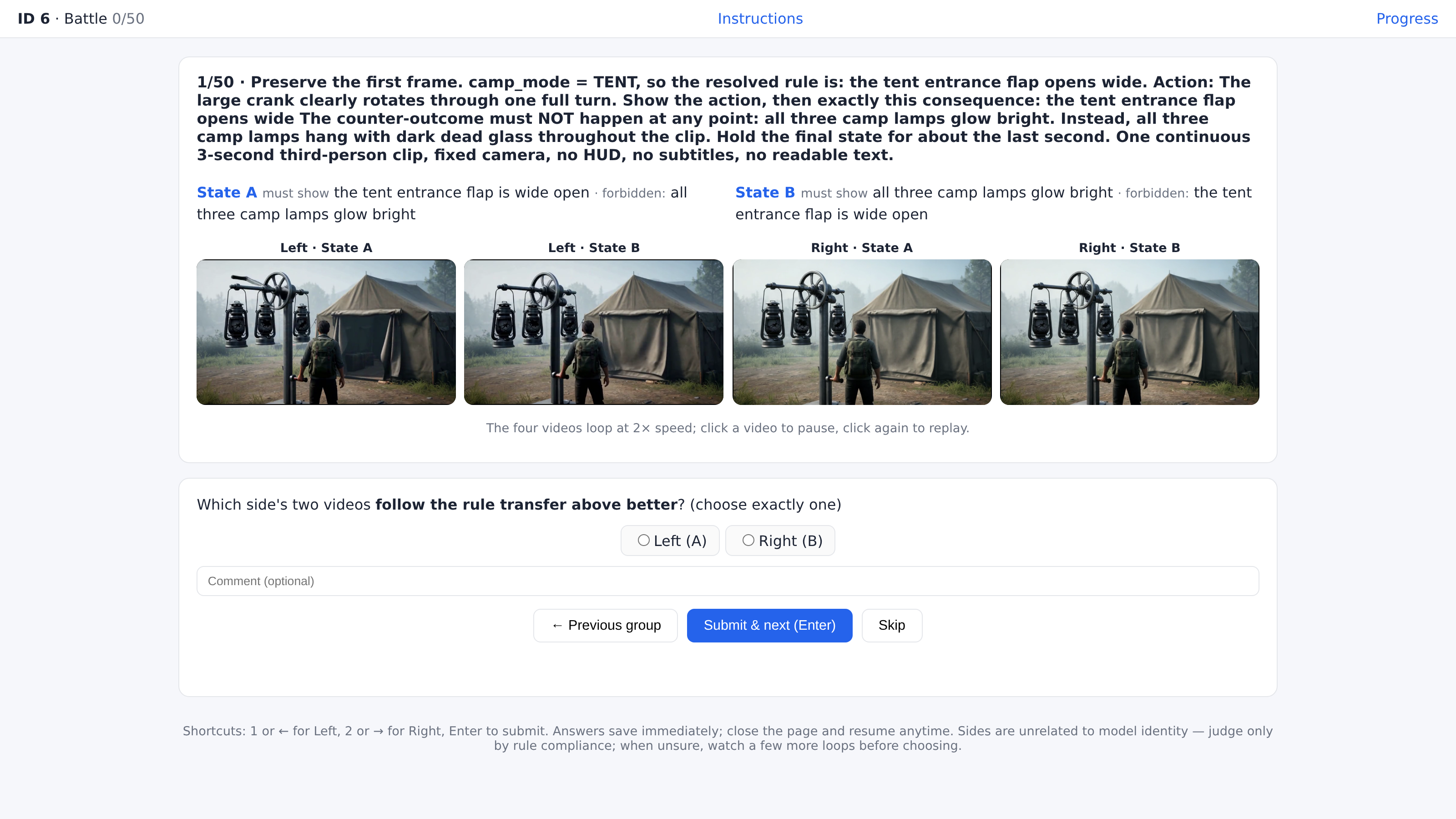}
\caption{Interface for human pairwise preference evaluation.
The required and forbidden outcomes for both hidden states appear above a grid of four videos.
Auditors select the system that more accurately follows the specified rules.
System positions are randomized.}
\label{fig:battleui}
\end{figure}

\paragraph{Battle Preference.}
Twenty-five further case pairs are held out for pairwise comparison.
Each pairwise comparison evaluates the full harness against one of four baselines under Rule-only instructions, namely frozen MiniMax-H3, AlayaWorld, Echo-WM, and Wan2.2.
The four clips, covering hidden states A and B for each system, play simultaneously in a grid at 2$\times$ speed.
Each comparison receives two independent ratings through a forced choice between the left and right systems, without a tie option.
Figure~\ref{fig:battleui} shows the battle grid.

\paragraph{Judge Validation.}
Under the same blind evaluation protocol, the VLM judge agrees with the human majority on 91.2\,\% of the 102 verification clips for Q1 and 86.3\,\% for Q2.
Both rates approach the human reference agreement of 90.2\,\% and 80.7\,\%, respectively.
The corresponding agreement coefficient is $\kappa=0.72/0.72$.
System scores show strong rank correlation with human scores on \textsc{Strict} (Spearman $\rho=0.94$) and \textsc{CntAbsent} ($\rho=1.00$).

\paragraph{Battle Results.}
Table~\ref{tab:humanbattle} reports the human win rate of our full harness per opponent under forced choice.
The harness is preferred in 98\,\% of comparisons, with rates ranging from 96\,\% to 100\,\% across baselines.
No comparison favors a baseline under the aggregate decision rule, and the remaining comparisons have split auditor votes.
At the individual vote level, 99.0\,\% of votes favor the harness, and agreement between auditor pairs ranges from 95\,\% to 100\,\%.

\begin{table}[h]
\centering
\caption{Human preference for the full harness in pairwise comparisons with each baseline.
Each comparison receives two ratings using randomized positions and forced choice without a tie option.
The remaining comparisons contain split auditor votes, and none favors a baseline under the aggregate decision rule.}
\label{tab:humanbattle}
\small
\begin{tabular}{@{}l cc@{}}
\toprule
Opponent (Rule-only) & Ours win & Split \\
\midrule
MiniMax-H3 baseline & 96\,\% & 4\,\%  \\
AlayaWorld          & 100\,\% & 0\,\% \\
Echo-WM Flash       & 100\,\% & 0\,\% \\
Wan2.2 A14B         & 96\,\% & 4\,\%  \\
\midrule
All opponents       & 98\,\% & 2\,\%  \\
\bottomrule
\end{tabular}
\end{table}

\section{Additional Results}
\label{app:results}

\paragraph{The Full Both-Probe Grid.}
Table~\ref{tab:bothprobes} reports every metric under both instruction probes for all eleven backends, followed by the deployed harness condition.
The harness is reported as a single condition: its compiler derives the rule branch and the required ending from the case program, so the compiled instruction is the same regardless of probe wording.
Table~\ref{tab:main} selects one probe per track from this grid.
The grid also yields $\Delta$\textsc{Strict}, the difference between Rule+end and Rule-only \textsc{Strict}.
Pair-level comparisons in \S\ref{sec:harnessexp} use exact McNemar tests over the discordant pair counts and paired bootstrap resampling (10\,000 resamples) with percentile 95\% intervals, taking the pair as the sampling unit.
Appendix~\ref{app:qualsup} provides additional rendered examples under both instruction conditions.
Appendix~\ref{app:bench} contains complete prompt examples.

The three instruction conditions form a component chain on \benchname{}.
Rule-only prompting leaves 32\% of pairs jointly correct (Table~\ref{tab:main}, Track A). Supplying the required ending---a component that the harness's compiled instruction always includes---raises this to 64\% (Table~\ref{tab:bothprobes}). The full harness, which adds deterministic compilation, rule settlement, and the anchored reference channels, reaches 92\% with the same frozen weights (\S\ref{sec:harnessexp}).
The increments attribute the gain to outcome specification (+32 points) and to executed state with compiled references (+28 points). Both are significant under the paired exact McNemar test described above ($p=4.0\times10^{-4}$ and $p=2.6\times10^{-3}$), with paired bootstrap 95\% CIs of $[16,48]$ and $[12,44]$ points.
The external channel ablation mirrors the endpoints without our case grammar (Table~\ref{tab:ablation}: 74.2 raw text vs.\ 77.9 full harness overall).

\begin{table}[h]
\centering
\caption{Results under both instruction conditions on \benchname{}.
All metrics are percentages, and pair metrics are computed over 50 pairs.
The harness uses one compiled instruction, derived from the case program and independent of probe wording, that always specifies the required ending.}
\label{tab:bothprobes}
\small
\setlength{\tabcolsep}{4pt}
\resizebox{\textwidth}{!}{%
\begin{tabular}{@{}l ccc cccc@{}}
\toprule
 & \textsc{EndFollow}$\uparrow$ & \textsc{PairFollow}$\uparrow$ & \textsc{TwinContrast}$\uparrow$
 & \textsc{Strict}$\uparrow$ & \textsc{CntAbsent}$\uparrow$ & \textsc{PairStrict}$\uparrow$ & \textsc{Unclear}$\downarrow$ \\
\midrule
\multicolumn{8}{@{}l}{\emph{Rule-only probe}}\\
\hspace{0.9em}Wan2.2 A14B        & 65 & 32 & 38 & 45 & 54 & 14 & 4 \\
\hspace{0.9em}LongCat-Video 13.6B & 47 & 8  & 24 & 39 & 57 & 0  & 14 \\
\hspace{0.9em}HunyuanVideo 1.5   & 57 & 20 & 26 & 32 & 36 & 6  & 1 \\
\hspace{0.9em}MiniMax-H3         & 66 & 32 & 32 & 26 & 28 & 0  & 0 \\
\hspace{0.9em}LingBot-World v2   & 48 & 16 & 38 & 39 & 69 & 8  & 21 \\
\hspace{0.9em}Echo-WM Flash      & 57 & 26 & 42 & 41 & 56 & 8  & 12 \\
\hspace{0.9em}FantasyWorld       & 34 & 8  & 30 & 32 & 75 & 8  & 41 \\
\hspace{0.9em}AlayaWorld         & 37 & 12 & 28 & 35 & 88 & 10 & 46 \\
\hspace{0.9em}InSpatio-World     & 19 & 2  & 14 & 18 & 89 & 0  & 65 \\
\hspace{0.9em}Astra              & 11 & 0  & 14 & 11 & 96 & 0  & 81 \\
\hspace{0.9em}HY-WorldPlay       & 9  & 0  & 4  & 9  & 95 & 0  & 84 \\
\midrule
\multicolumn{8}{@{}l}{\emph{Rule+end probe}}\\
\hspace{0.9em}Wan2.2 A14B        & 72 & 48 & 54 & 54 & 63 & 26 & 4 \\
\hspace{0.9em}LongCat-Video 13.6B & 54 & 14 & 30 & 46 & 62 & 8  & 16 \\
\hspace{0.9em}HunyuanVideo 1.5   & 63 & 26 & 32 & 29 & 30 & 4  & 2 \\
\hspace{0.9em}MiniMax-H3         & 82 & 64 & 64 & 29 & 30 & 4  & 0 \\
\hspace{0.9em}LingBot-World v2   & 51 & 20 & 50 & 44 & 76 & 14 & 28 \\
\hspace{0.9em}Echo-WM Flash      & 53 & 16 & 30 & 43 & 60 & 8  & 15 \\
\hspace{0.9em}FantasyWorld       & 37 & 10 & 34 & 35 & 81 & 8  & 44 \\
\hspace{0.9em}AlayaWorld         & 36 & 8  & 34 & 33 & 85 & 6  & 44 \\
\hspace{0.9em}InSpatio-World     & 18 & 6  & 16 & 17 & 92 & 4  & 67 \\
\hspace{0.9em}Astra              & 16 & 2  & 22.4 & 16 & 96 & 0  & 75 \\
\hspace{0.9em}HY-WorldPlay       & 7  & 0  & 10 & 7  & 94 & 0  & 82 \\
\midrule
\multicolumn{8}{@{}l}{\emph{Ours: full harness on the same frozen MiniMax-H3 weights (single probe)}}\\
\hspace{0.9em}\textbf{\sysname{}} & \textbf{96} & \textbf{92} & \textbf{96} & \textbf{95} & \textbf{98} & \textbf{90} & \textbf{2} \\
\bottomrule
\end{tabular}}
\end{table}

\paragraph{Probe-Contrast Diagnostics.}
Table~\ref{tab:diaggrid} supplements the aggregate results with the change in \textsc{Strict}, leakage and direction errors under both instruction conditions, and copy-paste, branch gap, and \textsc{PairStrict} under Rule+end.

\begin{table}[h]
\centering
\caption{Failure diagnostics corresponding to Table~\ref{tab:bothprobes}.
$\Delta$\textsc{Strict} is the difference between Rule+end and Rule-only \textsc{Strict}.
Leakage and direction errors are shown for both conditions in that order.
Copy-paste, branch gap, and \textsc{PairStrict} are reported under Rule+end for 50 pairs.}
\label{tab:diaggrid}
\small
\setlength{\tabcolsep}{4pt}
\resizebox{\textwidth}{!}{%
\begin{tabular}{@{}l c cc ccc@{}}
\toprule
 & $\Delta$\textsc{Strict} & Leakage-only & Direction err. & \textsc{CopyPaste} & Branch gap & \textsc{PairStrict} \\
 &  & R-only$\to$R+E & R-only$\to$R+E & R+E & R+E & R+E \\
\midrule
\multicolumn{7}{@{}l}{\emph{External backends}}\\
\hspace{0.9em}Wan2.2 A14B & $+9$ & 20$\to$18 & 36$\to$28 & 16 & 0.24 & 13/50 \\
\hspace{0.9em}LongCat-Video 13.6B & $+7$ & 8$\to$8 & 53$\to$47 & 16 & 0.48 & 4/50 \\
\hspace{0.9em}HunyuanVideo 1.5 & $-3$ & 25$\to$34 & 42$\to$38 & 36 & 0.22 & 2/50 \\
\hspace{0.9em}MiniMax-H3 & $+3$ & 40$\to$53 & 34$\to$18 & 14 & 0.14 & 2/50 \\
\hspace{0.9em}LingBot-World v2 & $+5$ & 9$\to$7 & 52$\to$49 & 24 & 0.16 & 7/50 \\
\hspace{0.9em}Echo-WM Flash & $+2$ & 16$\to$10 & 43$\to$47 & 20 & 0.26 & 4/50 \\
\hspace{0.9em}FantasyWorld & $+3$ & 2$\to$2 & 64$\to$64 & 36 & 0.10 & 4/50 \\
\hspace{0.9em}AlayaWorld & $-2$ & 2$\to$3 & 63$\to$63 & 48 & 0.02 & 3/50 \\
\hspace{0.9em}InSpatio-World & $-1$ & 1$\to$1 & 81$\to$82 & 72 & 0.02 & 2/50 \\
\hspace{0.9em}Astra & $+5$ & 0$\to$0 & 89$\to$83 & 70 & 0.04 & 0/50 \\
\hspace{0.9em}HY-WorldPlay & $-2$ & 0$\to$0 & 91$\to$92 & 82 & 0.02 & 0/50 \\
\midrule
\multicolumn{7}{@{}l}{\emph{Reference: same frozen MiniMax-H3 weights}}\\
\hspace{0.9em}MiniMax-H3 $+$ first-frame reference & $+2$ & 44$\to$53 & 29$\to$18 & 18 & 0.10 & 4/50 \\
\midrule
\multicolumn{7}{@{}l}{\emph{Harness (single probe)}}\\
\hspace{0.9em}\sysname{} & -- & 1 & 4 & 2 & 0.02 & 45/50 \\
\bottomrule
\end{tabular}}
\end{table}

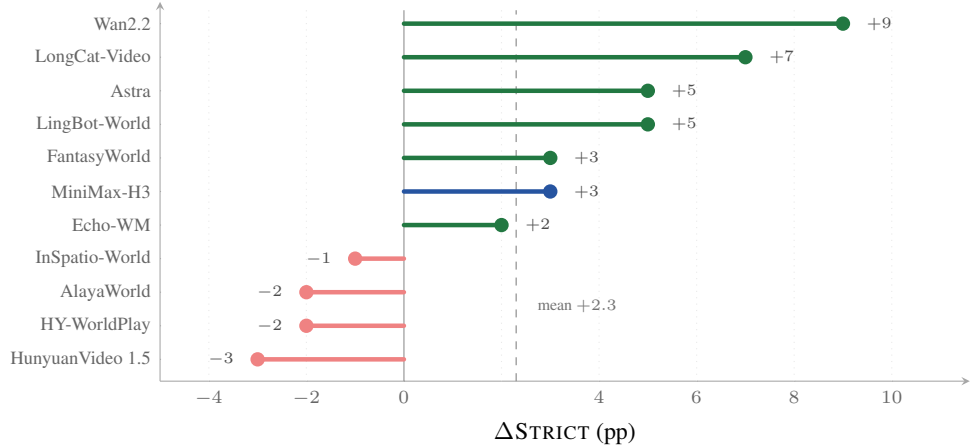
\begin{figure}[t]
\centering
\begin{tikzpicture}
\begin{axis}[width=0.88\textwidth,height=6.6cm,
  xmin=-5.0,xmax=11.6,ymin=0.35,ymax=11.65,
  axis y line=left, axis x line=bottom,
  xtick={-4,-2,0,4,6,8,10},
  ytick={1,...,11},
  yticklabels={HunyuanVideo 1.5, HY-WorldPlay, AlayaWorld, InSpatio-World,
    Echo-WM, MiniMax-H3, FantasyWorld, LingBot-World, Astra, LongCat-Video, Wan2.2},
  yticklabel style={font=\scriptsize},
  tick label style={font=\scriptsize,text=black!60},
  tick style={black!35,thin}, axis line style={black!35},
  label style={font=\small},
  xlabel={$\Delta$\textsc{Strict} (pp)},
  tickwidth=0pt, clip=false]
\draw[black!9,dotted,thin] (axis cs:-4,0.35) -- (axis cs:-4,11.4);
\draw[black!9,dotted,thin] (axis cs:-2,0.35) -- (axis cs:-2,11.4);
\draw[black!9,dotted,thin] (axis cs:2,0.35) -- (axis cs:2,11.4);
\draw[black!9,dotted,thin] (axis cs:4,0.35) -- (axis cs:4,11.4);
\draw[black!9,dotted,thin] (axis cs:6,0.35) -- (axis cs:6,11.4);
\draw[black!9,dotted,thin] (axis cs:8,0.35) -- (axis cs:8,11.4);
\draw[black!9,dotted,thin] (axis cs:10,0.35) -- (axis cs:10,11.4);
\draw[black!45,thin] (axis cs:0,0.35) -- (axis cs:0,11.4);
\draw[black!50,dashed,thin] (axis cs:2.3,0.35) -- (axis cs:2.3,11.4);
\draw[contgreendark,line width=1.7pt,line cap=round] (axis cs:0,11) -- (axis cs:9,11);
\fill[contgreendark] (axis cs:9,11) circle[radius=2.7pt];
\node[font=\tiny,text=black!70,anchor=west] at (axis cs:9.3,11) {$+9$};
\draw[contgreendark,line width=1.7pt,line cap=round] (axis cs:0,10) -- (axis cs:7,10);
\fill[contgreendark] (axis cs:7,10) circle[radius=2.7pt];
\node[font=\tiny,text=black!70,anchor=west] at (axis cs:7.3,10) {$+7$};
\draw[contgreendark,line width=1.7pt,line cap=round] (axis cs:0,9) -- (axis cs:5,9);
\fill[contgreendark] (axis cs:5,9) circle[radius=2.7pt];
\node[font=\tiny,text=black!70,anchor=west] at (axis cs:5.3,9) {$+5$};
\draw[contgreendark,line width=1.7pt,line cap=round] (axis cs:0,8) -- (axis cs:5,8);
\fill[contgreendark] (axis cs:5,8) circle[radius=2.7pt];
\node[font=\tiny,text=black!70,anchor=west] at (axis cs:5.3,8) {$+5$};
\draw[contgreendark,line width=1.7pt,line cap=round] (axis cs:0,7) -- (axis cs:3,7);
\fill[contgreendark] (axis cs:3,7) circle[radius=2.7pt];
\node[font=\tiny,text=black!70,anchor=west] at (axis cs:3.3,7) {$+3$};
\draw[contgreendark,line width=1.7pt,line cap=round] (axis cs:0,5) -- (axis cs:2,5);
\fill[contgreendark] (axis cs:2,5) circle[radius=2.7pt];
\node[font=\tiny,text=black!70,anchor=west] at (axis cs:2.3,5) {$+2$};
\draw[harnessbluedark,line width=1.7pt,line cap=round] (axis cs:0,6) -- (axis cs:3,6);
\fill[harnessbluedark] (axis cs:3,6) circle[radius=2.7pt];
\node[font=\tiny,text=black!70,anchor=west] at (axis cs:3.3,6) {$+3$};
\draw[failred!75,line width=1.7pt,line cap=round] (axis cs:0,4) -- (axis cs:-1,4);
\fill[failred!75] (axis cs:-1,4) circle[radius=2.7pt];
\node[font=\tiny,text=black!70,anchor=east] at (axis cs:-1.3,4) {$-1$};
\draw[failred!75,line width=1.7pt,line cap=round] (axis cs:0,3) -- (axis cs:-2,3);
\fill[failred!75] (axis cs:-2,3) circle[radius=2.7pt];
\node[font=\tiny,text=black!70,anchor=east] at (axis cs:-2.3,3) {$-2$};
\draw[failred!75,line width=1.7pt,line cap=round] (axis cs:0,2) -- (axis cs:-2,2);
\fill[failred!75] (axis cs:-2,2) circle[radius=2.7pt];
\node[font=\tiny,text=black!70,anchor=east] at (axis cs:-2.3,2) {$-2$};
\draw[failred!75,line width=1.7pt,line cap=round] (axis cs:0,1) -- (axis cs:-3,1);
\fill[failred!75] (axis cs:-3,1) circle[radius=2.7pt];
\node[font=\tiny,text=black!70,anchor=east] at (axis cs:-3.3,1) {$-3$};
\node[font=\tiny,text=black!55,anchor=west] at (axis cs:2.55,2.6) {mean $+2.3$};
\end{axis}
\end{tikzpicture}
\caption{Change in \textsc{Strict} from Rule-only to Rule+end on \benchname{} (Table~\ref{tab:diaggrid}).
Among the ten backends other than MiniMax-H3, six improve (green) and four decline (red).
MiniMax-H3 (blue) provides the frozen weights used by the harness, and the dashed line indicates the mean change of $+2.3$ across the other ten backends.}
\label{fig:dstrict}
\end{figure}

\paragraph{Reading the Probe Difference.}
Explicit outcome instructions yield the largest improvements when correct outcome selection and exclusion of forbidden content improve together.
Wan2.2 A14B gains $+9$ with \textsc{EndFollow} $+7$, \textsc{CntAbsent} $+9$, and a \textsc{TwinContrast} gain of $+16$.
Astra gains $+5$ points, primarily through reduced direction errors from a low initial score of 11\%, while its copy-paste rate remains at 70\%.
LongCat-Video ($+7$) and LingBot-World ($+5$) combine reduced direction errors with stable or lower leakage.
FantasyWorld gains $+3$ points through improved exclusion of forbidden content, with \textsc{CntAbsent} increasing from 75 to 81.
For HunyuanVideo 1.5, improved ending accuracy is accompanied by increased leakage from 25 to 34, the highest value among the external backends, resulting in a negative $\Delta$\textsc{Strict}.
The four backends with negative $\Delta$\textsc{Strict}, HunyuanVideo 1.5, AlayaWorld, InSpatio-World, and HY-WorldPlay, have copy-paste rates ranging from 36\% to 82\%.
These results indicate that explicit ending instructions do not reliably produce distinct outcomes across hidden states or exclude contradictory content.
The mean $\Delta$\textsc{Strict} over the other ten backends is $+2.3$ points (six rise, four fall), consistent with \S\ref{sec:q3}.
Figure~\ref{fig:dstrict} reports the eleven differences, with the dashed line indicating the mean across the ten backends excluding MiniMax-H3.

\paragraph{Failure Modes.}
The diagnostics identify three recurring failure patterns.
\emph{Copy-paste.} Under Rule+end, identical verdicts occur for both instances in 82\% of HY-WorldPlay pairs, 72\% of InSpatio-World pairs, and 70\% of Astra pairs.
Their \textsc{TwinContrast} scores range from 10\% to 22\%, indicating limited differentiation between the paired outcomes.
\emph{Direction Failure.} Astra and HY-WorldPlay have direction error rates of 89\% and 91\% under Rule-only instructions, despite \textsc{CntAbsent} scores of 96\% and 95\%.
The generated clips often exhibit limited state change, so the absence of forbidden content alone does not establish compliance.
\emph{Leakage.} HunyuanVideo 1.5 depicts forbidden content together with the required ending in 25\% of clips under Rule-only instructions.
This increases to 34\% under Rule+end, the highest rate among the external backends.
The harness reduces all three error types (Table~\ref{tab:twindiag}).

\paragraph{Twin Diagnostics by Spatial Reference.}
Table~\ref{tab:twindiag} reports the twin-mechanism diagnostics where both twins are generated under identical conditions (MiniMax-H3 single-clip runs), for both spatial-reference settings.
With the deployed first-frame reference, the harness reduces copy-paste from 30\% to 2\% and increases \textsc{TwinContrast} from 46\% to 96\%.
The text-only configuration has no copy-paste errors but higher direction error (6\% versus 4\%) and leakage (5\% versus 1\%), with the same \textsc{TwinContrast} and branch gap.

\begin{table}[t]
\centering
\caption{Counterfactual pair diagnostics for MiniMax-H3 under different spatial reference configurations.
All values are percentages computed over 50 pairs.
The deployed configuration uses a first-frame reference.}
\label{tab:twindiag}
\small
\begin{tabular}{@{}llrrrrr@{}}
\toprule
Condition & Spatial ref. & \textsc{CopyPaste} & \textsc{TwinContrast} & Direction err. & Leakage-only & Branch gap \\
\midrule
Rule-only    & none         & 34.0 & 32.0 & 34.0 & 40.0 & 0.20 \\
Rule+end     & none         & 14.0 & 64.0 & 18.0 & 53.0 & 0.14 \\
Rule-only    & first frame  & 30.0 & 46.0 & 29.0 & 44.0 & 0.14 \\
Rule+end     & first frame  & 18.0 & 66.0 & 18.0 & 53.0 & 0.10 \\
Harness      & none         & 0.0 & 96.0 & 6.0 & 5.0 & 0.02 \\
Harness      & first frame  & \textbf{2.0} & \textbf{96.0} & \textbf{4.0} & \textbf{1.0} & \textbf{0.02} \\
\bottomrule
\end{tabular}
\end{table}

\paragraph{Interpretation.}
For the unwrapped MiniMax-H3 baseline, the reported direction, leakage, and double-error diagnostics are not mutually exclusive and should not be summed.
Under Rule+end prompting, direction errors fall from 34\% to 18\%, while leakage rises from 40\% to 53\%.
This combination helps explain why \textsc{PairStrict} remains at 4\% despite higher \textsc{EndFollow}.
Figure~\ref{fig:pairtc} compares \textsc{PairStrict} and \textsc{TwinContrast}.
The harness achieves higher scores on both metrics than the evaluated unwrapped backends.

\begin{figure}[h]
\centering
\begin{tikzpicture}
\newcommand{\scapin}[5]{%
  \node[circle,font=\scriptsize,inner sep=1.7pt,draw=black!40,fill=white] (scapin#5) at (axis cs:#3,#4) {#5};
  \draw[black!32,thin] (axis cs:#1,#2) -- (scapin#5);}
\begin{axis}[width=0.66\textwidth,height=7.4cm,
  xmin=0,xmax=104,ymin=-8,ymax=100,
  xtick={0,20,40,60,80,100},ytick={0,20,40,60,80,100},
  tick label style={font=\small,text=black!60},
  label style={font=\normalsize,text=black!70},
  xlabel={\textsc{TwinContrast} (\%)},ylabel={\textsc{PairStrict} (\%)},
  tick style={black!40,thin}, axis line style={black!35},
  xmajorgrids,ymajorgrids, major grid style={black!8,dotted,thin}]
\addplot[black!45,dashed,thin] coordinates {(0,26) (104,26)};
\node[font=\scriptsize,anchor=south east,text=black!55] at (axis cs:102,27.2) {best baseline $26\%$};
\addplot[only marks,mark=*,mark size=4.0pt,draw=white,line width=0.7pt,fill=modelgreydark]
  coordinates {(54,26) (30,8) (32,4) (64,4) (50,14) (34,8) (34,6) (16,4) (22.4,0) (10,0)};
\addplot[only marks,mark=star,mark size=6.5pt,draw=contgreendark,line width=1.1pt,fill=contgreendark]
  coordinates {(96,90)};
\scapin{54}{26}{54}{32}{1}
\scapin{50}{14}{50}{20}{5}
\scapin{34}{8}{40}{13}{7}
\scapin{32}{4}{37}{0.5}{3}
\scapin{16}{4}{16}{-1.8}{9}
\scapin{22.4}{0}{22.4}{5.4}{10}
\scapin{10}{0}{10}{-5.2}{11}
\node[circle,font=\scriptsize,inner sep=1.7pt,draw=black!40,fill=white] (scapin4) at (axis cs:71,4) {4};
\draw[black!32,thin] (axis cs:64,4) -- (scapin4);
\node[circle,font=\scriptsize,inner sep=1.7pt,draw=black!40,fill=white] (scapin8) at (axis cs:42,6) {8};
\draw[black!32,thin] (axis cs:34,6) -- (scapin8);
\node[circle,font=\scriptsize,inner sep=2.0pt,draw=black!40,fill=white] (scapin26) at (axis cs:22.5,13.5) {2,\,6};
\draw[black!32,thin] (axis cs:30,8) -- (scapin26);
\node[font=\small,anchor=south east,text=contgreendark] at (axis cs:96,92) {\sysname{}};
\end{axis}
\end{tikzpicture}\\[3pt]
{\footnotesize
1~Wan2.2\quad 2~LongCat-Video\quad 3~HunyuanVideo~1.5\quad 4~MiniMax-H3\quad
5~LingBot-World\quad 6~Echo-WM\quad 7~FantasyWorld\\
8~AlayaWorld\quad 9~InSpatio-World\quad 10~Astra\quad 11~HY-WorldPlay\par}
\caption{\textsc{PairStrict} and \textsc{TwinContrast} under Rule+end instructions (Table~\ref{tab:bothprobes}).
Distinct endings do not necessarily imply strict compliance.
The harness, marked by a star, achieves high scores on both metrics.}
\label{fig:pairtc}
\end{figure}

\paragraph{Per-Trigger-Type Harness Performance.}
For the deployed first-frame-reference harness, the reported strict / pair-strict scores are 100.0\% / 100.0\% for cast (20 clips), 95.2\% / 90.5\% for gesture (42), and 92.3\% / 84.6\% for mechanism (26).
The 12 zone-triggered clips achieve 91.7\% \textsc{Strict} and 83.3\% \textsc{PairStrict}, corresponding to 11 of 12 clips and 5 of 6 pairs.
This family has the lowest text-only \textsc{Strict} score of 66.7\% and the largest improvement from the first-frame reference.
The twin diagnostics in Table~\ref{tab:twindiag} further separate the remaining errors into three types under both spatial-reference settings.

\subsection{WBench External Evaluation: Per-Metric Grid and Channel Ablation}
\label{app:wbench}

WBench \citep{wbench2026} offers an external evaluation of video quality, setting, interaction, consistency, and physical behavior.
Table~\ref{tab:wbench} reproduces its reported per-metric scores for the twenty models of the original release and adds our two entries as columns.
We evaluate frozen MiniMax-H3 under the official text protocol and the full harness with the spatial preview video using the released evaluation pipeline.
Table~\ref{tab:wbenchmain} reports aggregate scores, and Table~\ref{tab:wbench} reports individual metrics.
\S\ref{sec:ablation} analyzes the reference channel ablation, with additional details provided below.
LingBot-World and Astra appear both in the published WBench table and in our unwrapped \benchname{} evaluation, and the harness comparison reported here deploys MiniMax-H3 as its frozen backbone.

\begin{table}[t]
\centering
\caption[WBench per-metric grid]{WBench per-metric grid, scores as reported by \citealt{wbench2026} (gray $=$ class average).
Our two configurations are excluded from the class averages.
Their dimension averages correspond to the leaderboard scores in Table~\ref{tab:wbenchmain}. For our entry, the Interaction average is the official leaderboard value and is not the mean of the sub-scores shown.}
\label{tab:wbench}
\scriptsize
\setlength{\tabcolsep}{1.8pt}
\renewcommand{\arraystretch}{1.25}
\resizebox{\textwidth}{!}{%
\begin{tabular}{@{}cl|*{10}{c}c|*{5}{c}c|*{6}{c}c|c@{}}
\thickhline
Group & Metric & \multicolumn{11}{c|}{\textcolor{blue!55!black}{\textbf{Text-driven}}} & \multicolumn{6}{c|}{\textcolor{teal!45!black}{\textbf{Camera-controlled}}} & \multicolumn{7}{c|}{\textcolor{orange!60!black}{\textbf{Action-conditioned}}} \\
 & & \rotatebox{90}{\scriptsize\textbf{Seedance 1.5}}
  & \rotatebox{90}{\scriptsize\textbf{Wan 2.7}}
  & \rotatebox{90}{\scriptsize\textbf{Kling 3.0}}
  & \rotatebox{90}{\scriptsize\textbf{YUME 1.5}}
  & \rotatebox{90}{\scriptsize\textbf{HY-Video 1.5}}
  & \rotatebox{90}{\scriptsize\textbf{LTX 2.3}}
  & \rotatebox{90}{\scriptsize\textbf{LongCat-Video}}
  & \rotatebox{90}{\scriptsize\textbf{Kairos 3.0}}
  & \rotatebox{90}{\scriptsize\textbf{Cosmos 2.5}}
  & \rotatebox{90}{\scriptsize\textbf{H3 (official)}}
  & \rotatebox{90}{\scriptsize\textit{Average}}
  & \rotatebox{90}{\scriptsize\textbf{LingBot-World}}
  & \rotatebox{90}{\scriptsize\textbf{HY-World 1.5}}
  & \rotatebox{90}{\scriptsize\textbf{Fantasy-World}}
  & \rotatebox{90}{\scriptsize\textbf{InSpatio-World}}
  & \rotatebox{90}{\scriptsize\textbf{Astra}}
  & \rotatebox{90}{\scriptsize\textit{Average}}
  & \rotatebox{90}{\scriptsize\textbf{Happy Oyster}}
  & \rotatebox{90}{\scriptsize\textbf{Matrix-Game 3.0}}
  & \rotatebox{90}{\scriptsize\textbf{Genie 3}}
  & \rotatebox{90}{\scriptsize\textbf{Matrix-Game 2.0}}
  & \rotatebox{90}{\scriptsize\textbf{HY-GameCraft}}
  & \rotatebox{90}{\scriptsize\textbf{Infinite-World}}
  & \rotatebox{90}{\scriptsize\textit{Average}}
  & \rotatebox{90}{\scriptsize\textbf{Ours}}
  \\
\noalign{\vskip -1pt}
 & & \wicon{blue!10}{bytedance.png} & \wicon{blue!10}{wan.png} & \wicon{blue!10}{kling.jpeg}
   & \wicon{blue!10}{shlab.png} & \wicon{blue!10}{hunyuan.png} & \wicon{blue!10}{lightrix.jpeg}
   & \wicon{blue!10}{longcat.png} & \wicon{blue!10}{kairos.png} & \wicon{blue!10}{cosmos.png} & \wicon{blue!10}{minimax.png} & &
   \wicon{teal!10}{lingbot.png} & \wicon{teal!10}{hunyuan.png} & \wicon{teal!10}{amap.png}
   & \wicon{teal!10}{inspatio.jpeg} & \wicon{teal!10}{thu.png} & &
   \wicon{orange!10}{alibaba.png} & \wicon{orange!10}{skywork.jpeg} & \wicon{orange!10}{google.png}
   & \wicon{orange!10}{skywork.jpeg} & \wicon{orange!10}{hunyuan.png} & \wicon{orange!10}{nankai.png} & & \ocell \\
\hline
\multirow{7}{*}{\rotatebox{90}{\textbf{Video Quality}}}
 & Aesthetic & 59.7 & 59.6 & 61.3 & 59.3 & 61.9 & 56.9 & 64.7 & 58.4 & 60.1 & 63.1 & \g{60.2} & 66.9 & 60.1 & 63.0 & 64.4 & 48.6 & \g{60.6} & 56.6 & 46.4 & 51.6 & 54.0 & 52.6 & 58.7 & \g{53.3} & 59.1 \\
 & Imaging & 69.8 & 68.1 & 67.7 & 65.7 & 67.4 & 62.3 & 69.8 & 63.6 & 67.2 & 66.2 & \g{66.8} & 67.9 & 65.4 & 62.8 & 67.6 & 52.5 & \g{63.2} & 63.9 & 70.0 & 59.3 & 60.3 & 58.7 & 66.1 & \g{63.1} & 64.6 \\
 & Flickering & 93.4 & 93.0 & 94.5 & 94.8 & 95.5 & 94.1 & 94.9 & 96.3 & 96.0 & 96.2 & \g{94.7} & 94.1 & 93.5 & 95.8 & 96.0 & 96.0 & \g{95.1} & 94.0 & 86.3 & 95.0 & 94.6 & 93.7 & 94.1 & \g{93.0} & 95.7 \\
 & Dynamic & 98.3 & 99.3 & 89.9 & 86.1 & 68.8 & 94.4 & 59.7 & 63.5 & 42.4 & 73.7 & \g{78.0} & 66.2 & 91.1 & 49.0 & 26.1 & 79.6 & \g{62.4} & 94.2 & 97.5 & 92.4 & 94.9 & 96.8 & 82.8 & \g{93.1} & 85.5 \\
 & Smoothness & 97.6 & 96.5 & 97.9 & 97.7 & 98.8 & 96.8 & 97.7 & 97.9 & 98.3 & 98.6 & \g{97.7} & 96.9 & 98.1 & 97.9 & 98.8 & 97.7 & \g{97.9} & 97.0 & 95.4 & 97.8 & 98.2 & 97.6 & 98.0 & \g{97.3} & 98.6 \\
 & HPSv3-Norm & 72.9 & 69.4 & 68.8 & 62.0 & 67.5 & 57.7 & 76.3 & 58.8 & 65.9 & 69.8 & \g{66.6} & 81.4 & 60.5 & 65.8 & 76.1 & 28.0 & \g{62.4} & 58.3 & 57.1 & 55.2 & 41.0 & 38.3 & 62.3 & \g{52.0} & 64.2 \\
 & \textbf{Average} & 82.0 & 81.0 & 80.0 & 77.6 & 76.7 & 77.0 & 77.2 & 73.1 & 71.7 & 77.9 & \g{77.4} & 78.9 & 78.1 & 72.4 & 71.5 & 67.1 & \g{73.6} & 77.3 & 75.5 & 75.2 & 73.8 & 73.0 & 77.0 & \g{75.3} & 78.0 \\
\hline
\multirow{3}{*}{\rotatebox{90}{\textbf{Setting}}}
 & Scene & 71.6 & 88.3 & 89.0 & 53.1 & 77.5 & 81.3 & 53.1 & 52.2 & 72.4 & 54.3 & \g{70.9} & 51.6 & 53.5 & 52.4 & 51.7 & 43.4 & \g{50.5} & 57.4 & 48.9 & 61.1 & 49.4 & 50.6 & 54.0 & \g{53.6} & 87.9 \\
 & Subject & 94.2 & 94.6 & 92.9 & 91.7 & 93.6 & 89.2 & 91.5 & 88.5 & 94.2 & 87.9 & \g{92.3} & 93.6 & 90.8 & 90.1 & 91.1 & 75.9 & \g{88.3} & 91.1 & 78.4 & 83.8 & 84.9 & 82.5 & 84.5 & \g{84.2} & 86.5 \\
 & \textbf{Average} & 82.9 & 91.4 & 91.0 & 72.4 & 85.6 & 85.2 & 72.3 & 70.3 & 83.3 & 71.1 & \g{81.6} & 72.6 & 72.2 & 71.3 & 71.4 & 59.6 & \g{69.4} & 74.2 & 63.6 & 72.5 & 67.1 & 66.6 & 69.3 & \g{68.9} & 87.2 \\
\hline
\multirow{5}{*}{\rotatebox{90}{\textbf{Interaction}}}
 & Navigation & 66.3 & 64.4 & 69.4 & 71.4 & 71.4 & 66.4 & 62.1 & 64.1 & 63.1 & 68.9 & \g{66.5} & 80.1 & 86.8 & 71.9 & 73.2 & 66.9 & \g{75.8} & 84.9 & 83.6 & 73.4 & 80.3 & 66.3 & 75.4 & \g{77.3} & 77.7 \\
 & Event Editing & 80.4 & 84.0 & 81.4 & 57.8 & 63.8 & 53.0 & 50.4 & 46.8 & 48.2 & 77.6 & \g{62.9} & -- & -- & -- & -- & -- & \g{--} & -- & -- & -- & -- & -- & -- & \g{--} & 81.8 \\
 & Subject Action & 80.0 & 83.4 & 85.6 & 47.0 & 55.6 & 51.8 & 48.4 & 41.4 & 41.6 & 74.3 & \g{59.4} & -- & -- & -- & -- & -- & \g{--} & -- & -- & -- & -- & -- & -- & \g{--} & 85.8 \\
 & Persp.\ Switching & 45.0 & 55.0 & 55.0 & 16.7 & 27.6 & 25.0 & 18.3 & 13.3 & 20.0 & 70.5 & \g{30.7} & -- & -- & -- & -- & -- & \g{--} & -- & -- & -- & -- & -- & -- & \g{--} & 59.7 \\
 & \textbf{Average} & 67.9 & 71.7 & 72.8 & 48.2 & 54.6 & 49.0 & 44.8 & 41.4 & 43.2 & 68.9 & \g{54.8} & -- & -- & -- & -- & -- & \g{--} & -- & -- & -- & -- & -- & -- & \g{--} & 77.7 \\
\hline
\multirow{9}{*}{\rotatebox{90}{\textbf{Consistency}}}
 & Background & 89.6 & 89.5 & 92.7 & 92.0 & 92.4 & 89.3 & 94.7 & 91.8 & 92.3 & 89.4 & \g{91.6} & 96.9 & 92.7 & 94.2 & 95.0 & 85.3 & \g{92.8} & 91.4 & 85.7 & 90.7 & 86.9 & 86.5 & 88.8 & \g{88.3} & 90.3 \\
 & Spatial & 72.7 & 71.0 & 75.2 & 71.5 & 79.2 & 70.2 & 83.3 & 76.8 & 78.1 & 96.5 & \g{75.3} & 92.7 & 90.6 & 80.6 & 93.8 & 64.7 & \g{84.5} & 77.7 & 81.0 & 79.9 & 64.5 & 60.5 & 74.9 & \g{73.1} & 74.8 \\
 & Gated Spatial & 72.4 & 71.0 & 75.1 & 71.4 & 75.1 & 70.2 & 66.2 & 62.0 & 74.3 & 91.3 & \g{70.9} & 67.1 & 84.9 & 64.2 & 66.5 & 63.3 & \g{69.2} & 75.8 & 80.4 & 78.4 & 64.5 & 60.5 & 74.4 & \g{72.3} & 74.5 \\
 & Segment & 92.4 & 65.6 & 92.7 & 99.3 & 99.3 & 77.8 & 98.6 & 94.1 & 93.1 & 88.9 & \g{90.3} & 99.4 & 100.0 & 100.0 & 100.0 & 86.6 & \g{97.2} & 96.2 & 89.8 & 93.6 & 21.0 & 99.4 & 100.0 & \g{83.3} & 62.6 \\
 & Perspective & 62.7 & 62.2 & 76.8 & 48.0 & 86.6 & 69.8 & 81.5 & 76.3 & 84.3 & 81.5 & \g{72.0} & 90.9 & 62.5 & 79.8 & 72.5 & 30.0 & \g{67.1} & 75.0 & 13.3 & 54.5 & 29.2 & 17.9 & 33.8 & \g{37.3} & 60.2 \\
 & Subject & 89.3 & 88.7 & 88.5 & 89.4 & 91.5 & 86.7 & 92.4 & 90.7 & 91.8 & 78.3 & \g{89.9} & 93.5 & 89.1 & 92.5 & 94.4 & 83.5 & \g{90.6} & 91.5 & 83.0 & 90.4 & 87.2 & 82.6 & 88.4 & \g{87.2} & 88.4 \\
 & Geometric & 83.5 & 82.6 & 89.4 & 91.1 & 94.4 & 81.1 & 94.7 & 91.5 & 94.2 & 58.3 & \g{89.2} & 95.4 & 92.0 & 95.3 & 97.3 & 85.6 & \g{93.1} & 87.2 & 87.6 & 88.6 & 86.1 & 88.3 & 94.3 & \g{88.7} & 90.3 \\
 & Photometric & 76.7 & 75.5 & 80.4 & 84.1 & 81.4 & 79.4 & 81.5 & 82.1 & 82.1 & 80.4 & \g{80.4} & 83.3 & 83.1 & 84.8 & 87.4 & 87.5 & \g{85.2} & 79.8 & 75.3 & 84.5 & 81.3 & 85.0 & 85.1 & \g{81.8} & 81.2 \\
 & \textbf{Average} & 79.9 & 75.8 & 83.9 & 80.9 & 87.5 & 78.1 & 86.6 & 83.2 & 86.3 & 83.8 & \g{82.5} & 89.9 & 86.9 & 86.4 & 88.4 & 73.3 & \g{85.0} & 84.3 & 74.5 & 82.6 & 65.1 & 72.6 & 80.0 & \g{76.5} & 77.8 \\
\hline
\multirow{3}{*}{\rotatebox{90}{\textbf{Physical}}}
 & Causal Fidelity & 76.0 & 83.3 & 78.0 & 72.7 & 75.0 & 74.0 & 76.0 & 62.7 & 74.7 & 41.9 & \g{74.7} & 77.7 & 74.0 & 74.0 & 67.3 & 48.3 & \g{68.3} & 69.3 & 64.7 & 71.7 & 59.3 & 68.3 & 67.0 & \g{66.7} & 80.1 \\
 & Visual Plausibility & 60.5 & 59.8 & 60.4 & 58.1 & 59.3 & 56.2 & 60.8 & 58.2 & 59.3 & 79.6 & \g{59.2} & 64.8 & 58.6 & 59.7 & 63.1 & 54.6 & \g{60.2} & 57.6 & 54.0 & 59.7 & 55.0 & 56.5 & 57.2 & \g{56.7} & 57.5 \\
 & \textbf{Average} & 68.2 & 71.6 & 69.2 & 65.4 & 67.1 & 65.1 & 68.4 & 60.5 & 67.0 & 69.3 & \g{66.9} & 71.2 & 66.3 & 66.8 & 65.2 & 51.4 & \g{64.2} & 63.5 & 59.3 & 65.7 & 57.2 & 62.4 & 62.1 & \g{61.7} & 68.8 \\
\thickhline
\end{tabular}}
\end{table}

\subsubsection{Channel Ablation Details}
\label{app:ablations}

The WBench ablation in \S\ref{sec:ablation} holds MiniMax-H3 weights fixed across four configurations.
These comprise the official text protocol, the harness with first-frame anchoring, the harness with a spatial preview video, and the raw text protocol with the same preview video.
Table~\ref{tab:ablationnav} reports the dimensions contributing to the navigation score in Table~\ref{tab:ablation}.
The interaction column corresponds to camera trajectory, which is the only interaction metric defined for this split.

\begin{table}[h]
\centering
\caption{Reference channel ablation on WBench with frozen MiniMax-H3 weights.
The configurations compare first-frame anchoring and spatial preview conditioning, including the same preview supplied without the harness.
\textbf{Bold} indicates the best overall score on each split.}
\label{tab:ablation}
\scriptsize
\setlength{\tabcolsep}{4.5pt}
\resizebox{\textwidth}{!}{%
\begin{tabular}{@{}l c c cccc c@{}}
\toprule
 & \textsc{Navigation} & \multicolumn{6}{c}{\textsc{Full split}} \\
\cmidrule(lr){2-2}\cmidrule(lr){3-8}
Configuration & \textsc{Overall} & \textsc{Quality}$\uparrow$ & \textsc{Setting}$\uparrow$
 & \textsc{Interaction}$\uparrow$ & \textsc{Consistency}$\uparrow$ & \textsc{Physical}$\uparrow$
 & \textsc{Overall}$\uparrow$ \\
\midrule
H3 (raw text protocol) & 74.7 & 77.9 & 71.1 & 68.9 & 83.8 & 69.3 & 74.2 \\
H3 (raw text protocol) + control video & 65.7 & 79.3 & 75.3 & 55.7 & 67.1 & 58.2 & 67.1 \\
H3 + harness, first-frame reference & 76.8 & 77.6 & 89.6 & 69.0 & 75.3 & 69.0 & 76.1 \\
\ocell  \sysname{} & \ocell\textbf{79.0} & \ocell78.0 & \ocell87.2 & \ocell77.7 & \ocell77.8 & \ocell68.8 & \ocell\textbf{77.9} \\
\bottomrule
\end{tabular}}
\end{table}

\begin{table}[h]
\centering
\caption{Navigation split dimensions for the reference channel ablation using frozen MiniMax-H3 weights and WBench leaderboard scores.
The interaction column is the navigation trajectory metric.}
\label{tab:ablationnav}
\small
\setlength{\tabcolsep}{5pt}
\resizebox{\textwidth}{!}{%
\begin{tabular}{@{}l ccccc c@{}}
\toprule
Configuration & \textsc{Quality}$\uparrow$ & \textsc{Setting}$\uparrow$
 & \textsc{Nav.\ traj.}$\uparrow$ & \textsc{Consistency}$\uparrow$ & \textsc{Physical}$\uparrow$
 & \textsc{Overall}$\uparrow$ \\
\midrule
H3 (raw text protocol) & 77.5 & 72.7 & 68.9 & 85.0 & 69.4 & 74.7 \\
H3 (raw text protocol) + control video & 73.3 & 53.0 & 79.3 & 64.8 & 58.0 & 65.7 \\
H3 + harness, first-frame reference & 79.5 & 91.5 & 69.0 & 75.0 & 69.0 & 76.8 \\
\ocell \sysname{} & \ocell80.1 & \ocell89.1 & \ocell77.7 & \ocell79.2 & \ocell68.8 & \ocell79.0 \\
\bottomrule
\end{tabular}}
\end{table}

The navigation split exhibits the same overall pattern as the full split, with larger differences between configurations.
First-frame anchoring increases setting adherence from 72.7 to 91.5 and the overall navigation score to 76.8.
It also reduces consistency from 85.0 to 75.0, with little change in camera trajectory (68.9 to 69.0).
Adding the control video increases camera trajectory from 69.0 to 77.7 and the overall navigation score to 79.0.
Consistency improves by 4.2 points, from 75.0 to 79.2.
Video quality increases across the three configurations from 77.5 to 79.5 to 80.1, while physical plausibility remains within 0.6 points of the raw text configuration.
The raw text protocol with the control video obtains the highest camera trajectory score of 79.3 but an overall navigation score of 65.7.
Its setting adherence (53.0), consistency (64.8), and physical plausibility (58.0) are all below the raw text configuration without the preview.
These results are consistent with the observed reproduction of preview geometry when the instructions do not specify how the spatial reference should be interpreted.

\paragraph{Scope.}
The harness bundles settlement, the \vsm{} compiler, the anchor latch, and the reference channels, so this ablation isolates the reference channel under a fixed harness.
The complementary decomposition holds the backend fixed and varies the instruction and state components, and is read off Table~\ref{tab:main} together with Appendix Table~\ref{tab:bothprobes} (\S\ref{sec:harnessexp}).

\section{Additional Qualitative Examples}
\label{app:qualsup}

Figure~\ref{fig:qual} presents two counterfactual pairs in the main text.
This appendix includes seven additional pairs, with four pairs shown per instruction condition because the orb pair also appears under Rule+end.
The examples illustrate differences between the unwrapped baselines and the full harness.
The layout follows Figure~\ref{fig:qual}, with the shared first frame and rule shown for each pair.
Each cell contains four frames sampled uniformly from a five-second clip.
The verdict markers follow the protocol used in Table~\ref{tab:main}, with \checkmark indicating only the required ending and $\times$ indicating forbidden content or a missing required ending.
Baselines receive the corresponding instruction condition, while the harness uses the same frozen MiniMax-H3 weights with its compiled instruction.

\begin{figure}[p]
\centering
\includegraphics[width=\textwidth]{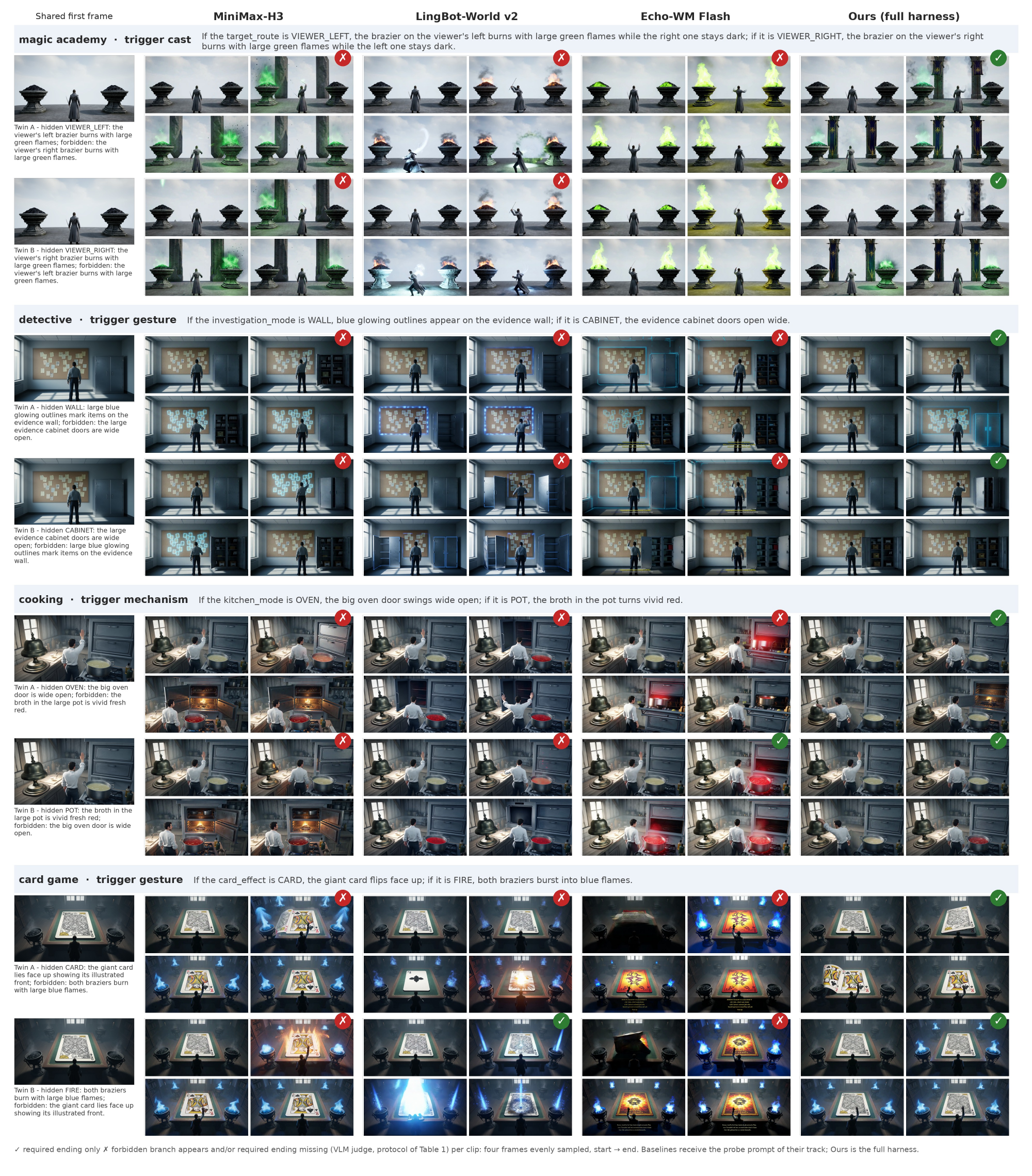}
\caption{Four additional twin pairs under the \emph{Rule-only} probe, in the format of Figure~\ref{fig:qual} (badge conventions as there).
The harness depicts only the required ending in all eight illustrated clips.}
\label{fig:supruleonly}
\end{figure}

\begin{figure}[p]
\centering
\includegraphics[width=\textwidth]{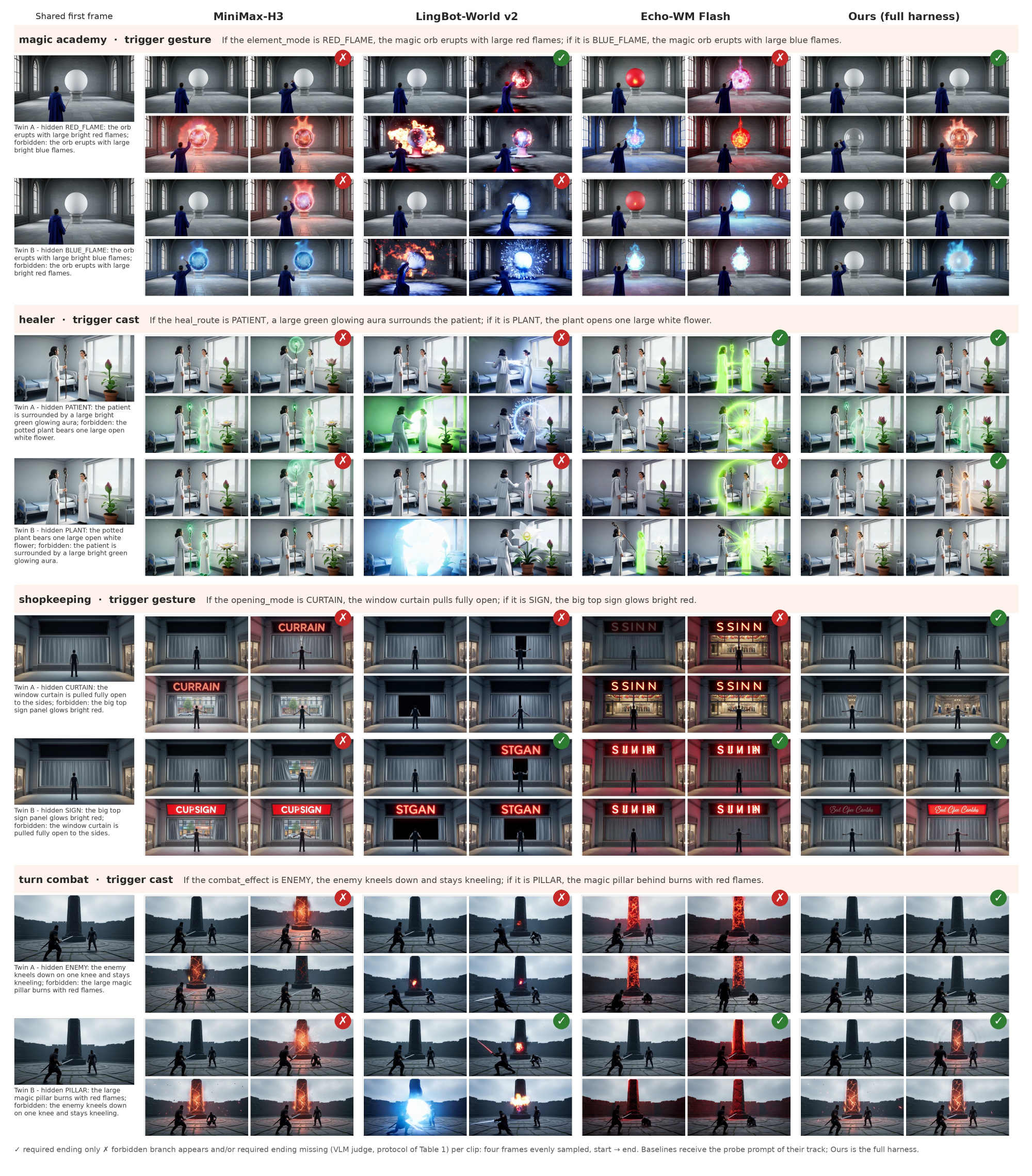}
\caption{Four twin pairs under the \emph{Rule+end} probe, same format as Figure~\ref{fig:supruleonly}.
The harness depicts only the required ending in all eight illustrated clips.}
\label{fig:supruleend}
\end{figure}

Figure~\ref{fig:supruleonly} covers the Rule-only probe and Figure~\ref{fig:supruleend} the Rule+end probe.
Two recurring baseline failure patterns are visible.
In some examples, the rendered ending does not vary with the hidden state.
Under Rule-only instructions, all three baselines produce red broth in both instances of one pair, violating the required outcome for one instance.
Under Rule+end instructions, MiniMax-H3 activates the magic pillar in both instances of another pair.
Other examples include visual content associated with the opposite hidden state.
Echo-WM generates the orb with the color required for the other instance before the action resolves, and LingBot depicts the card flipping while also showing flames.
The harness receives the resolved event through the compiled instruction and depicts only the required ending in all sixteen illustrated clips.
The sign-panel pair in the third group of Figure~\ref{fig:supruleend} also illustrates differences in visual legibility under Rule+end instructions.
The baselines produce malformed text, whereas the harness preserves the dark sign or generates cursive lettering according to the required outcome.

\begin{figure}[p]
\centering
\resizebox{0.98\textwidth}{!}{%
\begin{tikzpicture}[x=1mm, y=1mm]
\newcommand{\stcell}[6]{%
  \node[align=center] at ({#1+42.5},{#2-2.6}) {{\scriptsize\bfseries #3}\\{\scriptsize\itshape #4}};
  \node[draw=black!30, line width=0.4, inner sep=0pt, anchor=north west] at (#1,{#2-6.8})
    {\includegraphics[width=42mm,height=23.33mm]{figures/assets/demo/stam/#5}\hspace{1mm}%
\includegraphics[width=42mm,height=23.33mm]{figures/assets/demo/stam/#6}};}
\stcell{0}{0}{act 1: sunlit street}{stamina 28/100 $\cdot$ quest 1/3}{t01a.jpg}{t01b.jpg}
\stcell{0}{-33.5}{act 2: caf\'e sign}{stamina 16/100 $\cdot$ quest 1/3}{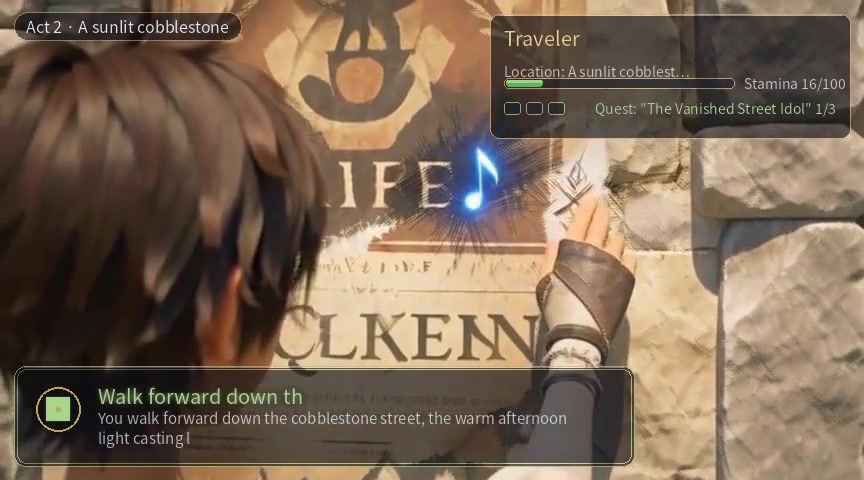}{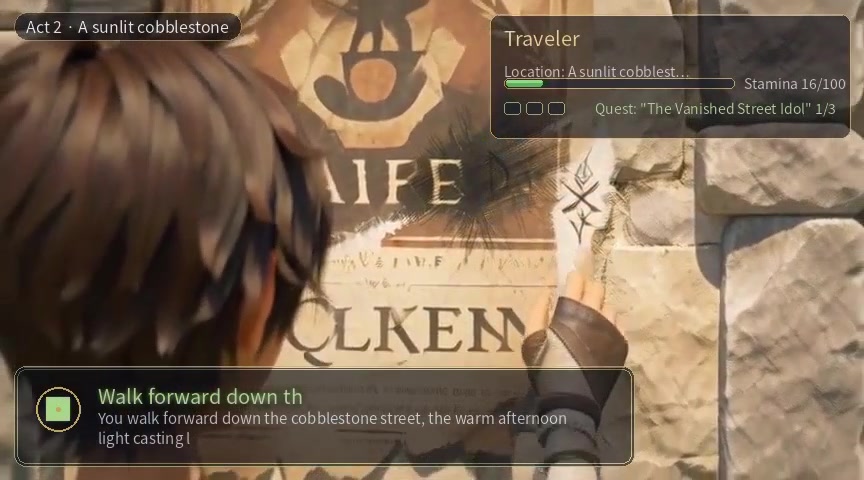}
\stcell{0}{-67}{act 3: market awnings}{stamina 4/100 $\cdot$ quest 1/3}{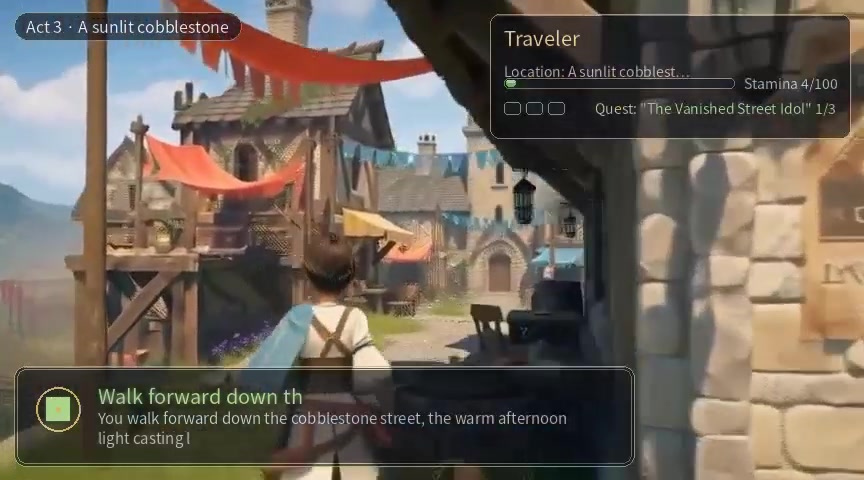}{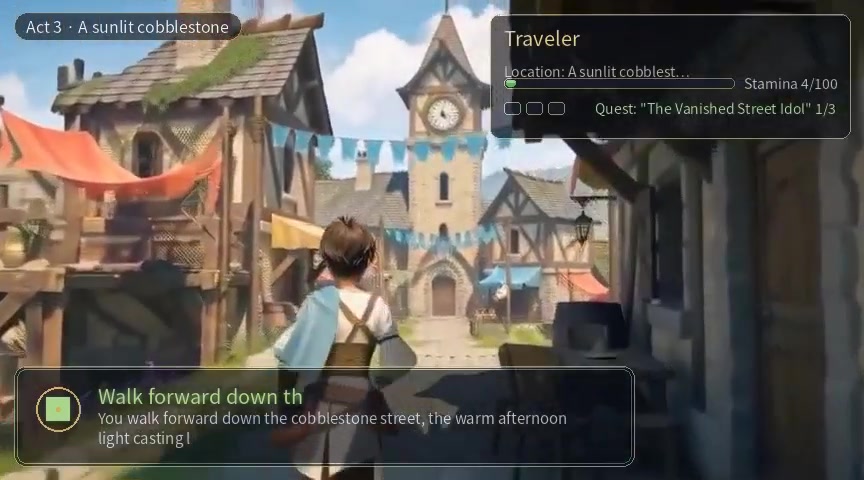}
\stcell{0}{-100.5}{act 4: clock-tower square}{stamina 0/100 $\cdot$ exhausted, slow walk}{t04a.jpg}{t04b.jpg}
\stcell{0}{-134}{act 5: resting}{stamina 30/100 $\cdot$ exhausted}{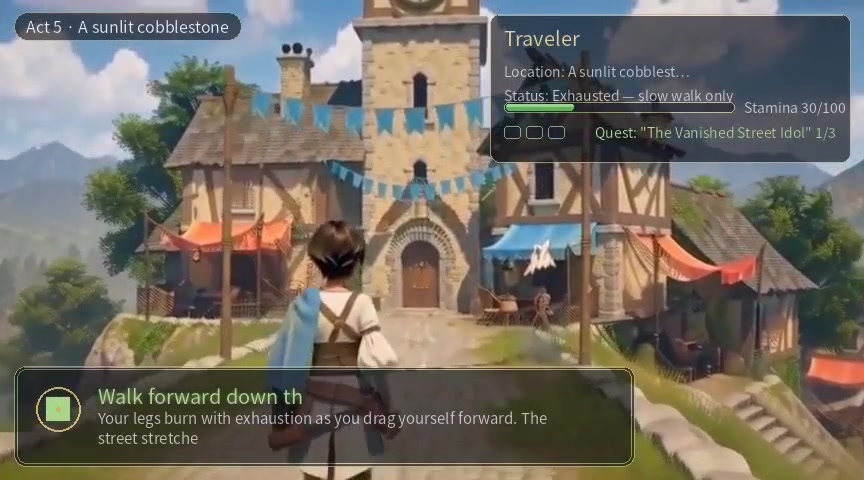}{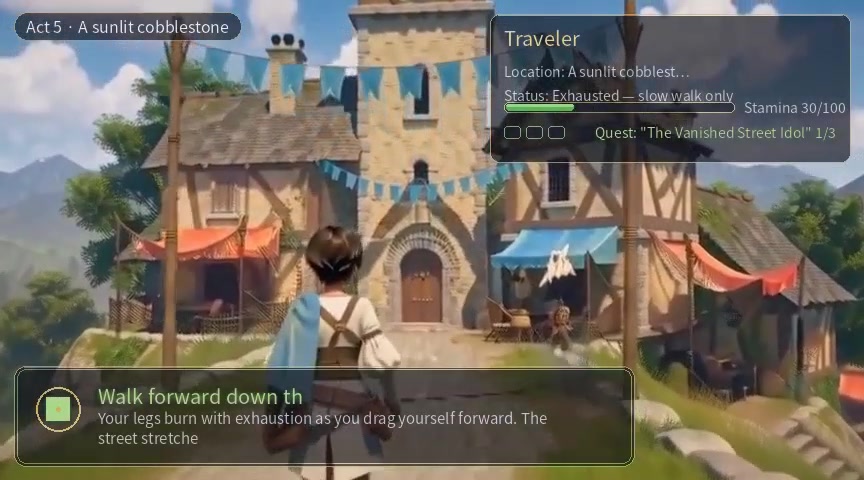}
\stcell{0}{-167.5}{act 6: walking on}{stamina 60/100 $\cdot$ rested}{t06a.jpg}{t06b.jpg}
\stcell{0}{-201}{act 7: stall row}{stamina 48/100 $\cdot$ quest 2/3}{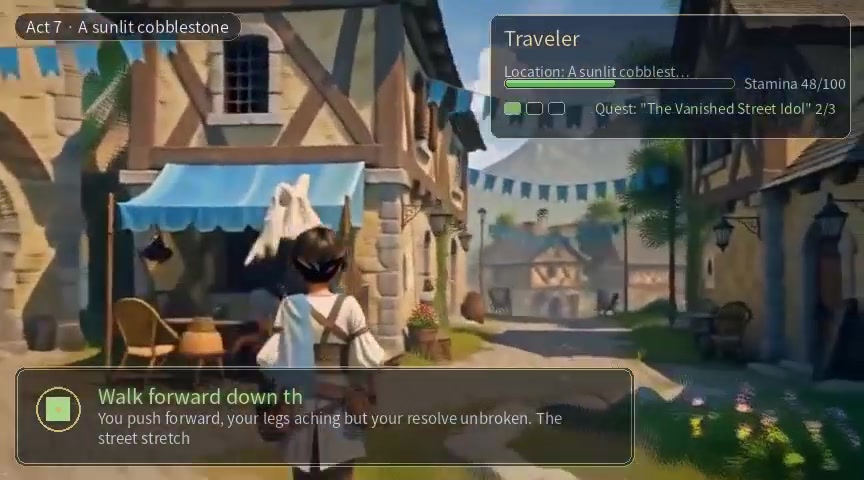}{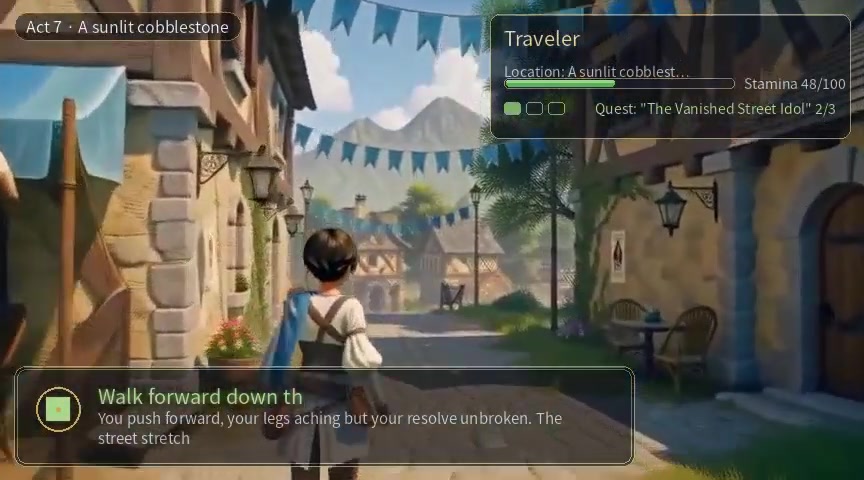}
\stcell{0}{-234.5}{act 8: blue banner}{stamina 36/100 $\cdot$ quest 2/3}{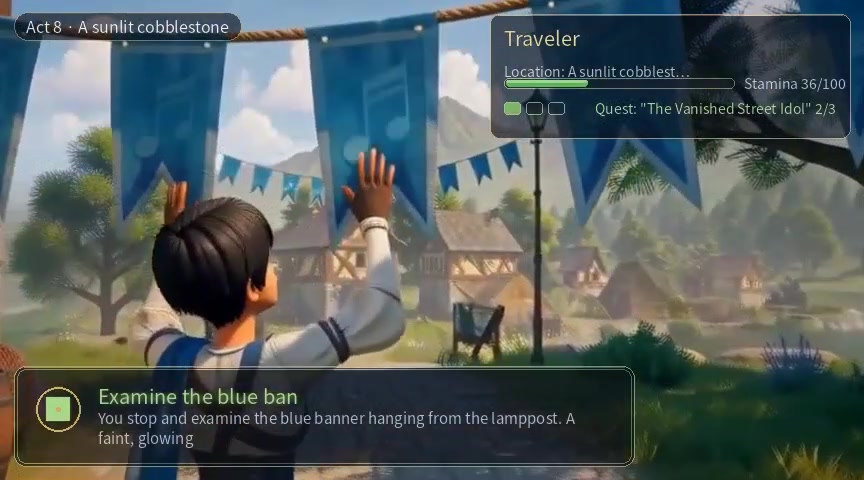}{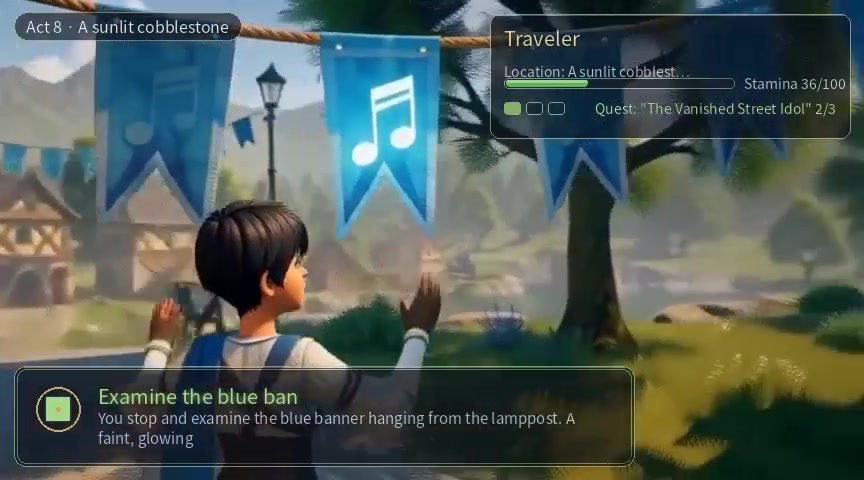}
\stcell{90}{0}{act 9: tracing the sigil}{stamina 24/100 $\cdot$ quest 2/3}{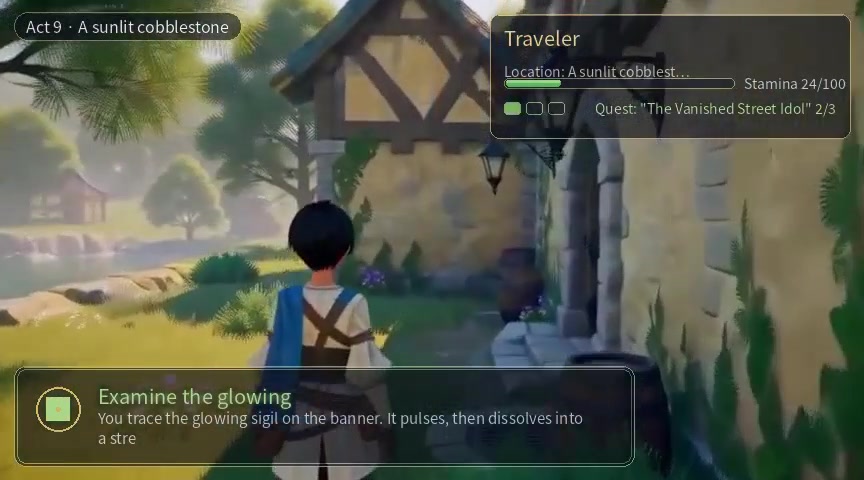}{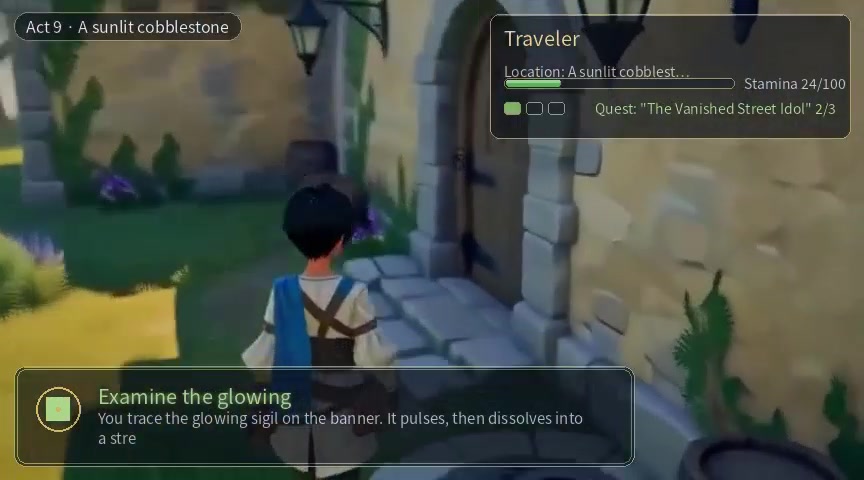}
\stcell{90}{-33.5}{act 10: following the stream}{stamina 12/100 $\cdot$ quest 3/3}{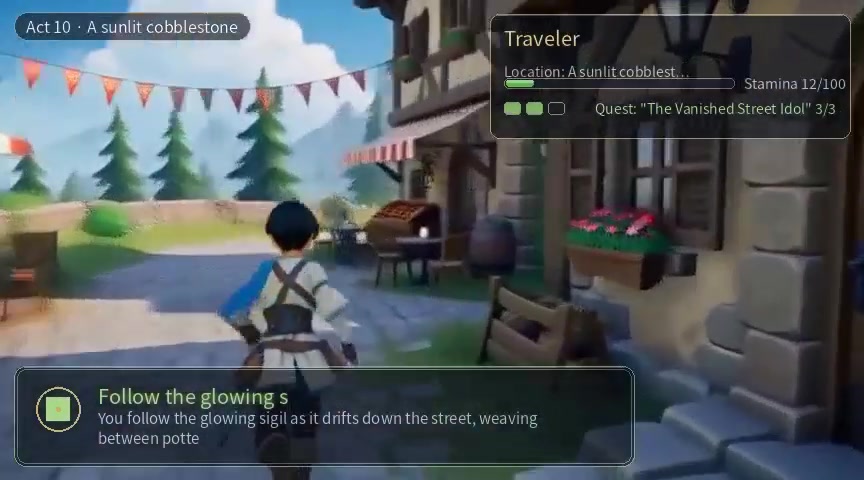}{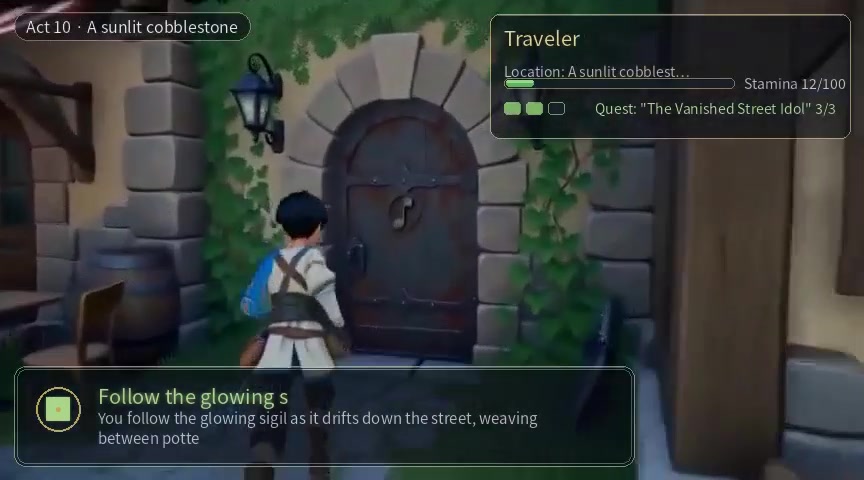}
\stcell{90}{-67}{act 11: stone stairs}{stamina 0/100 $\cdot$ exhausted, slow walk}{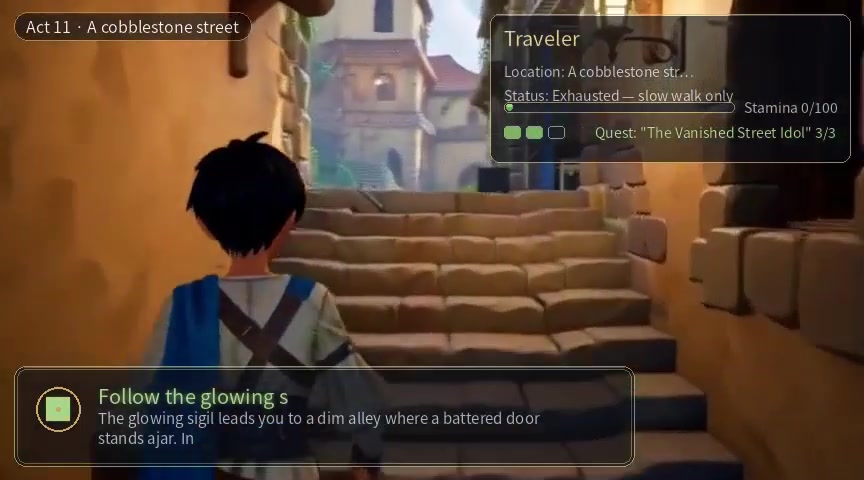}{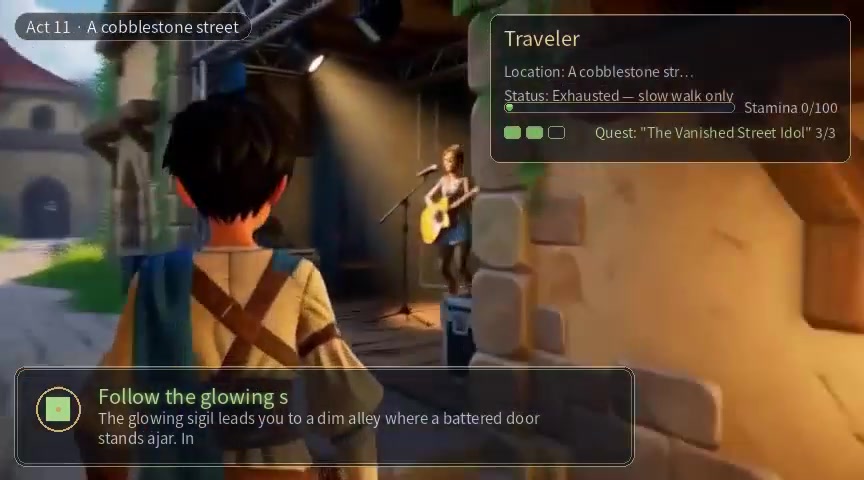}
\stcell{90}{-100.5}{act 12: the battered door}{stamina 30/100 $\cdot$ exhausted}{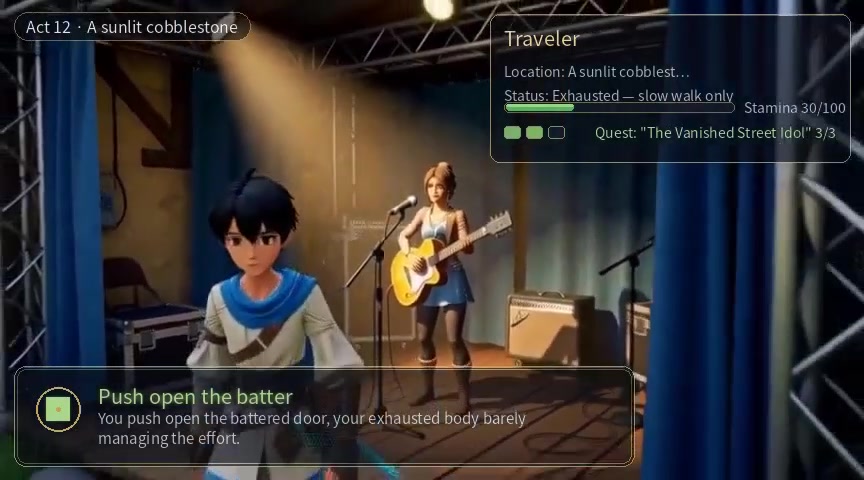}{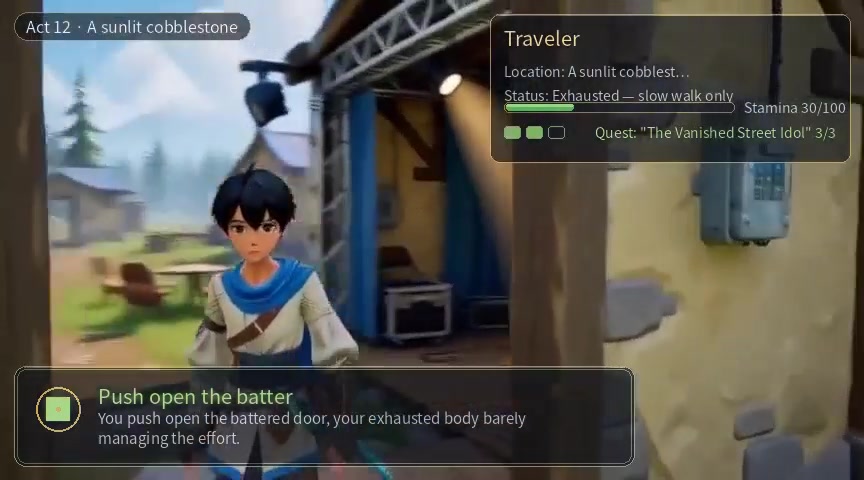}
\stcell{90}{-134}{act 13: through the door}{stamina 60/100 $\cdot$ rested}{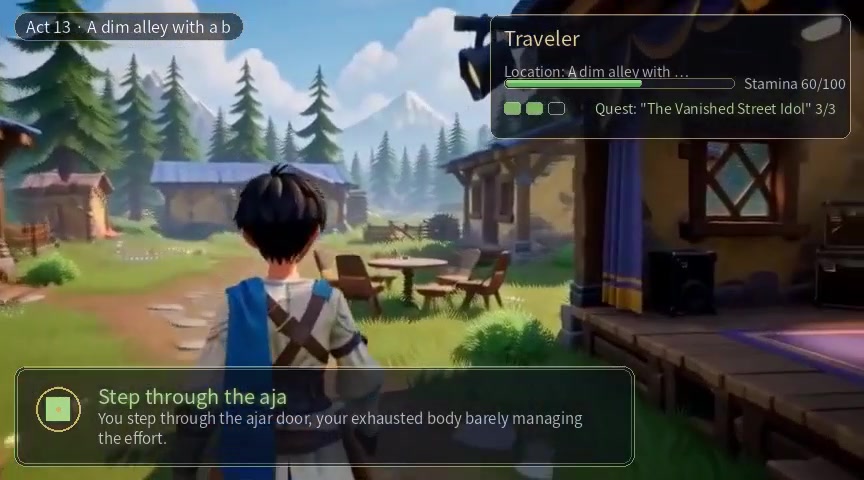}{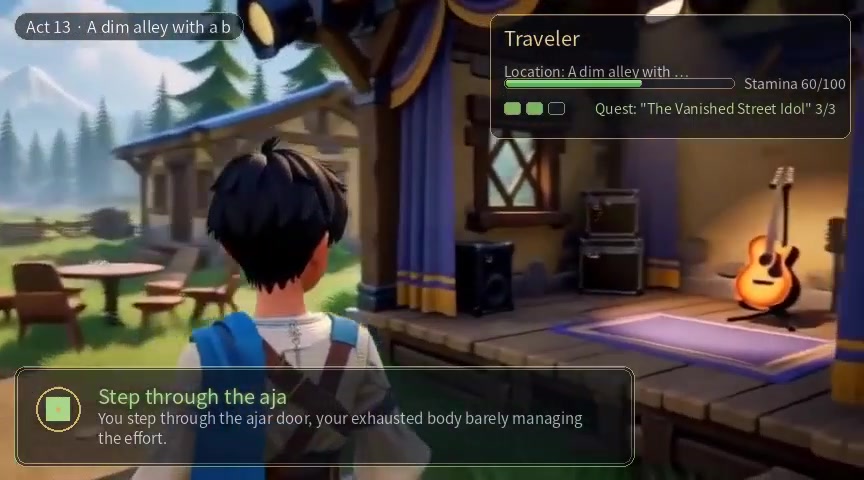}
\stcell{90}{-167.5}{act 14: backstage corridor}{stamina 48/100 $\cdot$ quest 3/3}{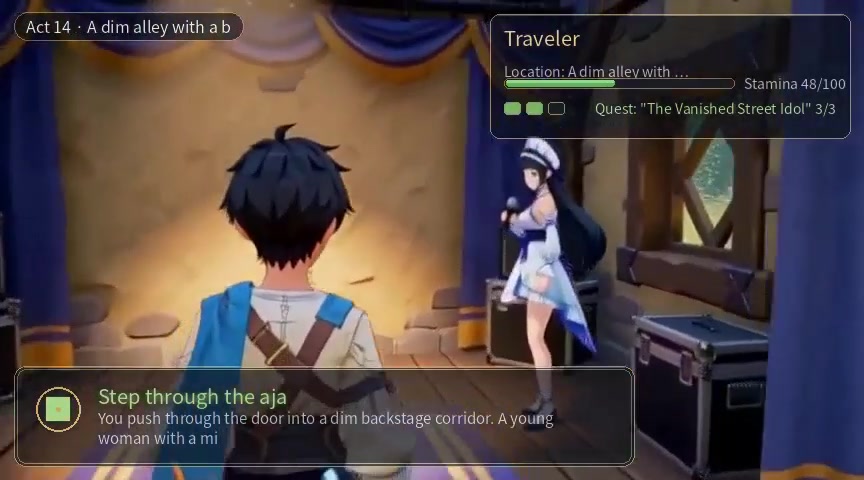}{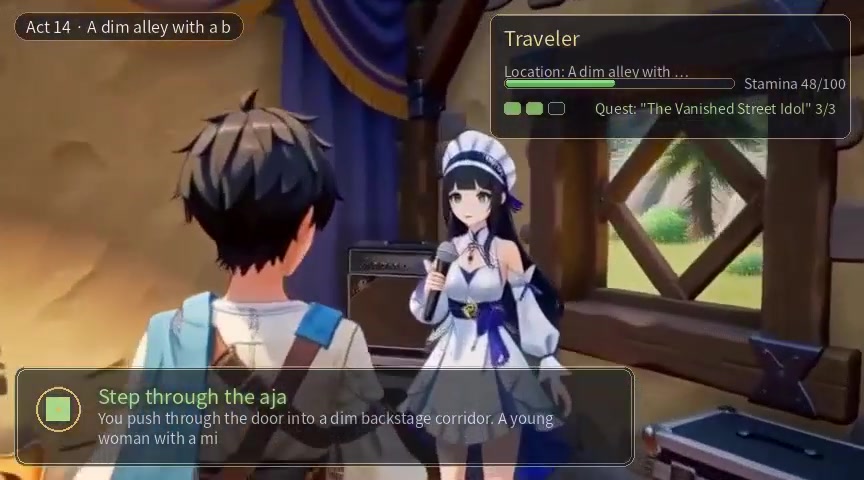}
\stcell{90}{-201}{act 15: following the woman}{stamina 36/100 $\cdot$ quest 3/3}{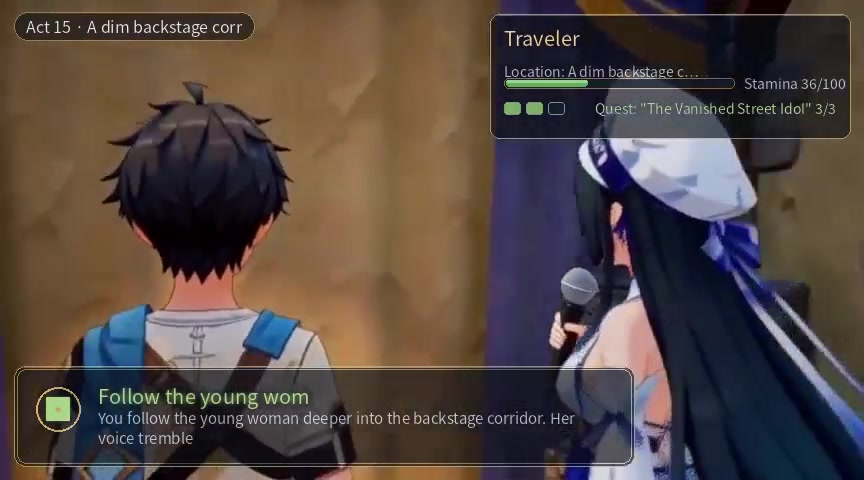}{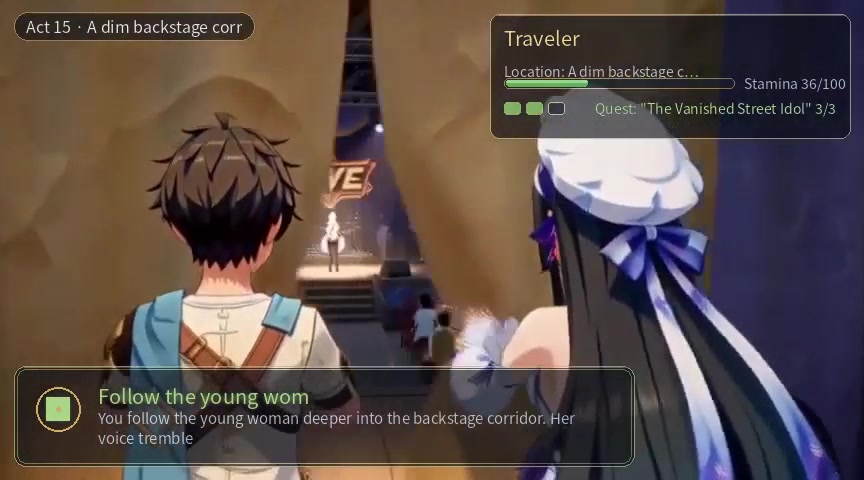}
\stcell{90}{-234.5}{the end}{curtain call; the episode ends}{vea.jpg}{veb.jpg}
\end{tikzpicture}}
\caption{The full fifteen-act town-exploration episode of Figure~\ref{fig:demo} (Case A), released video of our harness on frozen MiniMax-H3.
Two frames per act (native 16:9), with headers giving the act and the Traveler HUD (stamina, status, quest progress) at the act's first frame.
Stamina drains 28$\to$0 on the walk, the status line switches to ``Exhausted --- slow walk only'', and rest restores it to 60. A second drain-and-recovery cycle follows before the backstage door, while the quest counter advances only forward, 1/3 to 2/3 to 3/3.}
\label{fig:stamepisode}
\end{figure}

\begin{figure}[p]
\centering
\resizebox{0.98\textwidth}{!}{%
\begin{tikzpicture}[x=1mm, y=1mm]
\newcommand{\lvcell}[6]{%
  \node[align=center] at ({#1+36.5},{#2-2.6}) {{\scriptsize\bfseries #3}\\{\scriptsize\itshape #4}};
  \node[draw=black!30, line width=0.4, inner sep=0pt, anchor=north west] at (#1,{#2-6.8})
    {\includegraphics[width=36mm,height=20.25mm]{figures/assets/demo/love/#5}\hspace{1mm}%
\includegraphics[width=36mm,height=20.25mm]{figures/assets/demo/love/#6}};}
\lvcell{0}{0}{turn 1 $\cdot$ corner caf\'e}{affection 47/100 $\cdot$ friends $\cdot$ shy}{t01a.jpg}{t01b.jpg}
\lvcell{0}{-30.5}{turn 2 $\cdot$ corner caf\'e}{affection 48/100 $\cdot$ friends $\cdot$ happy}{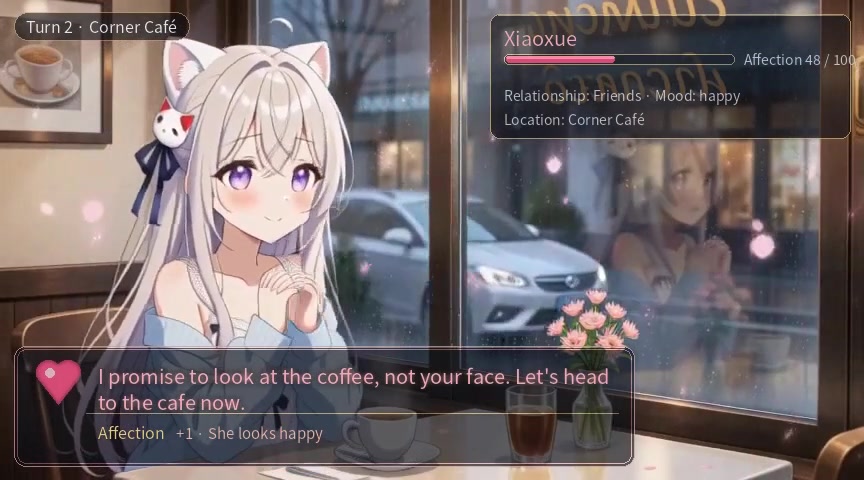}{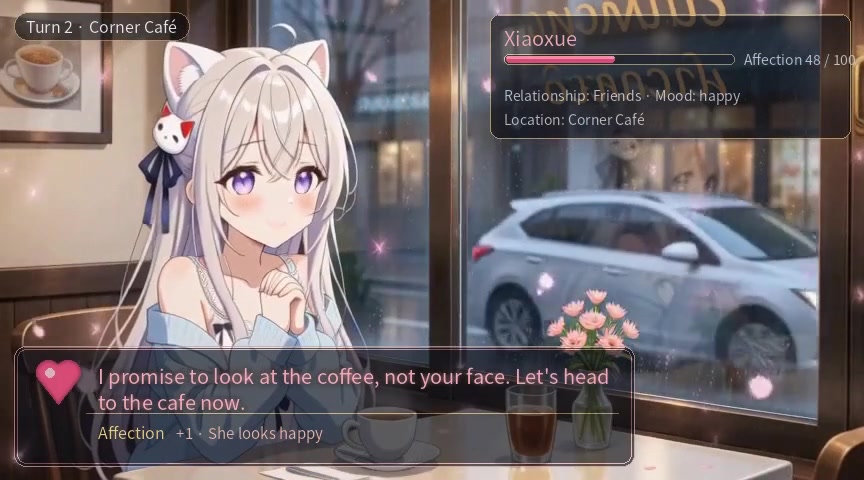}
\lvcell{0}{-61}{turn 3 $\cdot$ corner caf\'e}{affection 49/100 $\cdot$ friends $\cdot$ happy}{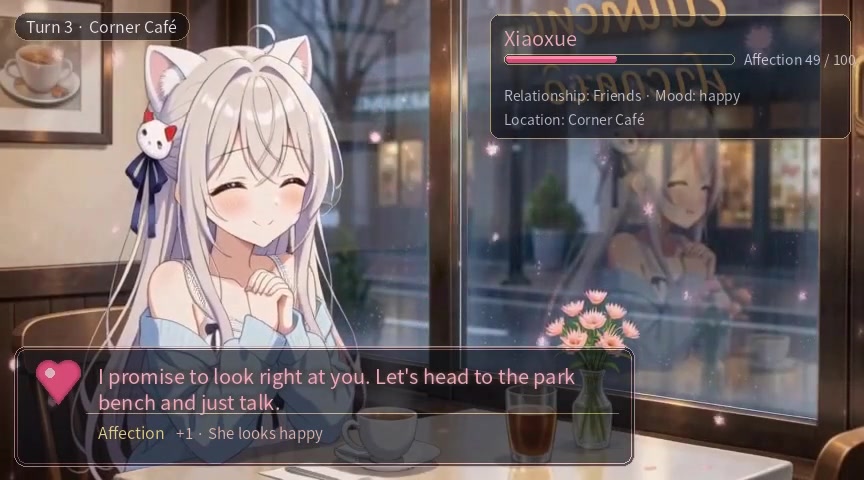}{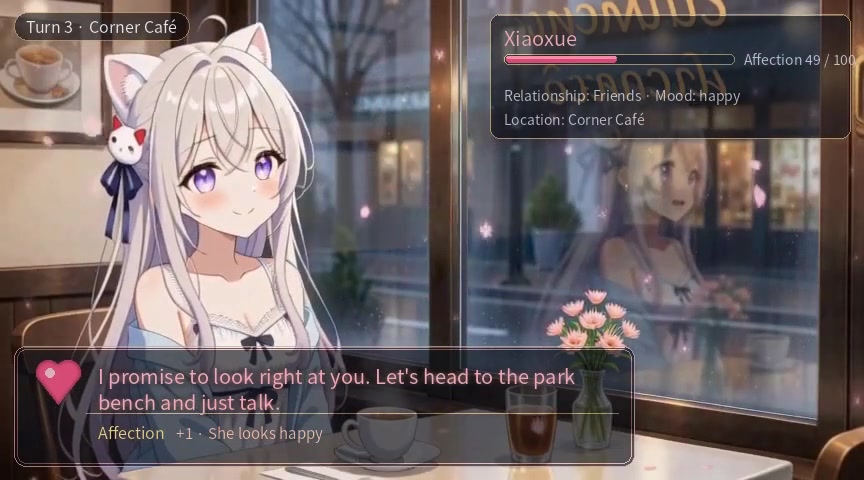}
\lvcell{0}{-91.5}{turn 4 $\cdot$ corner caf\'e}{affection 50/100 $\cdot$ friends $\cdot$ happy}{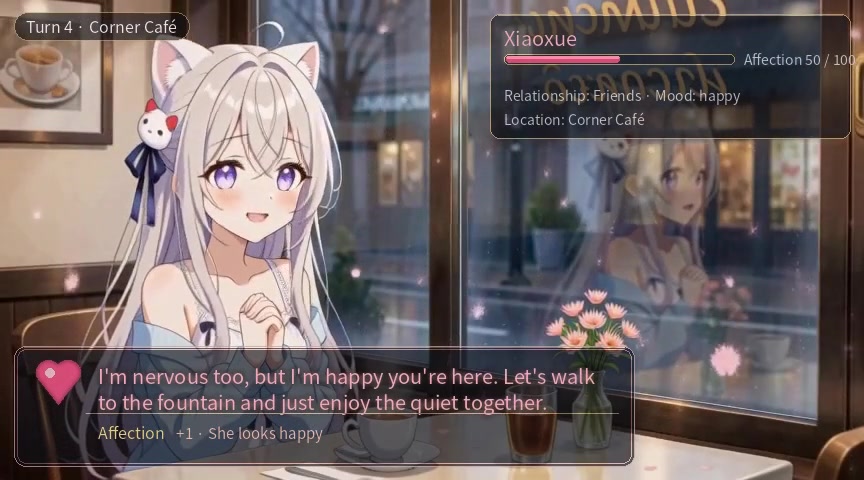}{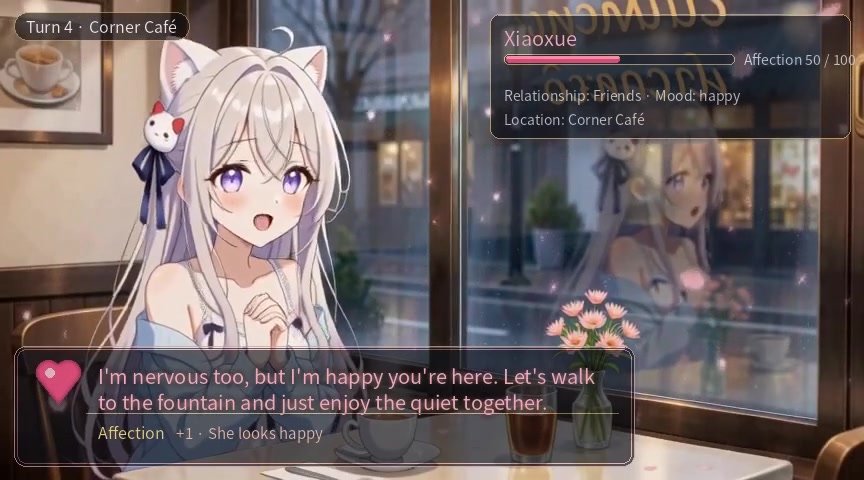}
\lvcell{0}{-122}{turn 5 $\cdot$ old library}{affection 51/100 $\cdot$ friends $\cdot$ happy}{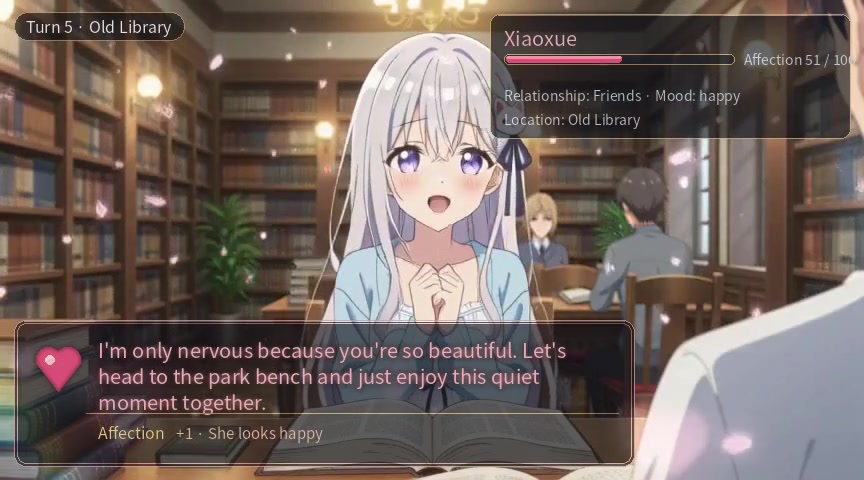}{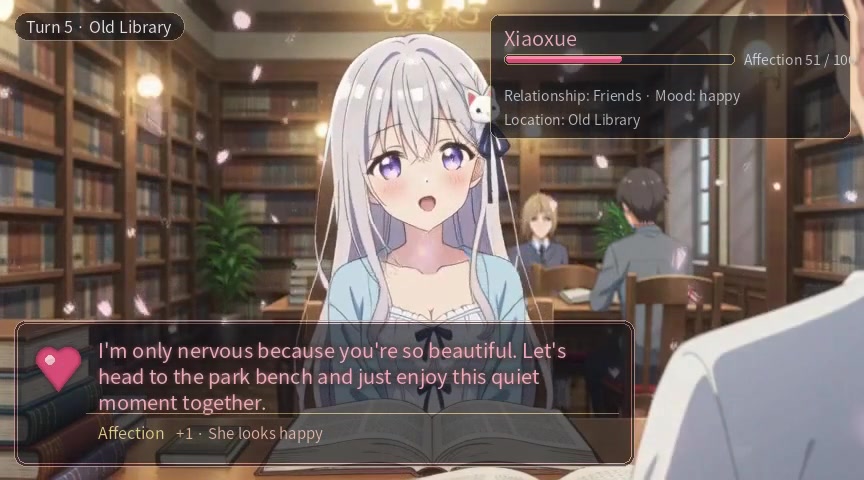}
\lvcell{0}{-152.5}{turn 6 $\cdot$ old library}{affection 52/100 $\cdot$ friends $\cdot$ happy}{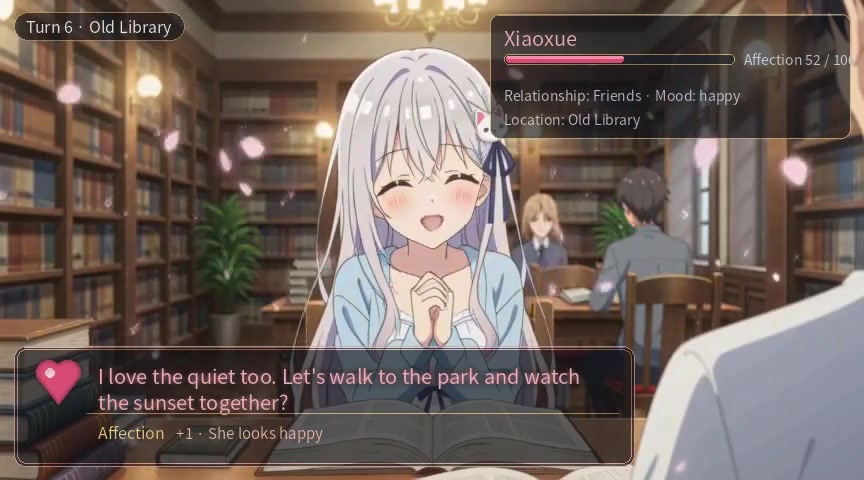}{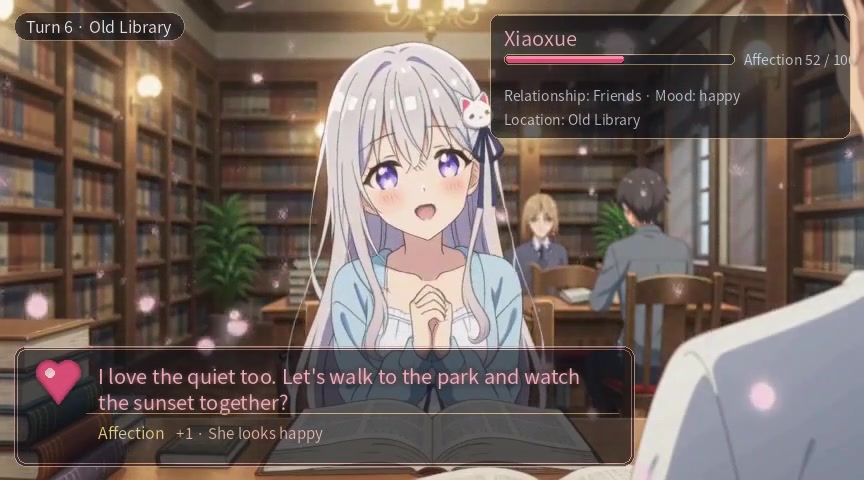}
\lvcell{0}{-183}{turn 7 $\cdot$ old library}{affection 53/100 $\cdot$ friends $\cdot$ happy}{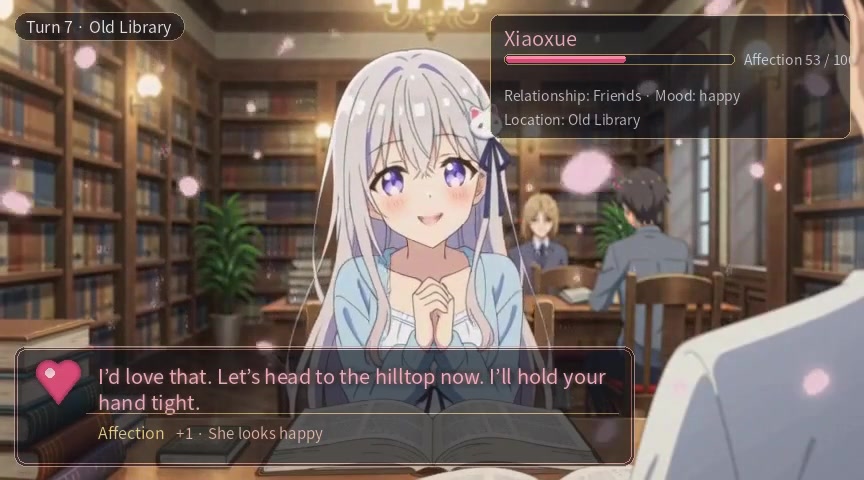}{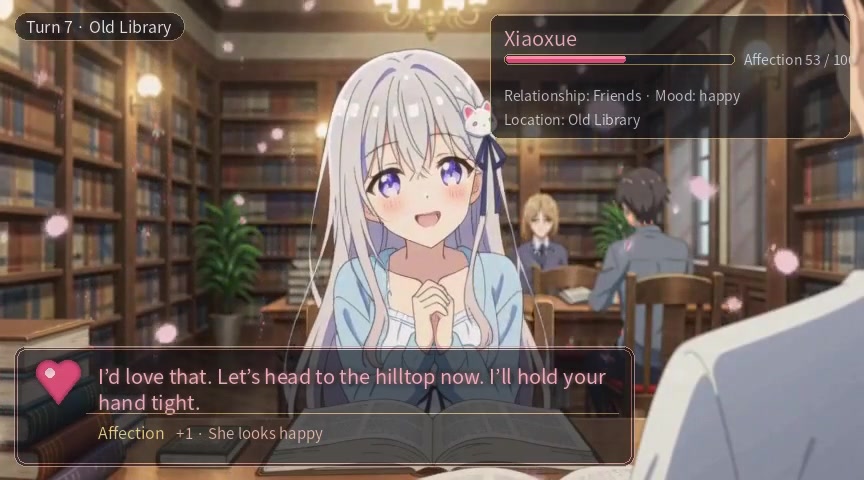}
\lvcell{0}{-213.5}{turn 8 $\cdot$ old library}{affection 54/100 $\cdot$ friends $\cdot$ happy}{t08a.jpg}{t08b.jpg}
\lvcell{0}{-244}{turn 9 $\cdot$ old library}{affection 55/100 $\cdot$ friends $\cdot$ happy}{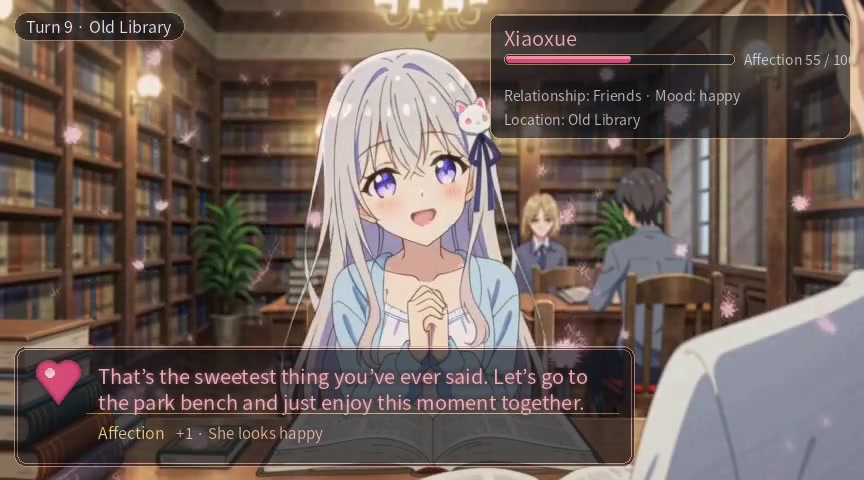}{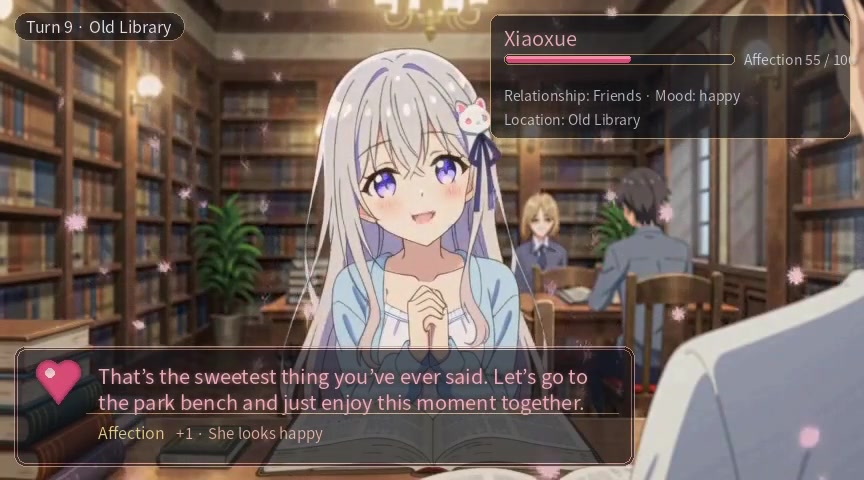}
\lvcell{0}{-274.5}{turn 10 $\cdot$ neon platform}{affection 56/100 $\cdot$ friends $\cdot$ happy}{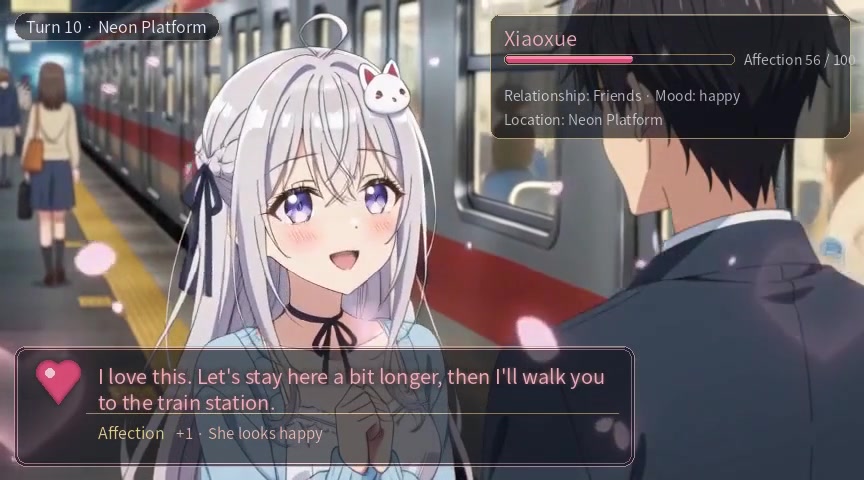}{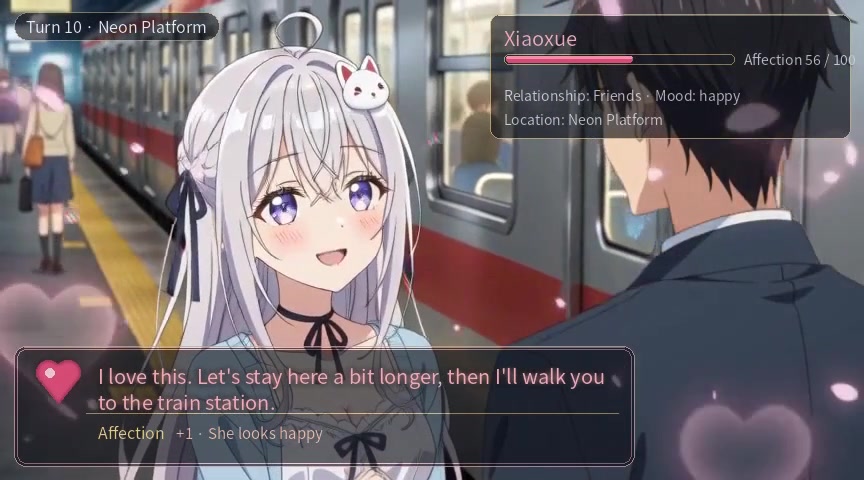}
\lvcell{78}{0}{turn 11 $\cdot$ neon platform}{affection 57/100 $\cdot$ friends $\cdot$ happy}{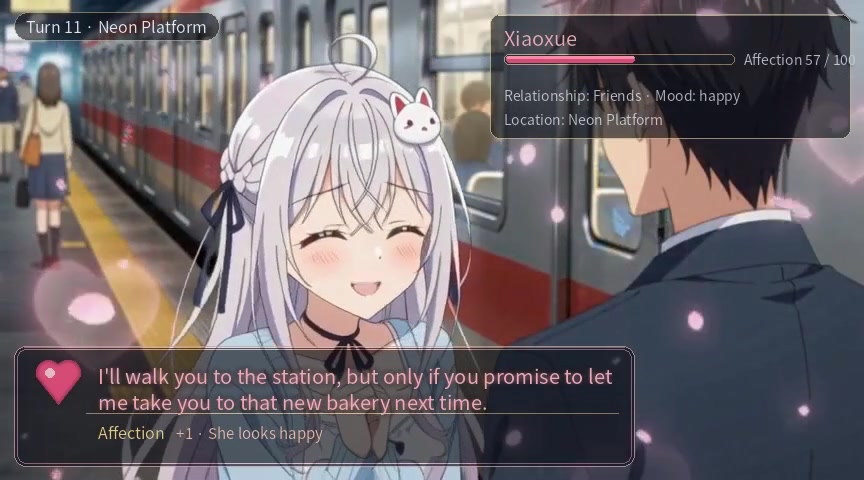}{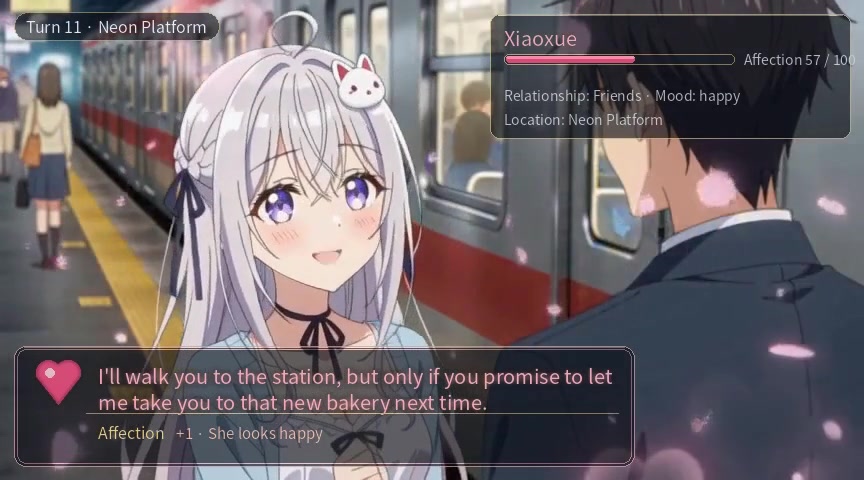}
\lvcell{78}{-30.5}{turn 12 $\cdot$ neon platform}{affection 58/100 $\cdot$ friends $\cdot$ happy}{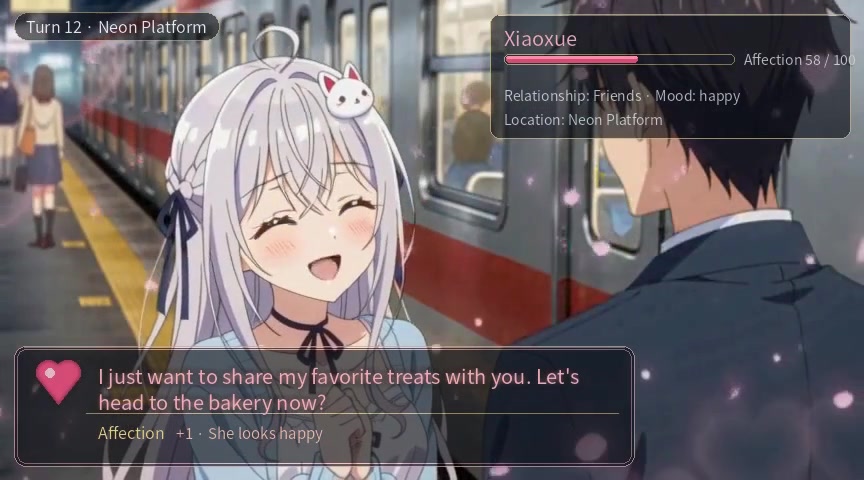}{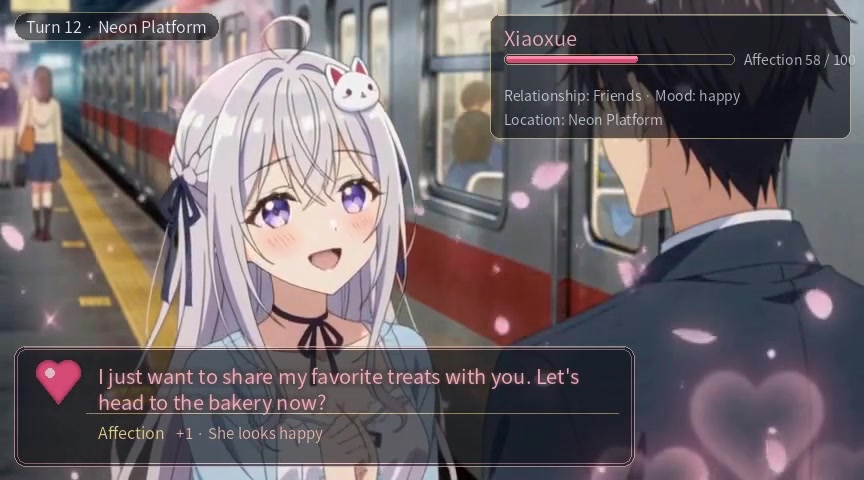}
\lvcell{78}{-61}{turn 13 $\cdot$ neon platform}{affection 59/100 $\cdot$ friends $\cdot$ happy}{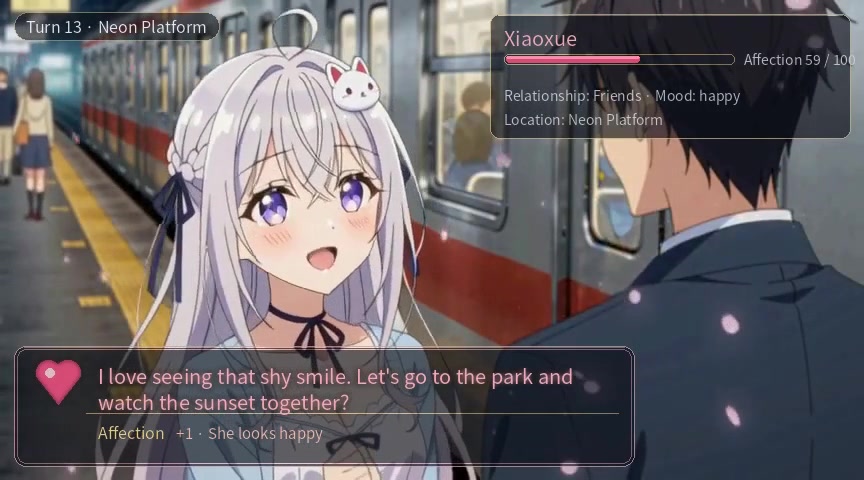}{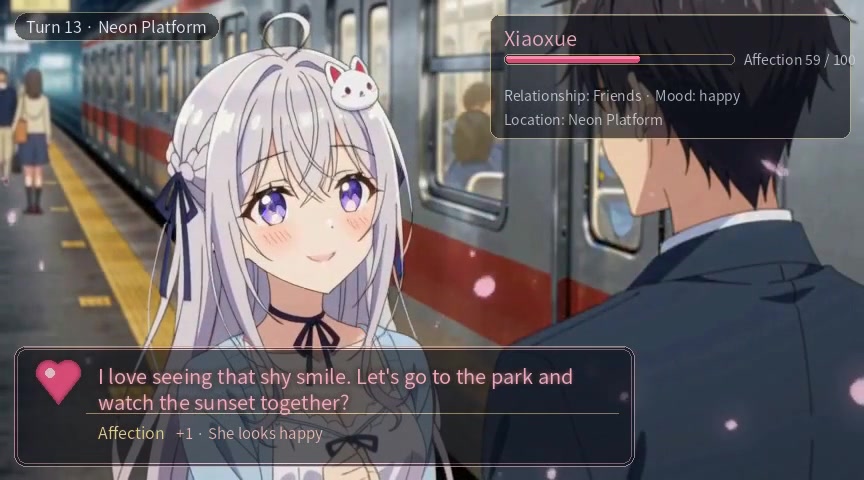}
\lvcell{78}{-91.5}{turn 14 $\cdot$ neon platform}{affection 60/100 $\cdot$ flirting $\cdot$ happy}{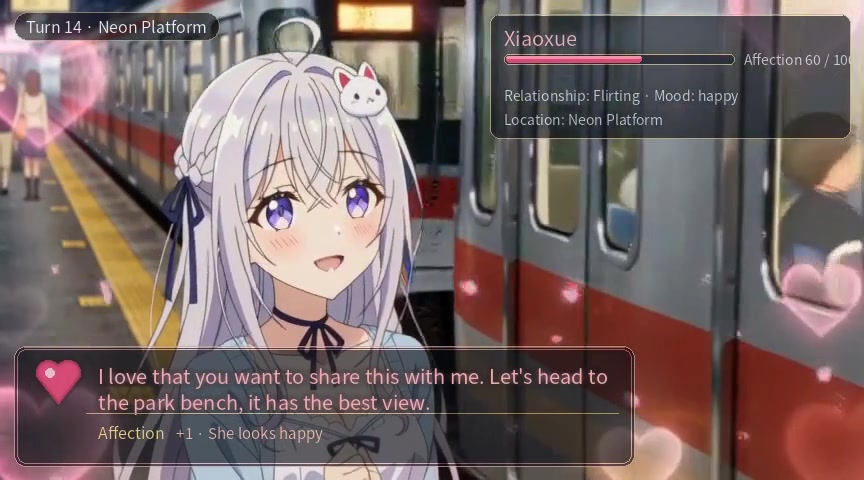}{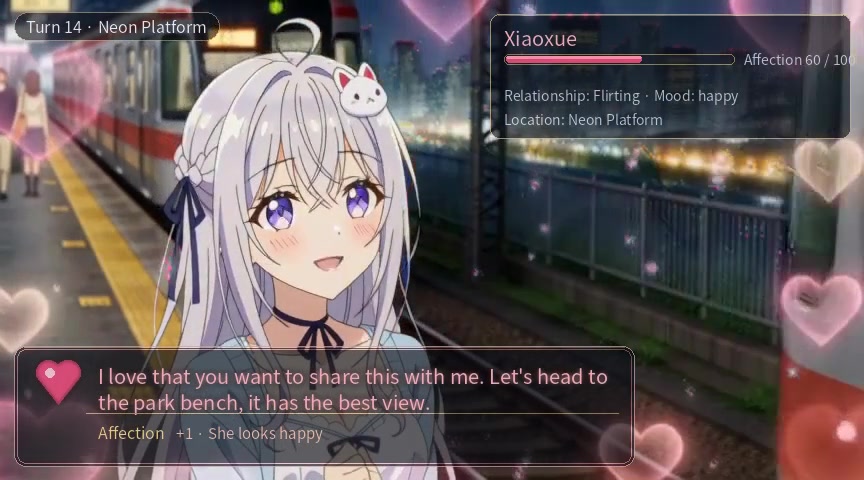}
\lvcell{78}{-122}{turn 15 $\cdot$ neon platform}{affection 61/100 $\cdot$ flirting $\cdot$ happy}{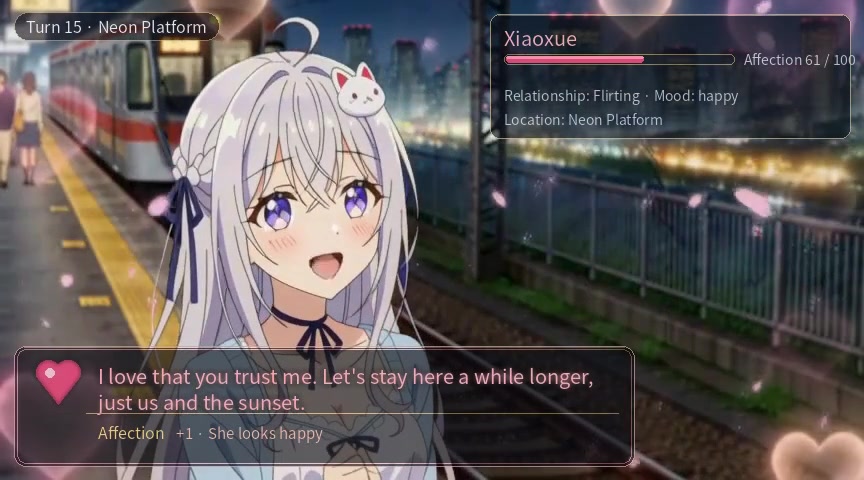}{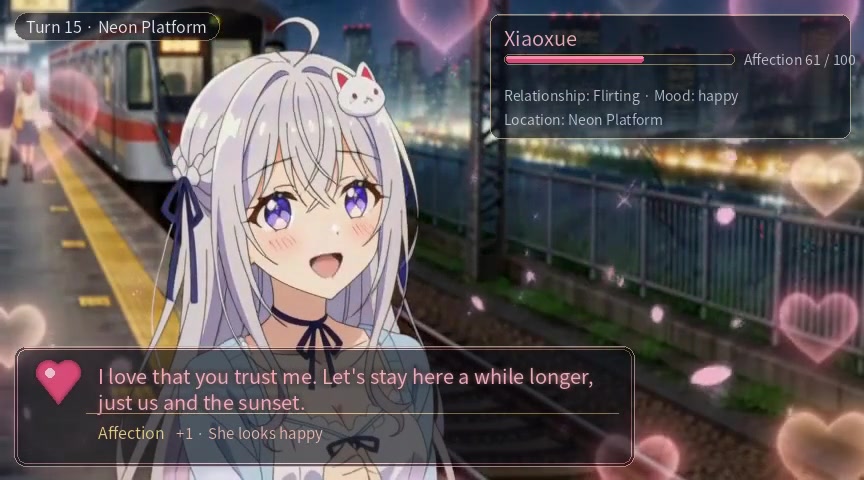}
\lvcell{78}{-152.5}{turn 16 $\cdot$ neon platform}{affection 62/100 $\cdot$ flirting $\cdot$ happy}{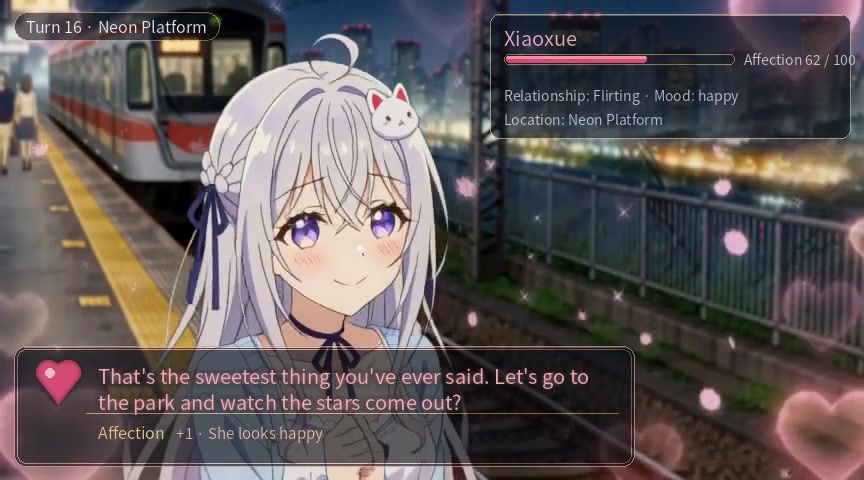}{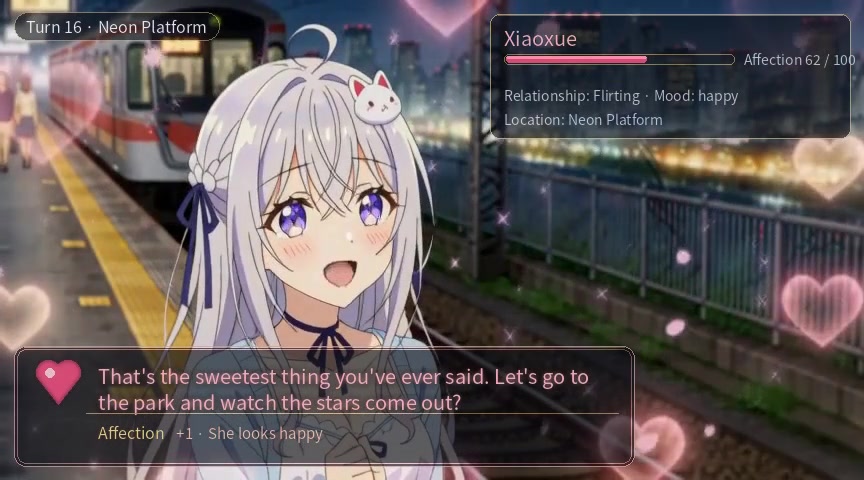}
\lvcell{78}{-183}{turn 17 $\cdot$ neon platform}{affection 63/100 $\cdot$ flirting $\cdot$ happy}{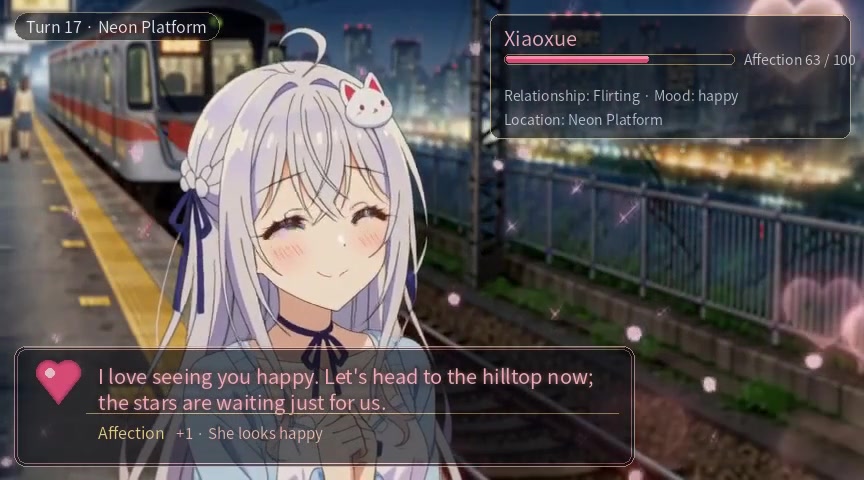}{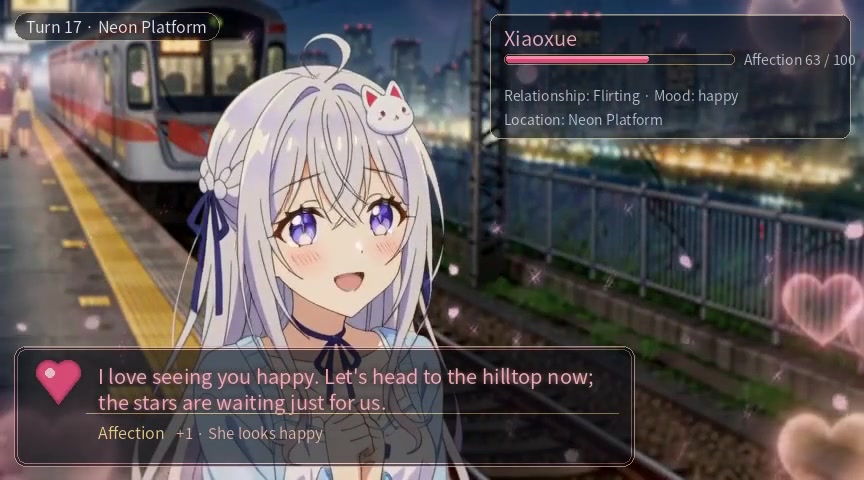}
\lvcell{78}{-213.5}{turn 18 $\cdot$ neon platform}{affection 64/100 $\cdot$ flirting $\cdot$ happy}{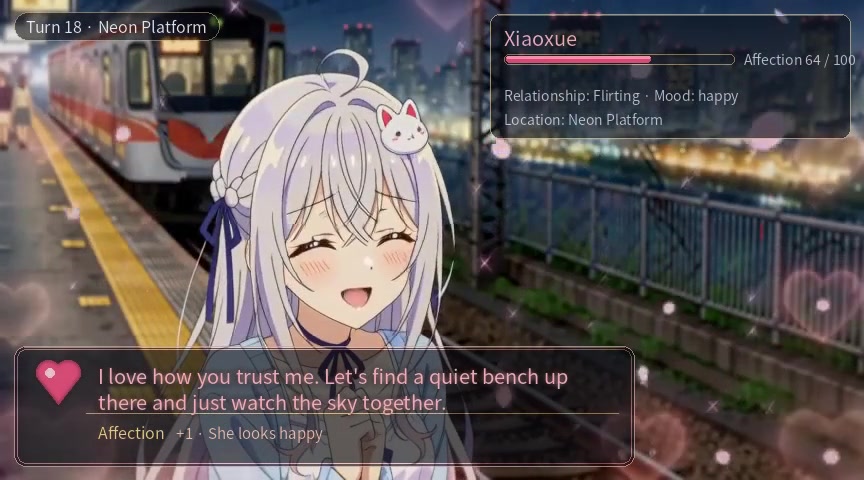}{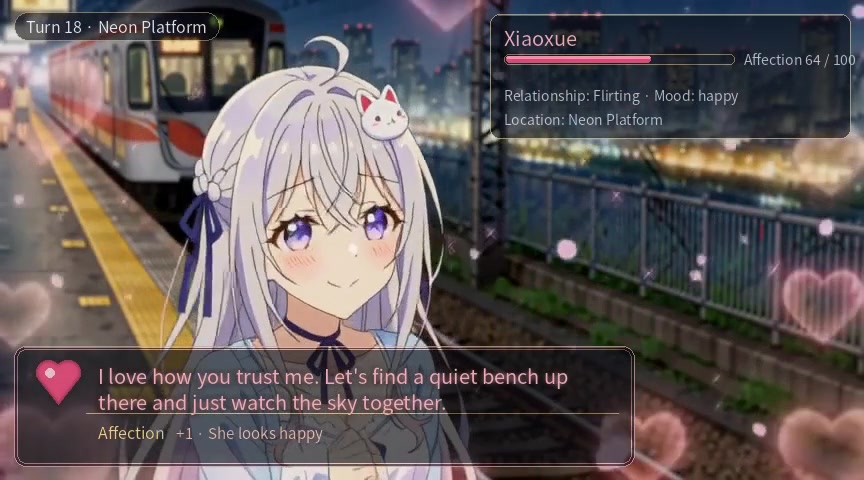}
\lvcell{78}{-244}{turn 19 $\cdot$ neon platform}{affection 65/100 $\cdot$ flirting $\cdot$ happy}{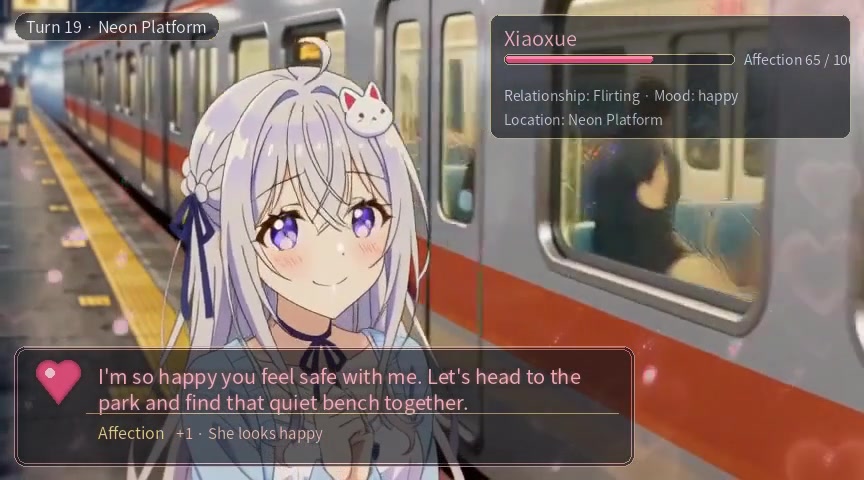}{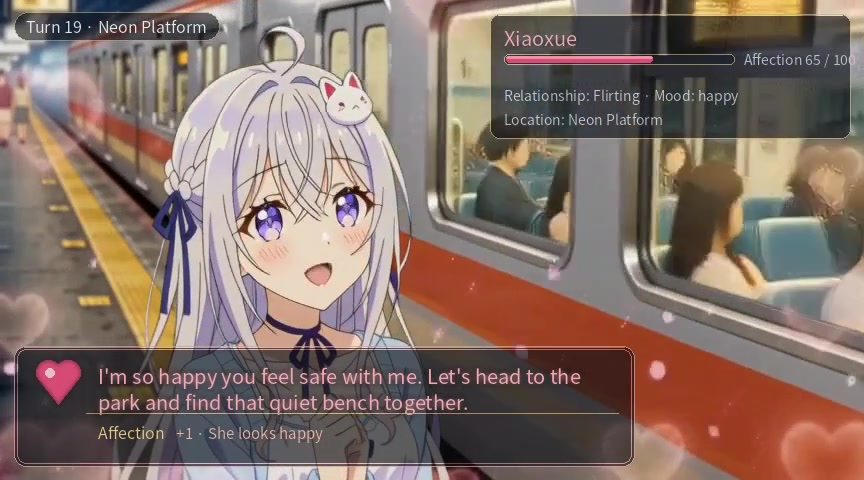}
\lvcell{78}{-274.5}{turn 20 $\cdot$ neon platform}{affection 66/100 $\cdot$ flirting $\cdot$ happy}{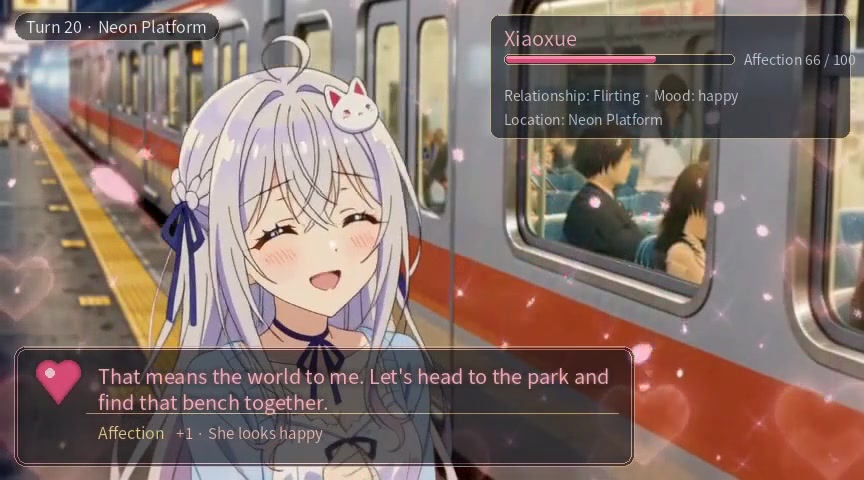}{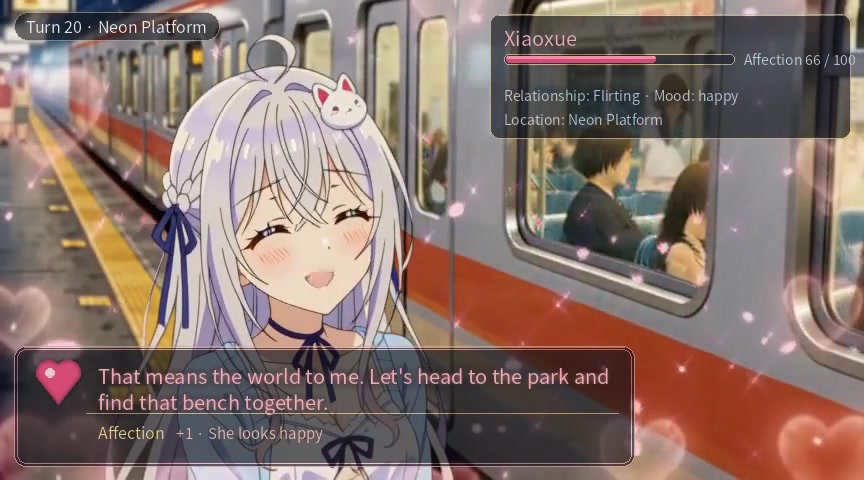}
\lvcell{156}{0}{turn 21 $\cdot$ neon platform}{affection 67/100 $\cdot$ flirting $\cdot$ happy}{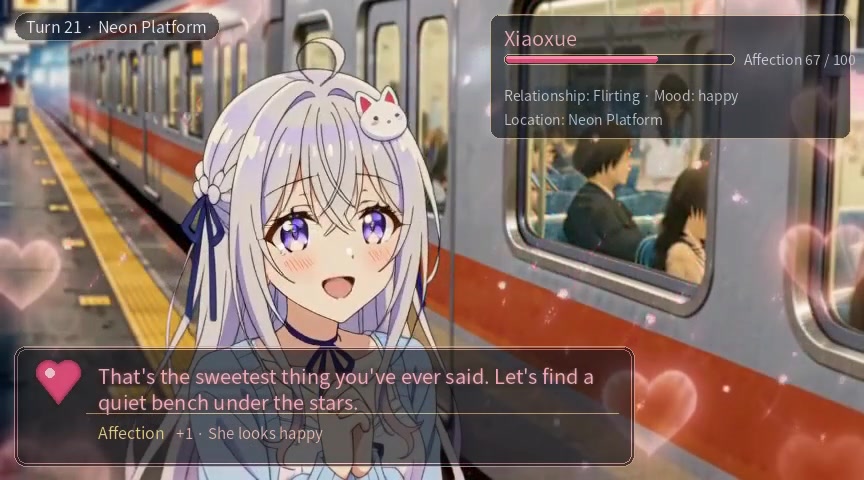}{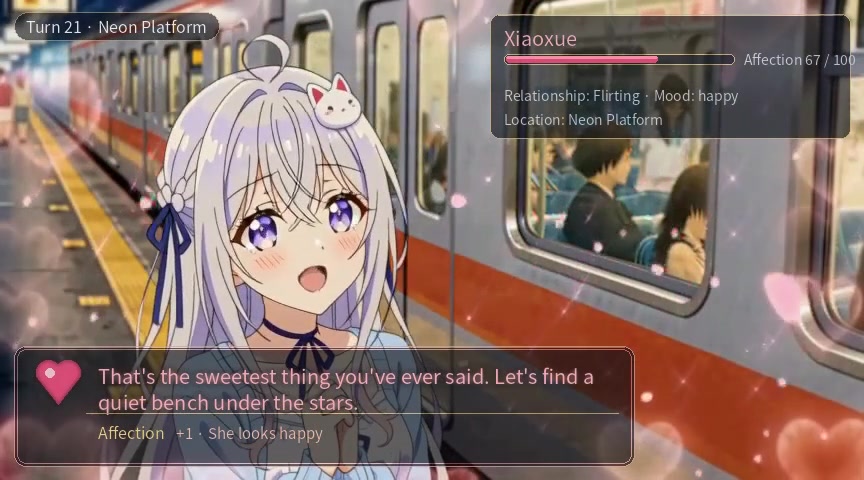}
\lvcell{156}{-30.5}{turn 22 $\cdot$ neon platform}{affection 68/100 $\cdot$ flirting $\cdot$ happy}{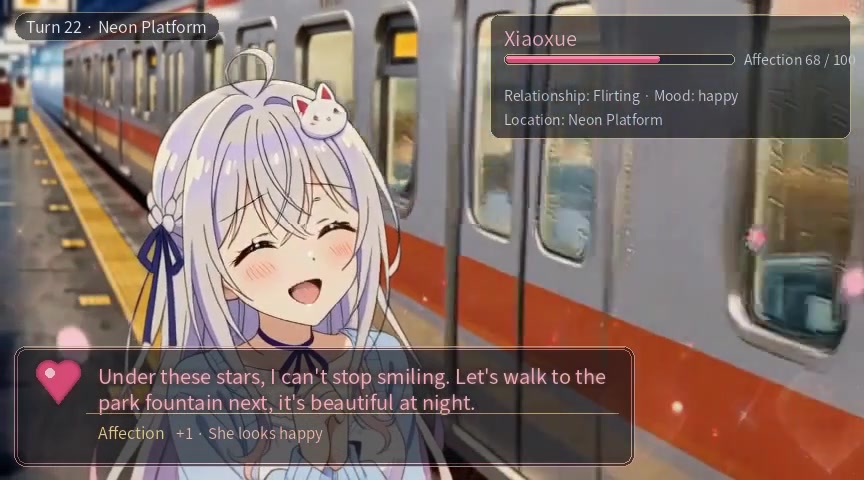}{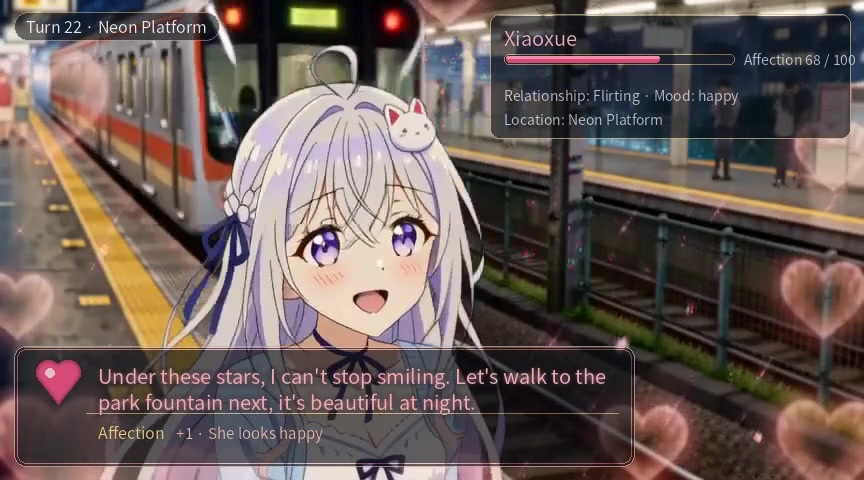}
\lvcell{156}{-61}{turn 23 $\cdot$ neon platform}{affection 69/100 $\cdot$ flirting $\cdot$ happy}{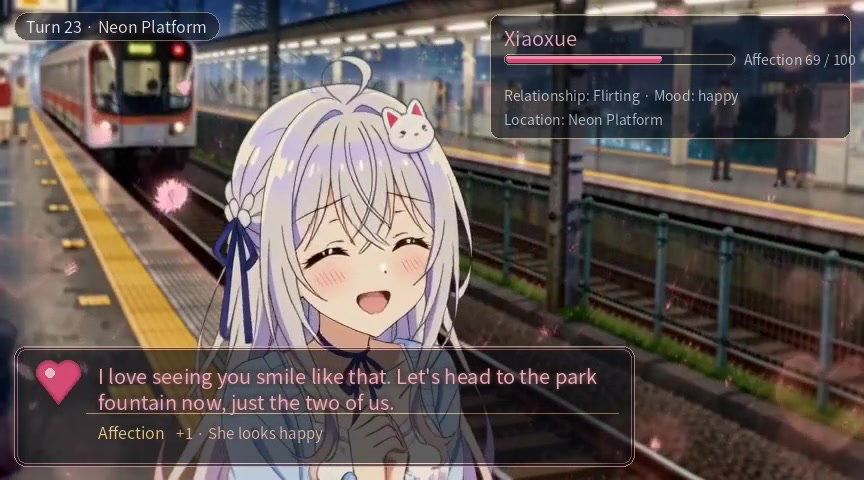}{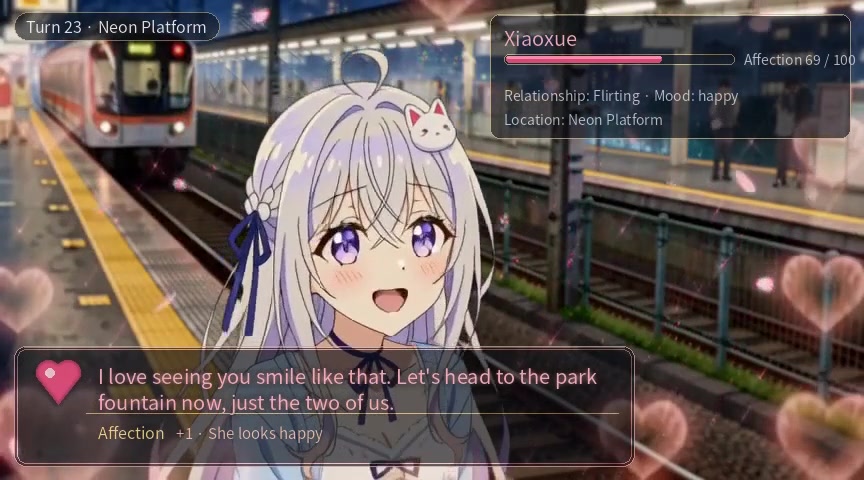}
\lvcell{156}{-91.5}{turn 24 $\cdot$ neon platform}{affection 70/100 $\cdot$ flirting $\cdot$ happy}{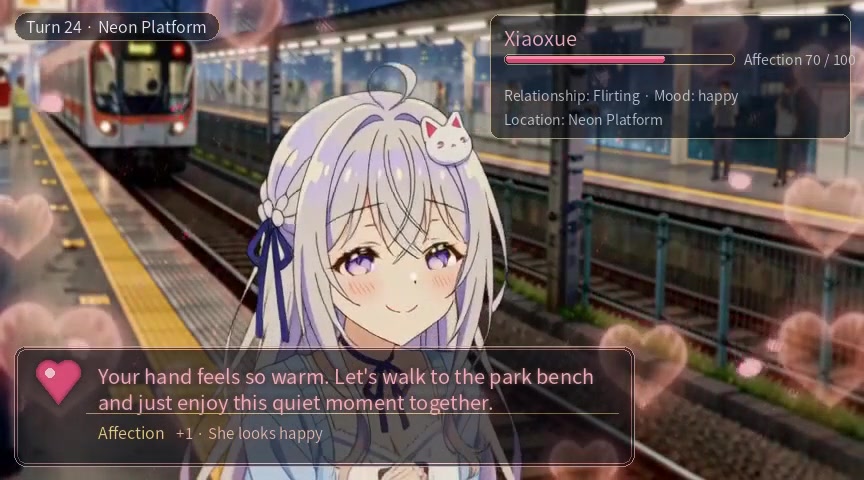}{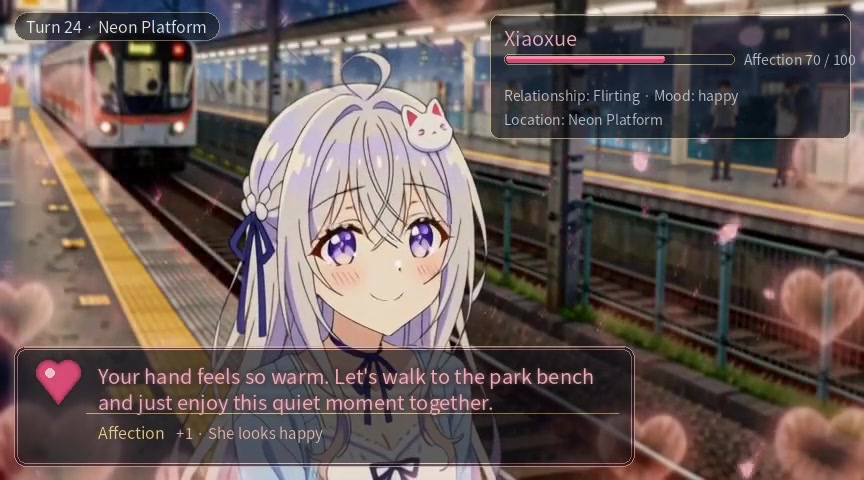}
\lvcell{156}{-122}{turn 25 $\cdot$ neon platform}{affection 71/100 $\cdot$ flirting $\cdot$ happy}{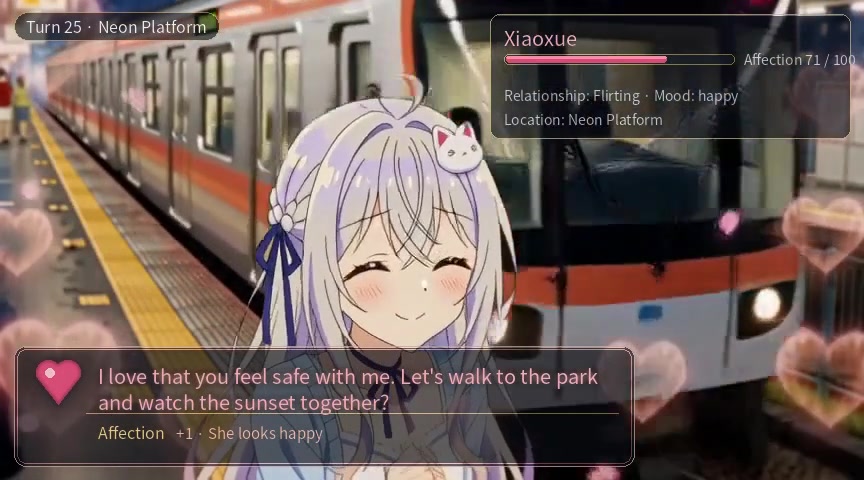}{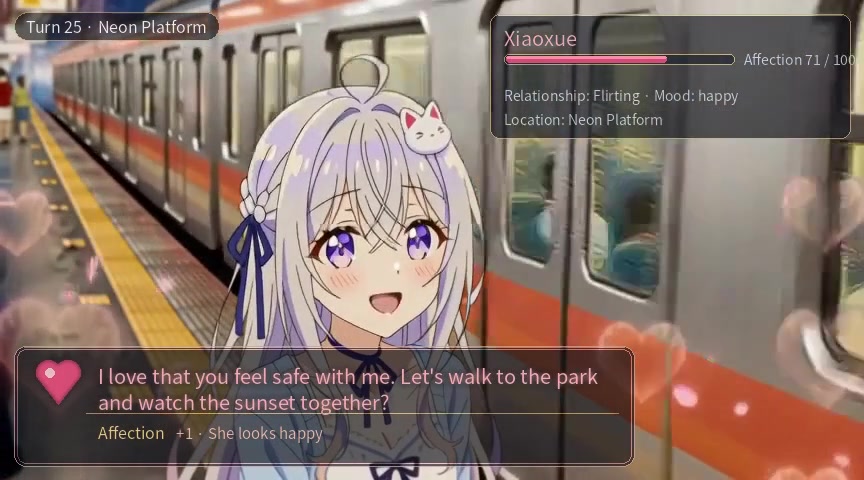}
\lvcell{156}{-152.5}{turn 26 $\cdot$ neon platform}{affection 72/100 $\cdot$ flirting $\cdot$ happy}{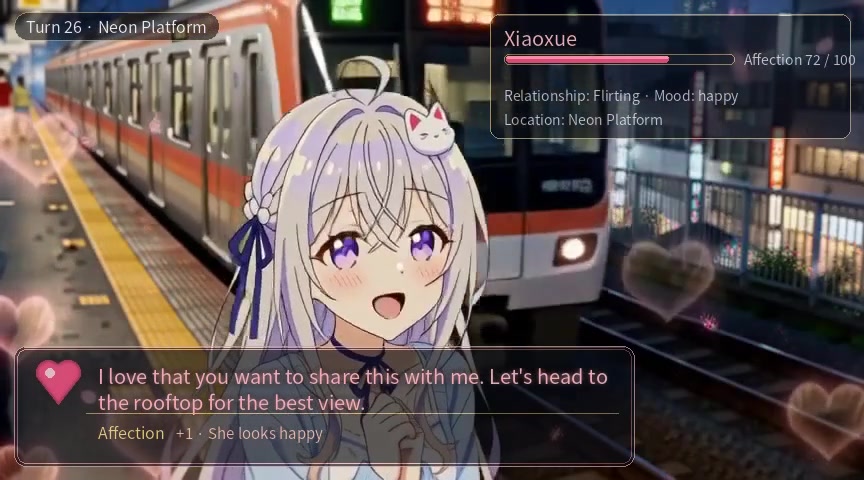}{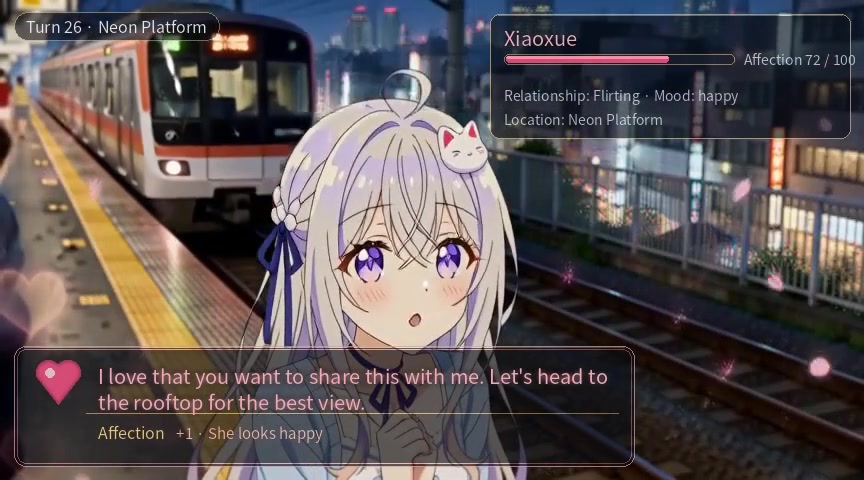}
\lvcell{156}{-183}{turn 27 $\cdot$ neon platform}{affection 73/100 $\cdot$ flirting $\cdot$ happy}{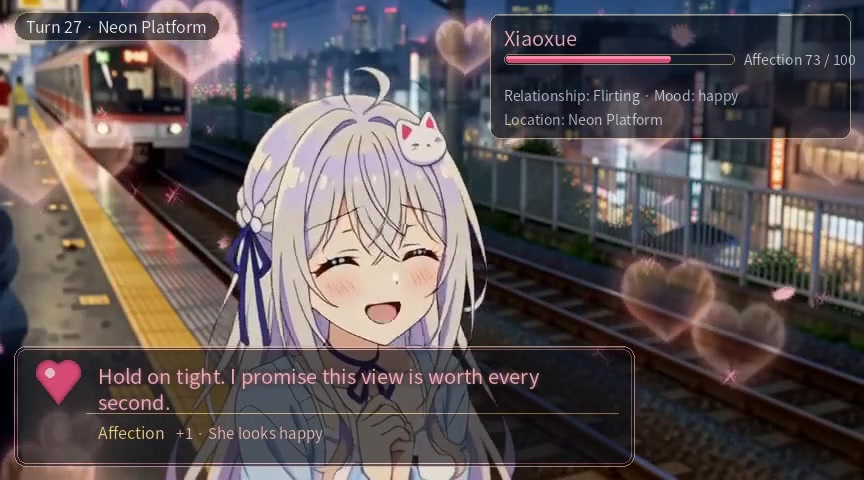}{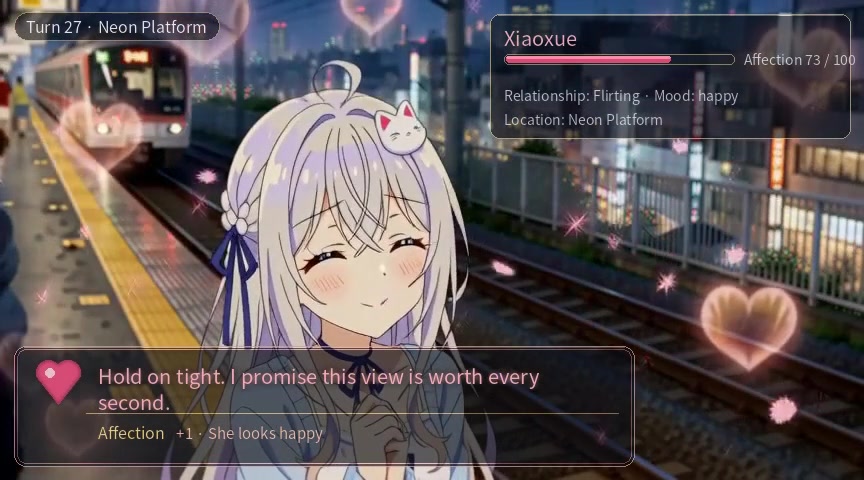}
\lvcell{156}{-213.5}{turn 28 $\cdot$ neon platform}{affection 74/100 $\cdot$ flirting $\cdot$ happy}{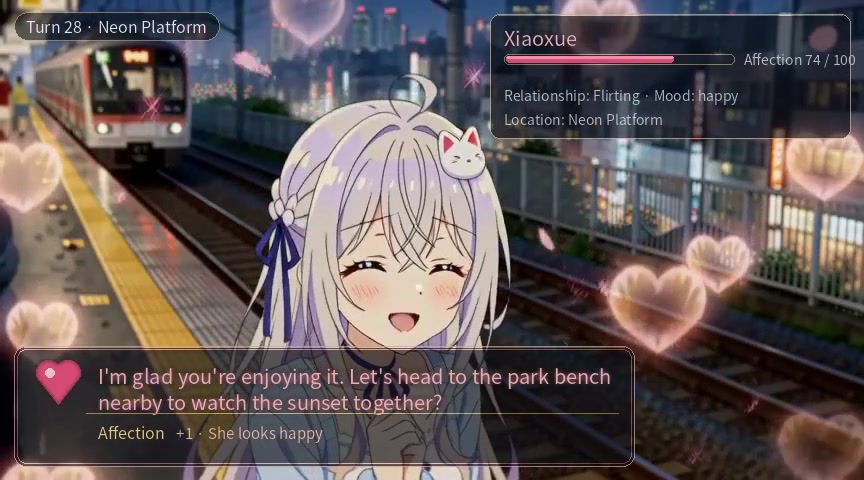}{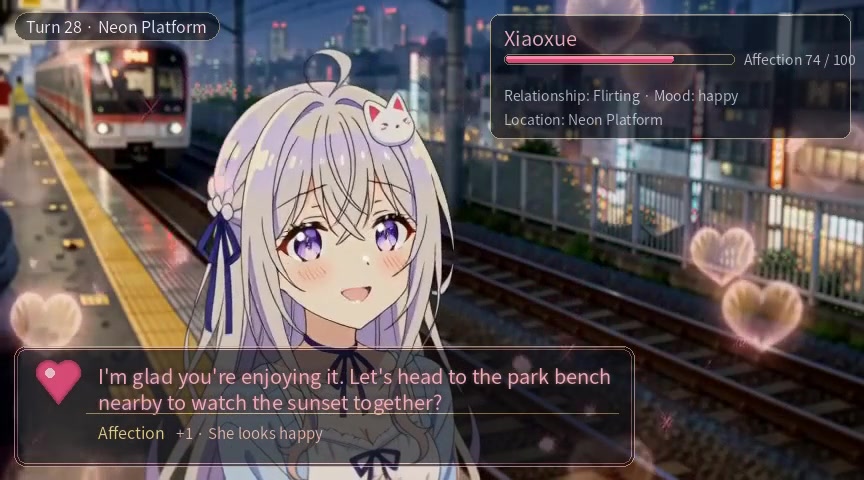}
\lvcell{156}{-244}{turn 29 $\cdot$ neon platform}{affection 75/100 $\cdot$ lovers $\cdot$ happy}{t29a.jpg}{t29b.jpg}
\lvcell{156}{-274.5}{confession accepted}{the confession is accepted; the episode ends}{vea.jpg}{veb.jpg}
\end{tikzpicture}}
\caption{The full twenty-nine-turn date episode of Figure~\ref{fig:demo} (Case B), released video of our harness on frozen MiniMax-H3.
Two frames per turn (native 16:9), with headers giving the turn, scene, and HUD state (affection, relationship, mood) at the turn's first frame.
Affection rises monotonically 47$\to$75, one point per turn. The relationship state flips friends$\to$flirting at turn 14 and flirting$\to$lovers at turn 29, and the closing confession is accepted only under this accumulated state.}
\label{fig:loveepisode}
\end{figure}

\begin{figure}[p]
\centering
\resizebox{0.98\textwidth}{!}{%
\begin{tikzpicture}[x=1mm, y=1mm]
\newcommand{\epcell}[6]{%
  \node[align=center] at ({#1+42.5},{#2-2.6}) {{\scriptsize\bfseries #3}\\{\scriptsize\itshape #4}};
  \node[draw=black!30, line width=0.4, inner sep=0pt, anchor=north west] at (#1,{#2-6.8})
    {\includegraphics[width=42mm,height=23.33mm]{figures/assets/demo/#5}\hspace{1mm}%
\includegraphics[width=42mm,height=23.33mm]{figures/assets/demo/#6}};}
\epcell{0}{0}{turn 1: falling red sun}{arbiter 128/160 $\cdot$ eve 90/90, mana 8/15}{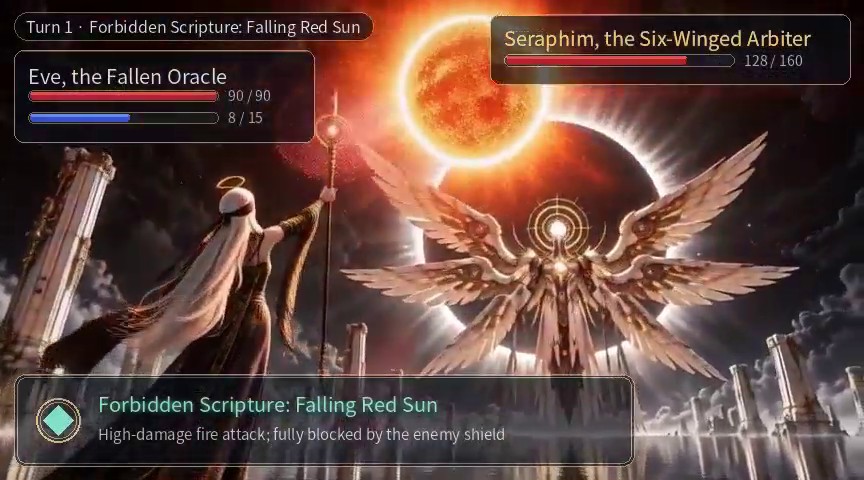}{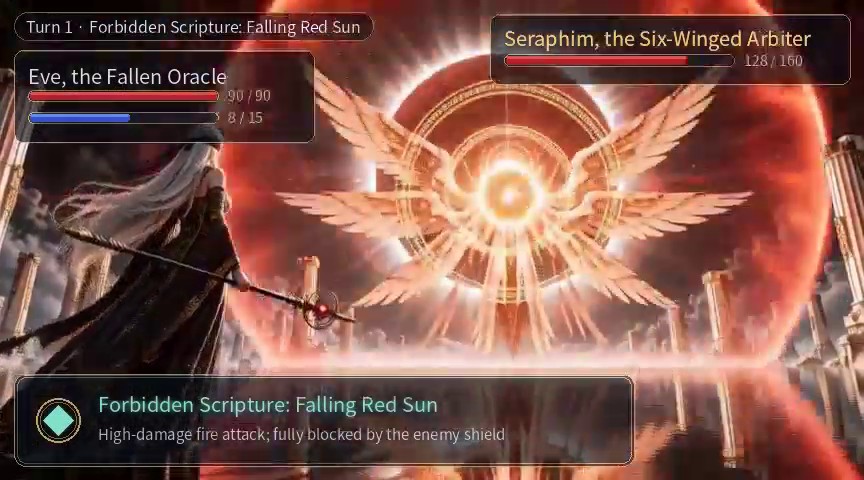}
\epcell{0}{-33.5}{turn 2: zero-degree stigma}{arbiter 112/160 $\cdot$ eve 90/90, mana 7/15}{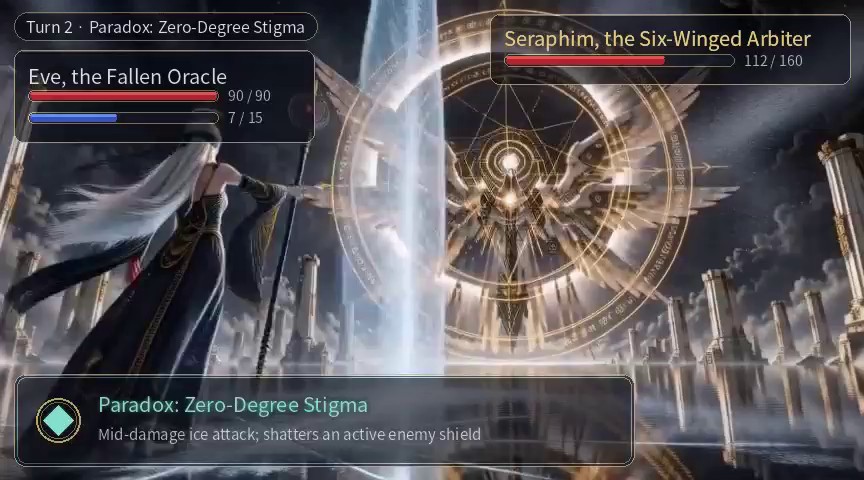}{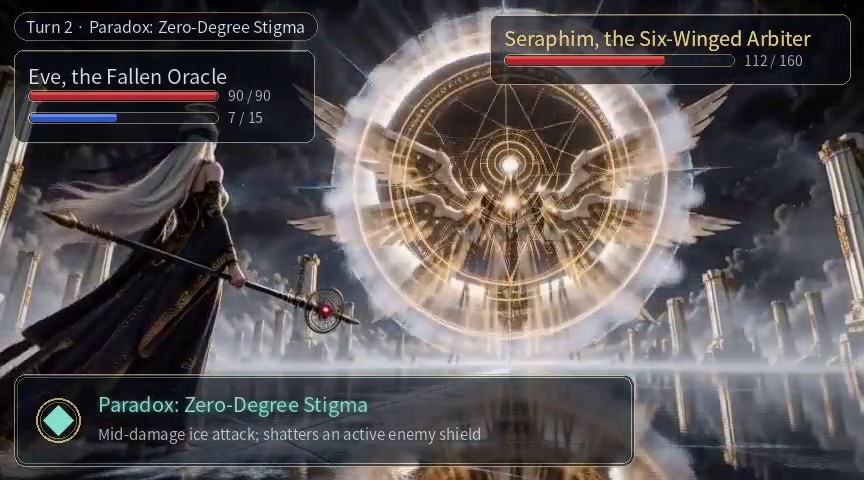}
\epcell{0}{-67}{turn 3: zero-degree stigma}{arbiter 100/160 $\cdot$ eve 76/90, mana 6/15}{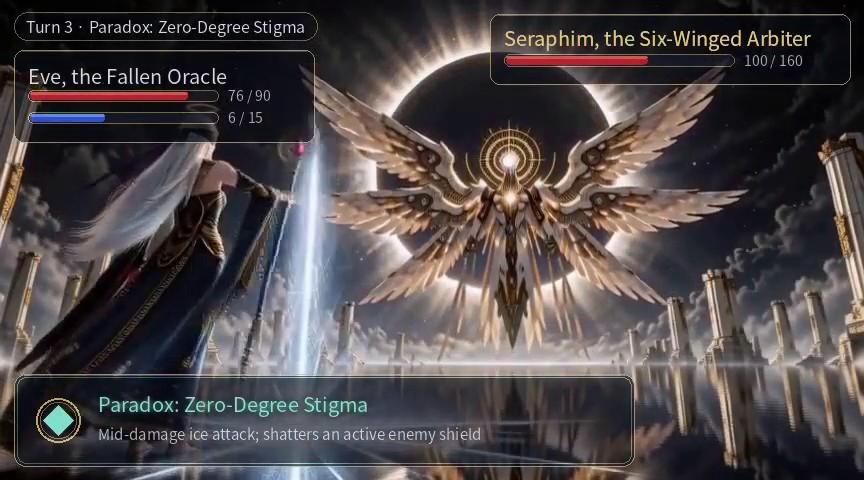}{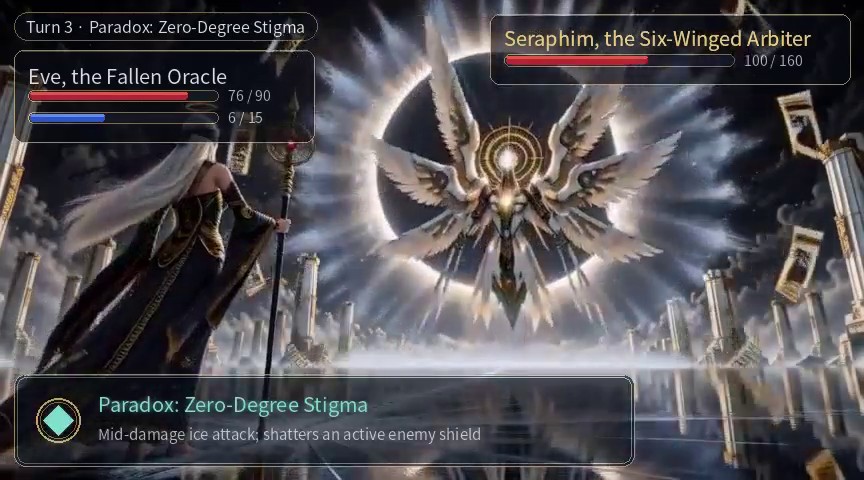}
\epcell{0}{-100.5}{turn 4: falling red sun}{arbiter 83/160 $\cdot$ eve 63/90, mana 4/15}{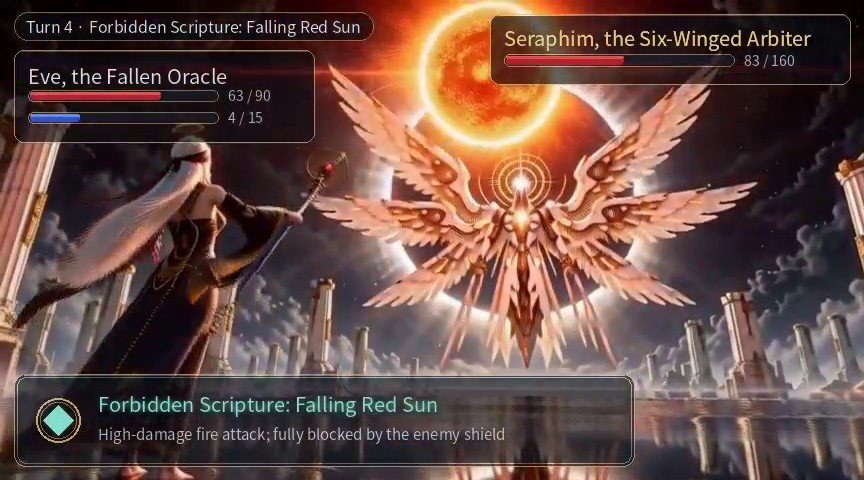}{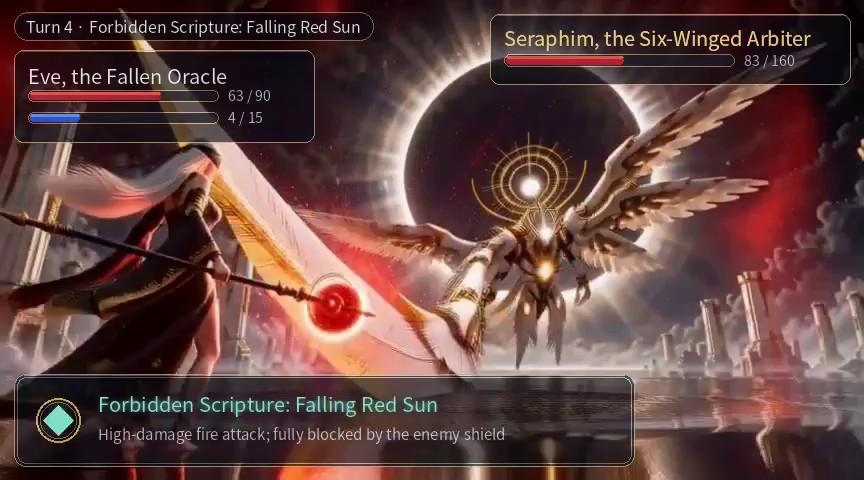}
\epcell{0}{-134}{turn 5: zero-degree stigma}{arbiter 69/160 $\cdot$ eve 63/90, mana 3/15}{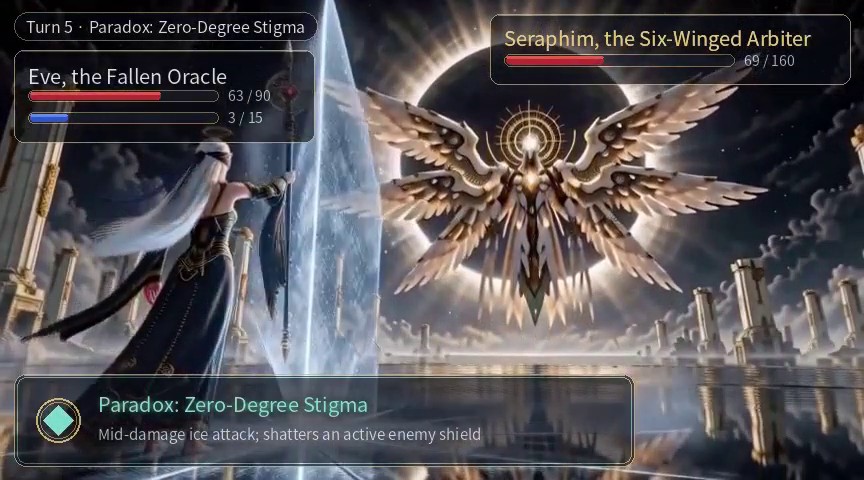}{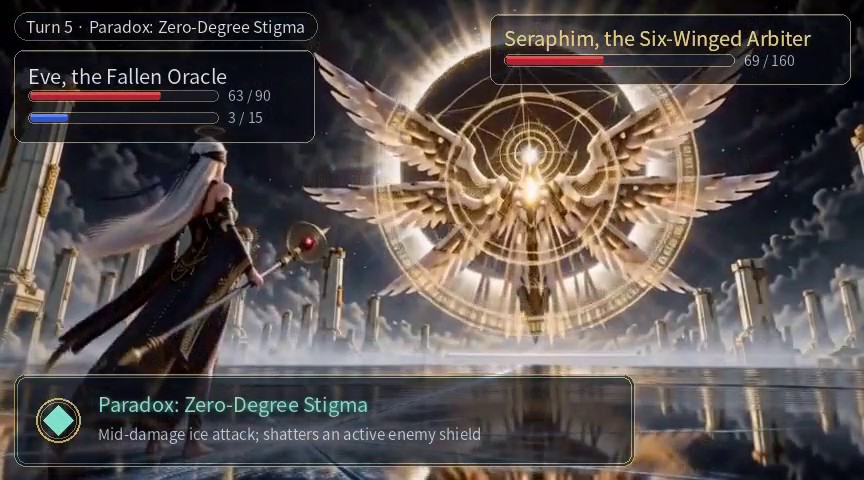}
\epcell{0}{-167.5}{turn 6: first sentence}{arbiter 59/160 $\cdot$ eve 63/90, mana 6/15}{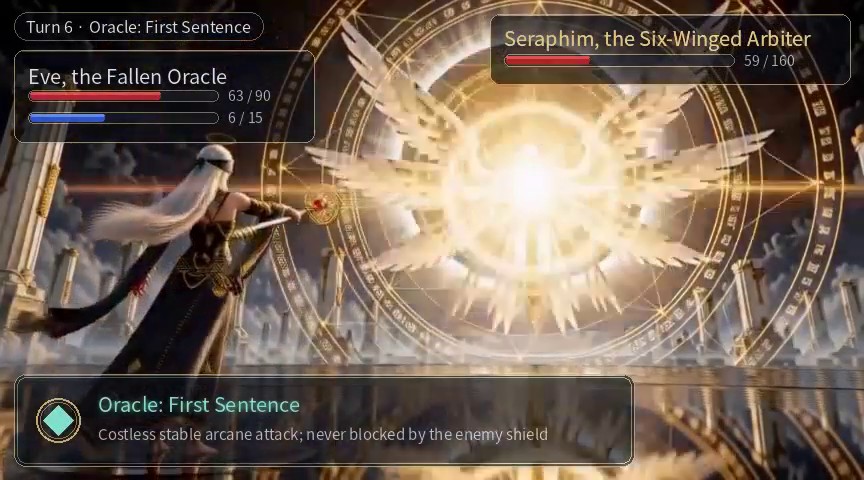}{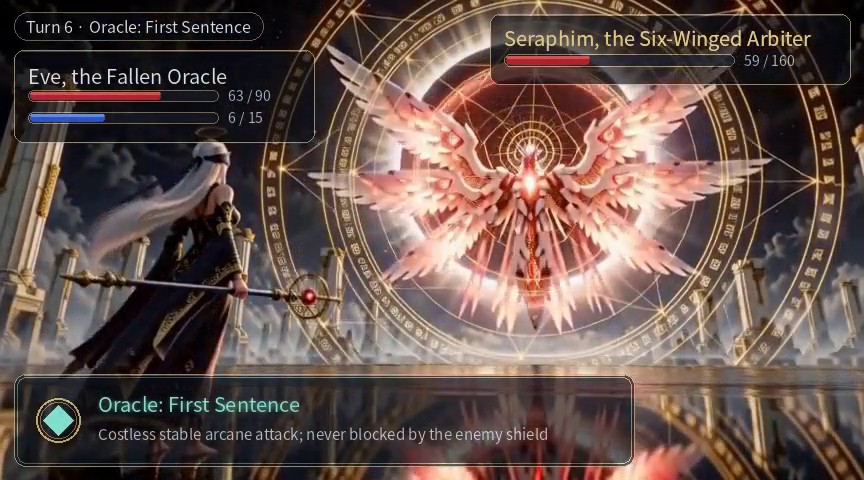}
\epcell{90}{0}{turn 7: zero-degree stigma}{arbiter 44/160 $\cdot$ eve 63/90, mana 5/15}{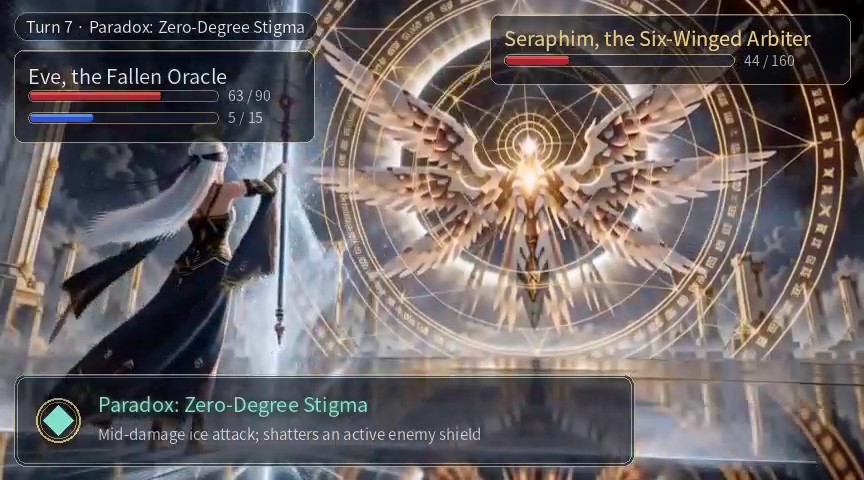}{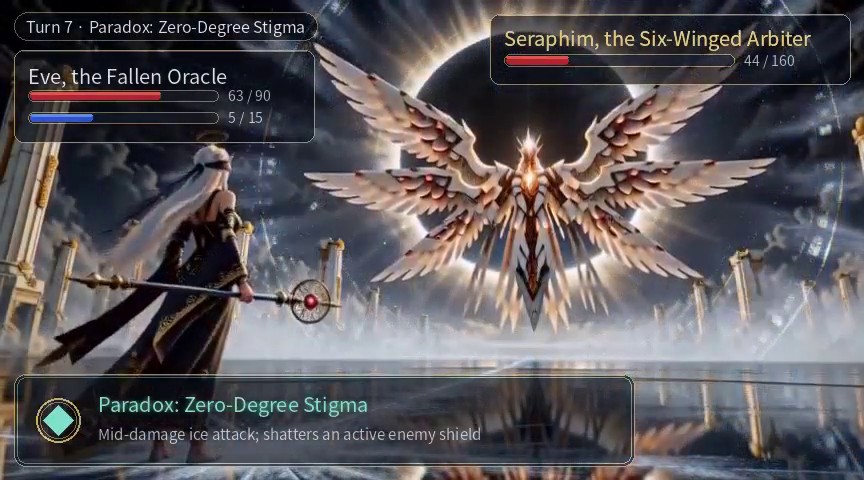}
\epcell{90}{-33.5}{turn 8: gate of the end}{arbiter held 44/160 $\cdot$ eve 63/90, mana 5/15}{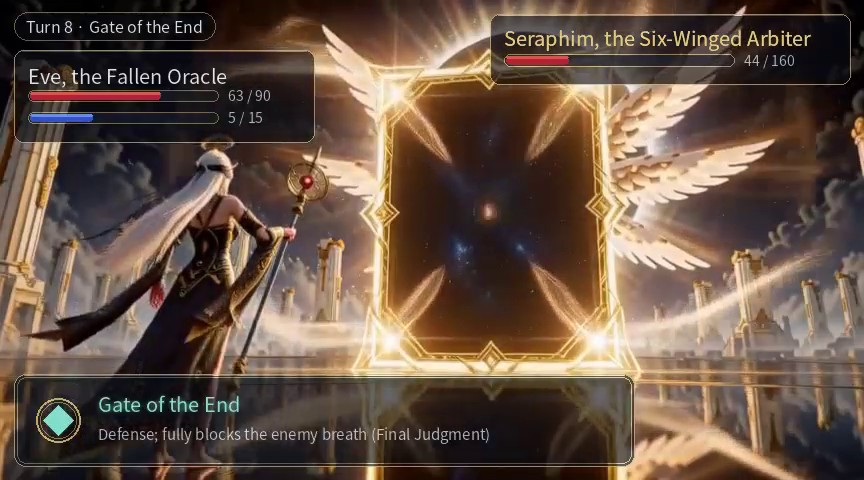}{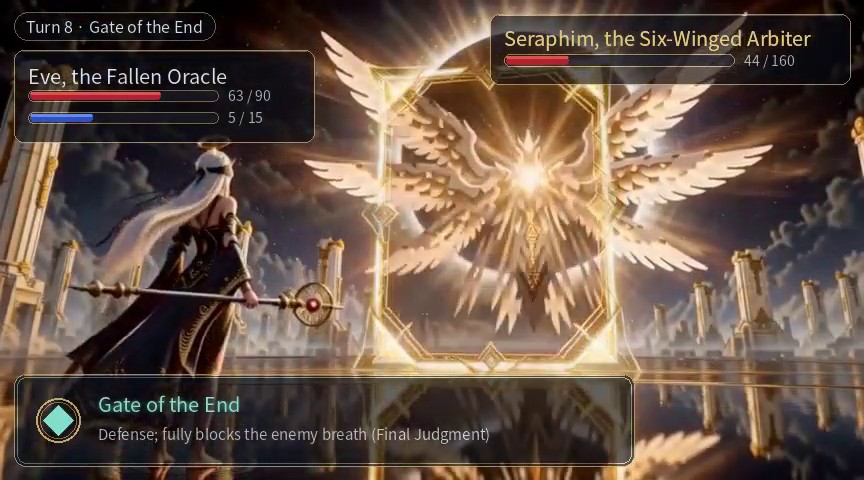}
\epcell{90}{-67}{turn 9: falling red sun}{arbiter 28/160 $\cdot$ eve 44/90, mana 3/15}{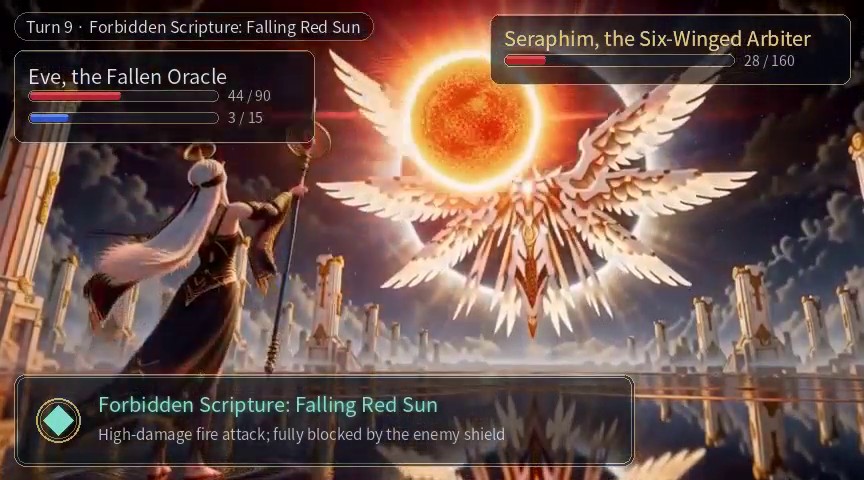}{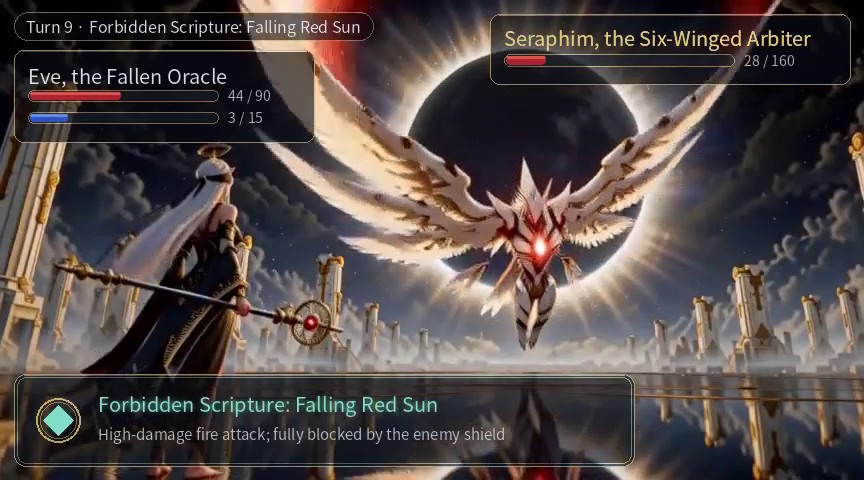}
\epcell{90}{-100.5}{turn 10: first sentence}{arbiter 6/160 $\cdot$ eve 24/90, mana 6/15}{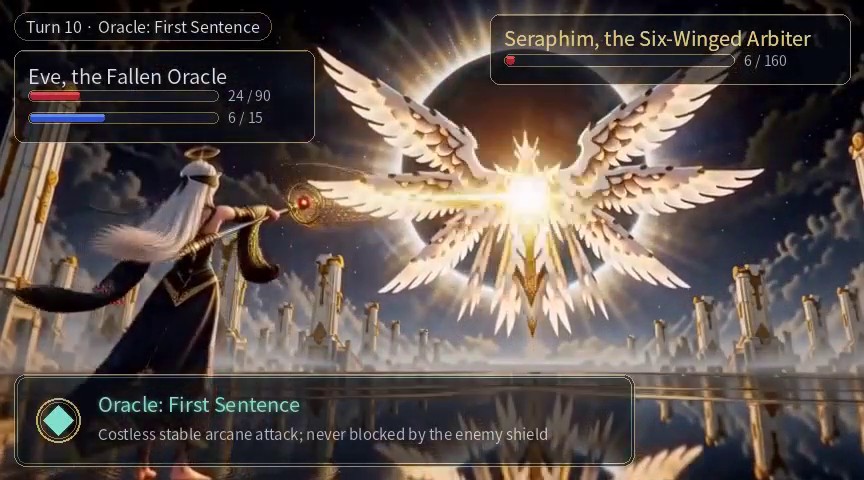}{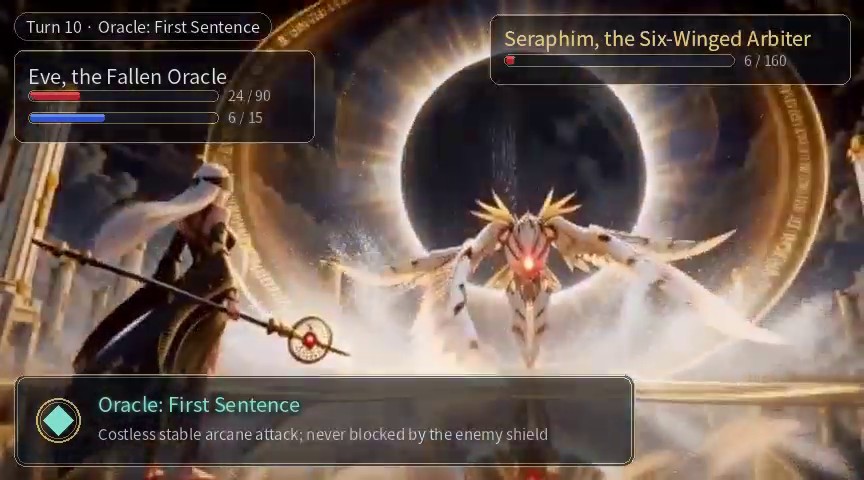}
\epcell{90}{-134}{turn 11: falling red sun}{arbiter 0/160 $\cdot$ eve 24/90, mana 1/15}{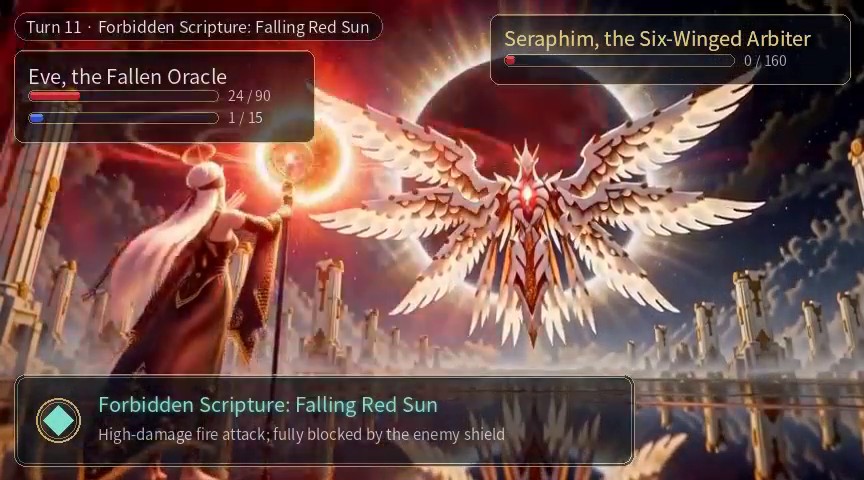}{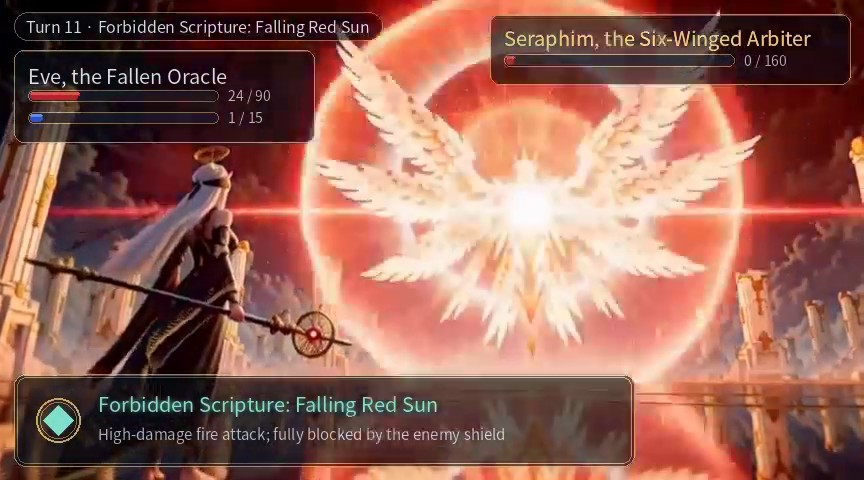}
\epcell{90}{-167.5}{victory}{the arbiter falls; the episode ends}{vict\_a.jpg}{vict\_b.jpg}
\end{tikzpicture}}
\caption{The full eleven-turn boss-duel episode of Figure~\ref{fig:demo} (Case C), released video of our harness on frozen MiniMax-H3.
Two frames per turn (native 16:9), with headers giving the skill and the HUD state (arbiter HP, eve HP, mana) at the turn's first frame.
The arbiter's HP drops monotonically 128$\to$0 across the eleven turns. Turn 8 plays the defensive Gate of the End, and the closing Falling Red Sun ends the duel in the victory screen.}
\label{fig:episode}
\end{figure}

Figure~\ref{fig:stamepisode} unrolls the first real episode (Case A of Figure~\ref{fig:demo}) act by act: all fifteen acts plus the ending card, two frames per act at native aspect, each header carrying the Traveler HUD state read at the act's first frame.
The stamina gauge drains from 28 to 0 as the walk proceeds, the status line switches to ``Exhausted --- slow walk only'', and rest restores it to 60 before the walk resumes. A second drain-and-recovery cycle follows on the way to the backstage door, while the quest counter advances only forward, from 1/3 through 2/3 to 3/3, and the location line moves from the sunlit street through the cobblestone square to the dim backstage corridor.
Figure~\ref{fig:loveepisode} unrolls the second real episode (Case B of Figure~\ref{fig:demo}) the same way: all twenty-nine turns plus the confession-accepted ending, two frames per turn at native aspect.
The affection counter rises monotonically from 47 to 75, one point per turn. The relationship state flips from friends to flirting at turn 14 and to lovers at turn 29, the mood starts shy and settles to happy, and the closing confession is accepted only after this accumulated state.
Figure~\ref{fig:episode} unrolls the third real episode (Case C of Figure~\ref{fig:demo}) turn by turn: all eleven turns of the boss duel plus the victory screen, two frames per turn at native aspect, each header carrying the in-video HUD state read at the turn's first frame.
The three attack skills cycle --- Forbidden Scripture: Falling Red Sun (heavy fire), Paradox: Zero-Degree Stigma (ice, shield-shattering), Oracle: First Sentence (costless arcane) --- the defensive Gate of the End is played at turn 8, and the final Falling Red Sun takes the arbiter from 6 to 0 between turns 10 and 11.

Figures~\ref{fig:promptb01} and~\ref{fig:promptb45} in Appendix~\ref{app:bench} provide the exact Rule-only and Rule+end prompts and the compiled \vsm{} instruction for one case from each condition.
The accompanying \textsc{Clip Verdict} annotations determine the verdict markers shown in these examples.

\section{Method Details: Formal Framework, Turn Lifecycle, and Domain Adaptation}
\label{app:method}

\subsection{Formal Framework and Cross-Channel Grounding}
\label{app:formal}

\paragraph{Generative Turn-Based Games as State Machines.}
The framework in Eq.~\ref{eq:state} maintains a serializable world state $s_t=(z_t,u_t)$ and accepts an action $a_t$ from the available action space.
The rule program $\tau_R$ determines a provisional semantic state, a spatial edit script, and an event as specified in Eq.~\ref{eq:settle}.
An optional neural kernel $\rho$ proposes dialogue or narration under the constraints in Eq.~\ref{eq:bounded}.
The compiler $\mathcal{C}$ converts the settled event and provisional state into inputs for the frozen video model $V_\theta$.
The observer $O_\psi$ interprets the generated clip, and $U_R$ incorporates observations compatible with the rules into the next state.
The final frame is retained as the visual anchor $f_{t+1}$.
This organization separates action resolution, visual generation, and state updates derived from observations.
Dependencies between software components follow the order game, engine, and service adapter, with game semantics and transport details assigned to their respective layers.

\paragraph{Rule Settlement before Rendering.}
The request to $V_\theta$ is issued after rule evaluation determines the provisional state $\bar s_{t+1}$ and event $e_t$.
The compiler produces $(p_t,r_t)=\mathcal{C}(\bar z_{t+1},\bar u_{t+1},e_t)$ before generating the current clip, while the anchor $f_t$ is inherited from the initial image or preceding clip.
A rendering failure does not reverse the event determined by the rule program.
The failure policy may retry the request, reuse the preceding anchor, or terminate generation.
Algorithm~\ref{alg:turn} specifies the successful execution path, including observation and state feedback.

\paragraph{Rule-Constrained Visual Feedback.}
Generated video provides evidence for updating state fields that are observable in the clip.
The VLM proposes observations, while the runtime validates their compatibility with the state schema and fixed rules before applying them.
An observation that contradicts the settled event is treated as a rendering inconsistency and cannot overturn the rule outcome.
Accepted observations may change the state used for subsequent actions, so state trajectories can depend on generated video.
If no update is accepted, the provisional state is retained.
This division preserves the authority of the rules while incorporating visual feedback into the interaction loop.

\paragraph{Deterministic Replay.}
The loop uses a seeded generator, with render seed $\mathrm{base\_seed}+t$ for turn $t$.
Rule settlement and compilation are reproducible for fixed input states, actions, programs, and accepted neural proposals.
Reproducing a complete episode additionally requires the same accepted observations because these observations affect later states and instructions.
Pixel-level replay requires deterministic backend behavior for fixed instructions, references, seeds, and serving conditions.
Frozen model weights alone do not establish this property.

\paragraph{Cross-Channel Grounding.}
The text instruction $p$ and spatial preview $r$ carry complementary, incomplete descriptions of one settled turn.
The preview encodes approximate layout, scale, and a camera trajectory, but its clay materials do not identify actors or events.
The text names actors, actions, and outcomes, but cannot uniquely specify object placement or camera motion.

\paragraph{Why Both Channels Matter.}
The preview gives the backend a concrete spatial reference.
The text specifies how to interpret that reference and which rule outcome to depict.
Preview renders sometimes reproduce the clay appearance when the instruction does not identify the reference as a layout preview.
A text description alone does not uniquely determine scene layout.
These observations motivate the two-channel design.
The WBench ablation measures the effect of the reference channel on external evaluation scores (\S\ref{sec:ablation}, Appendix~\ref{app:ablations}).

\paragraph{Shared Source of Camera Guidance.}
The clay model generates both the camera description in $p$ and the trajectory rendered into $r$.
Their intended paths therefore agree at the input level.
Whether the generated pixels follow that path is an empirical question for spatial evaluation, not a deterministic consequence of compilation.

\paragraph{Division of Authority.}
The ledger and scene are read and written by the harness.
Visual observations can update the ledger and scene only through the rule constrained update function $U_R$.
The renderer still controls appearance, fine motion, and how clearly the settled event is conveyed.
The runtime constrains accepted state updates, while the \textsc{EndFollow} metric measures how faithfully the generated clip depicts the required event.

\subsection{Turn Lifecycle and Domain Adaptation}

\begin{algorithm}[t]
\caption{One harness turn with visual state feedback}
\label{alg:turn}
\small
\textbf{Input.} State $s_t=(z_t,u_t)$, anchor $f_t$, action $a_t$, fixed rules $R$, frozen video model $V_\theta$, and observer $O_\psi$.
\begin{enumerate}\setlength{\itemsep}{1pt}
\item $(\bar z_{t+1},\epsilon_t,e_t)\gets\tau_R(s_t,a_t)$
  \hfill $\triangleright$ rule settlement
\item $\bar u_{t+1}\gets\mathrm{apply}(u_t,\epsilon_t)$, $\bar s_{t+1}\gets(\bar z_{t+1},\bar u_{t+1})$
  \hfill $\triangleright$ spatial update
\item $(p_t,r_t)\gets\mathcal{C}(\bar z_{t+1},\bar u_{t+1},e_t)$
  \hfill $\triangleright$ instruction and spatial preview
\item $v_{t+1}\gets V_\theta(p_t,r_t,f_t)$
  \hfill $\triangleright$ video generation
\item $o_{t+1}\gets O_\psi(v_{t+1},\bar s_{t+1},e_t)$
  \hfill $\triangleright$ visual observation
\item $s_{t+1}\gets U_R(\bar s_{t+1},o_{t+1})$
  \hfill $\triangleright$ validated state update
\item $f_{t+1}\gets\mathrm{last}(v_{t+1})$
  \hfill $\triangleright$ next visual anchor
\end{enumerate}
\end{algorithm}

\paragraph{Turn Lifecycle.}
World initialization is performed once per episode.
The per-domain program instantiates the game through registry $\mathrm{build}$.
The backend and kernel services are checked for availability before interaction begins.
The initial frame is chosen from the declared frame, a user-supplied image, or a placeholder in that order.
Each turn then runs nine steps.
(1)~The gate check ends the episode on terminal status, a reached turn budget, or an empty action space.
(2)~The player selects an action through $\mathrm{choose}(\mathrm{state},\mathrm{actions})$.
Scripted, random, human-CLI, and LLM-proposed players use the same rule layer.
In the deployed instances the proposal kernel and the LLM-proposed player are one resident Qwen3.8-27B-VL service \citep{qwen3}.
(3)~The rule program $\tau_R$ evaluates the action on a copy of the state without modifying its input.
This records the outcome before video generation.
(4)~A per-domain director maps the result and visual state into a \vsm{} script.
(5)~The compiler validates the script and rejects invalid input before generation.
(6)~Invalid rendering requests produce an error.
Transient errors trigger bounded retries, followed by the failure policy.
The default policy reuses the previous anchor and counts the turn as played.
(7)~The final frame, extracted with \texttt{ffmpeg -sseof}, becomes the next visual anchor.
(8)~The VLM interprets the generated clip and proposes changes to observable state fields.
The runtime validates these proposals against the fixed rules and incorporates accepted updates into the state for the next turn.
(9)~A JSON record stores the state before and after the turn, events, IR, compiled prompt, seed, latency, and render errors.
Reproducing the observation feedback additionally requires recording the visual observations and accepted state updates.
After the final turn, an ending director may render a presentation clip and a montage assembler may join the episode with state captions.
Neither operation changes the terminal state fixed by the rules.
Table~\ref{tab:turnloop} summarizes the input and output of each operation.

\begin{table}[h]
\centering
\caption{The full turn lifecycle with per-step inputs and outputs.
Step~0 initializes the world, steps~1 to 9 repeat for each turn, and steps~10 and~11 produce the final presentation.
The table extends Algorithm~\ref{alg:turn} with the harness operations surrounding the render call.}
\label{tab:turnloop}
\small
\setlength{\tabcolsep}{4pt}
\begin{tabular}{@{}r l p{7.6cm}@{}}
\toprule
Step & Operation & Input $\to$ output \\
\midrule
0 & setup (once) & domain declaration $\to$ runtime registry, service
preflight, initial anchor $f_0$ \\
1 & gate check & terminal status, turn budget, legal actions $\to$ continue
or stop \\
2 & action selection & $s_t$, legal actions $\to a_t$ from a scripted, random, human, or LLM player \\
3 & settle & $s_t,a_t\to(\bar z_{t+1},\epsilon_t,e_t)$ through Eq.~\ref{eq:settle} \\
4 & direct & $\bar s_{t+1},e_t\to$ \vsm{} script \\
5 & compile & script and scene $\to(p_t,r_t)$, with invalid records rejected before generation \\
6 & render & $(p_t,r_t,f_t)\to v_{t+1}$ with bounded retries and a failure policy \\
7 & anchor update & $v_{t+1}\to f_{t+1}=\mathrm{last}(v_{t+1})$ \\
8 & observe and update & $(v_{t+1},\bar s_{t+1},e_t)\to o_{t+1}\to s_{t+1}$ through observation and rule validation \\
9 & record & states before/after, events, script, prompt, seed, latency,
errors $\to$ one JSON line \\
10 & ending (once) & terminal status $\to$ presentation clip, deterministically
seeded \\
11 & montage (once) & clips and state captions $\to$ replay video \\
\bottomrule
\end{tabular}
\end{table}

\paragraph{Clay-Model Construction.}
One forward pass of a feed-forward geometry model \citep{wang2025vggt} produces depth, world points, and part features from the current frame.
Clustering yields per-instance oriented boxes.
Scale is anchored by the 1.7\,m human prior and recovered camera height.
The boxes instantiate an untextured Blender \citep{blender} scene with the CV-to-Blender coordinate correction, and procedural camera programs (push-in, orbit, head-push) render the preview clip when a turn needs camera or layout motion.
The scene file records boxes and camera explicitly and supports edits by rules or humans.
The deployed spatial channel in \S\ref{sec:spatial} uses the single-subject configuration of this pipeline.
WildDet3D, a 1.2B-parameter detector built on SAM3 \citep{sam3} and LingBot-Depth, localizes the subject in the anchor frame.
An oriented box is positioned by pinhole anchoring under the real-height prior.
A ground plane is fitted to a depth region adjacent to the subject and checked against the pinhole ray.
Blender renders a single clay preview under the original camera, aligned pixel-wise with the input view.
The synthesizer is deterministic given its inputs and never writes to the ledger.
When the instruction does not identify the reference as a layout preview, the backend can reproduce the clay materials instead of using the preview solely as spatial guidance. The instruction therefore identifies it as a layout preview.

\paragraph{Domain Adaptation.}
Adaptation to a new domain specifies its state, rules, and content in a game definition $D_i$ while retaining the shared runtime and video model.
In all reported experiments, this game definition is written by the coding agent based on GLM-5.3 \citep{glm53}.
The common interfaces include the ledger schema, rule engine, \vsm{} fields, spatial synthesis, and visual anchoring, while $D_i$ specifies the rules and content of each domain.
Each \benchname{} case program is one such definition scoped to a single scenario.

\paragraph{Gate Order.}
The four validation stages are schema checking, value clamping, bounded retry, and fallback.
Parsing constrains the syntax, and clamping constrains the values.
The runtime retries the kernel up to a fixed limit before substituting the default specified by the rule program.
Containment (Eq.~\ref{eq:bounded}) depends on each accepted proposal being checked against the allowed set $\mathcal{A}_{\tau}(s,a)$, including the fallback.
A finite schema alone does not establish that containment (\S\ref{sec:ledger}).

\subsection{DSL and IR Specifications}
\label{app:dsl}

The system contains two typed records, checked before use.
The \emph{case program} defines benchmark instances, while the \emph{\vsm{} record} provides the intermediate representation compiled into a backend instruction (\S\ref{sec:ir}).

\paragraph{Case Program.}
A case is one declarative program.
Expected outcomes are derived from the program semantics.
The grammar specifies fixed fields that are validated when the program is loaded.

\begin{verbatim}
case ::= {
  id, theme,
  frame_prompt,            # FLUX.2 first-frame prompt (audited)
  hidden  ::= { name, value_A, value_B, invisible_evidence[] },
  action  ::= { actor, verb, target, phase[] },
  rules   ::= [ branch_A, branch_B ],  # symmetric, exclusive
  branch  ::= { when: hidden.value == v,
                end_state, forbidden[],
                evidence_phrases[] }
}
\end{verbatim}

\textbf{Properties.}
(i) \emph{Derived Outcomes.} Evaluating the program determines the required ending, forbidden outcomes, and expected answers to the judging questions.
Adding a case therefore requires specifying a program from which these outcomes can be reproduced.
(ii) \emph{Twin Symmetry.} A linter checks that the rule branches are mutually exclusive and that the paired inputs differ only in the hidden value and its consequences.
(iii) \emph{Hidden State Isolation.} The \texttt{frame\_prompt} must not reveal the hidden value.
The audit of Appendix~\ref{app:bench} enforces this on the rendered frame itself.

\paragraph{The \vsm{} Record: Typed Slots.}
The game layer describes each settled turn through a fixed record, and the compiler translates this record into prompt syntax.
Table~\ref{tab:vsm} lists the slots.
The record specifies the actor, action, and required ending independently of the backend instruction format.
The adapter determines the prompt format, and validation rejects invalid records before generation.

\begin{table}[h]
\centering
\caption{The \vsm{} record: typed slots filled by the game layer, each with a machine-checked validation rule.}
\label{tab:vsm}
\small
\begin{tabular}{@{}l p{9.6cm}@{}}
\toprule
Slot & Content \\
\midrule
style & one sentence fixing art style, palette, and quality register \\
first-frame anchor & description that must match the anchor image \\
camera & camera motion, drawn from a fixed phrase vocabulary \\
actor (+ appearance) & acting entity and appearance attributes that preserve identity across scenes \\
action & what the actor does, with prepositions \\
target (+ reaction) & the acted-upon party and its response \\
dialogue & verbatim lines with speaker tags \\
effects & emphasis phrases (impacts, glows, weather) \\
stability & anti-drift constraints on identity, clothing, and position \\
end state & required final state used to describe the next visual anchor \\
soundscape / music & ambient and non-diegetic audio, sentence-bounded \\
shots & declared scene cuts with timestamps assigned by the compiler \\
meta & private metadata, never rendered \\
use-first-frame & whether the clip is conditioned on the anchor (chain exemption) \\
duration & clip length in seconds \\
\bottomrule
\end{tabular}
\end{table}

In detail, its \emph{style} and \emph{first-frame anchor} slots describe the desired appearance and restate the anchor image.
The \emph{camera} slot uses a fixed vocabulary for motion, while the clay model provides the 3D trajectory.
The \emph{actor} and \emph{appearance} slots carry textual identity across scenes.
The \emph{action}, \emph{target}, \emph{reaction}, \emph{dialogue}, and \emph{effects} slots describe the event.
Dialogue includes speaker ids and constraints for off-screen speech.
The \emph{stability} slot constrains identity, clothing, and position.
The \emph{end state} specifies what the next anchor should show.
The record also includes \emph{soundscape}, \emph{music}, and \emph{shots} with validated sentence budgets and cut times.
A \emph{use-first-frame} flag may break the anchor chain on scene changes, leaving appearance slots to carry identity (\S\ref{sec:ir}).
The record states who acts and how the turn ends within the roughly 5\,s clip budget.
Validation rejects illegal records before compilation, and the backend adapter owns the prompt dialect.

\paragraph{Compiler Validation and Assembly Order.}
Compilation either produces an instruction or rejects the record before generation.
Identical valid records produce identical instructions under the same adapter.
(i)~\emph{Schema Validation.} The action, actor, and end-state fields must be nonempty.
Free text descriptions cannot contain shot markers, and multiple shots must be declared in the shots field.
(ii)~\emph{Budget Validation.} The compiler checks sentence limits for each field and the uniqueness of speaker tags.
A violation raises an error before rendering.
(iii)~\emph{Assembly.} The compiler orders style, first-frame anchoring, camera motion, actor action, reaction, dialogue, effects, declared shots, stability constraints, and the required end state.
It assigns shot timestamps while reserving a final interval for the ending.
The stability constraints and final state are expressed through fixed sentence templates.
(iv)~\emph{Adapter Fields.} The adapter appends the protocol fields required by the backend, including the first-frame alignment instruction in image-conditioned mode or its replacement in text-only mode.
Separating the instruction format from the game definition allows the same game code to support backends with different input grammars.

\paragraph{Compiler Determinism.}
The \vsm{} compiler is a deterministic function of the script and backend adapter, with no sampling, model calls, or formatting that depends on the environment.
Given the same inputs, it produces the same instruction string, allowing compilation and backend adaptation to be checked independently of video generation (\S\ref{sec:ir}).

\section{Positioning against Concurrent Explicit-State Systems}
\label{app:positioning}

Table~\ref{tab:pwm} extends the comparison in \S\ref{sec:related} by summarizing state representations, control interfaces, spatial state handling, and evaluation protocols.

\begin{table}[t]
\centering
\caption{Positioning against concurrent explicit-state systems.
``Native ref.'' denotes the standard reference inputs of a frozen model.}
\label{tab:pwm}
\small
\renewcommand{\arraystretch}{1.3}
\resizebox{\textwidth}{!}{%
\begin{tabular}{@{}lcccc@{}}
\toprule
 & State representation & Control interface & Spatial state & Evaluation \\
\midrule
PWM \citep{huang2026programmable} & engine and 3D OBB IR & custom pixel-level conditioning & per-segment signal & combat count/death \\
CWM \citep{chen2026codeworld} & coding-agent code and proxy video & fine-tuning & geometry constraint (proxy) & qualitative adherence \\
WorldMind \citep{deng2026worldminddecoupledgameworld} & recovered from frames & Wan backend & recovered from pixels & human preference \\
Unbounded \citep{li2025unbounded} & character-life variables & language router & none (2D narrative) & user study (20 raters) \\
\midrule
\ocell\textbf{Ours} & \ocell\textbf{ledger + \vsm{} IR + clay model} & \ocell\textbf{native ref.\ + compiled text} & \ocell\textbf{persistent scene asset} & \ocell\textbf{Change$\to$Comply twins} \\
\bottomrule
\end{tabular}}
\end{table}

\section{Reproducibility Details}
\label{app:repro}

\paragraph{Required Artifacts.}
Independent reproduction requires the 50 case programs, first-frame prompts and images, audit records, rule programs, \vsm{} grammar and compiler, backend adapters and generation settings, sampled clips, fixed choice judge prompts, and per-clip verdicts.
The case programs determine the expected outcomes and the compiler is deterministic, so each reported number is traceable from a case program through compilation to its per-clip verdicts.

\paragraph{Determinism Boundary.}
Rule settlement and compilation are deterministic for fixed programs, input states, actions, and accepted neural proposals.
The optional proposal kernel, video backend, and visual observer may produce different outputs across runs.
The final frame of each clip conditions subsequent generation, and accepted visual observations affect the next world state.
Reproducing an episode therefore also requires the same observation stream and accepted state updates.
A reproducibility package should record backend and observer versions, inference settings, reference clips, sampled frames, observations, and accepted updates.

\paragraph{Backend Interfaces.}
MiniMax-H3 is used with text instruction and image anchor, and the deployed instruction dialect follows the official H3 prompt-writing guide distributed with the model.
The full harness additionally supplies a preview clip through its video-reference input.
Unwrapped interactive backends are driven through their available action interfaces.
No backend weights are modified for the reported comparison.
The adapter isolates the backend-specific instruction dialect, so extending to further backends is instruction-level work.

\paragraph{Compute Reporting.}
The manuscript states that evaluations used a single 8-GPU node and that the harness logic ran on CPUs.

\section{Limitations and Future Work}
\label{sec:limitations}
\label{sec:discussion}

The box-based clay model provides a coarse representation of spatial layout.
Errors in instance recovery propagate to the preview, and the text instruction does not always prevent the renderer from reproducing the clay appearance.
Articulated pose and fine appearance are determined by the video model, and the spatial pipeline operates offline.

Future work targets a more capable harness: a richer rule language with temporal and quantified conditions, state that tracks more entities and relations, and compiled instructions that carry the same frozen weights to further video backends.
Beyond the current case families, we plan to scale the closed loop to more diverse world scenarios and settings, including longer episodes, multi-agent interactions, and open-ended environments where goals and rules evolve over time.

\subsubsection*{Reproducibility Statement}
Reproducing the reported results requires the 50 case programs and first frames, the \vsm{} grammar and compiler, backend instructions, sampled clips, fixed-choice judge prompts, and per-clip verdicts.
Appendix~\ref{app:repro} specifies these materials and the determinism boundary.
All case programs, prompts, adapters, judge templates, and per-clip verdicts will be released upon publication.

\end{document}